\pdfoutput=1
\documentclass{article}

 \usepackage[main, final]{neurips_2026}
\usepackage[utf8]{inputenc}
\usepackage[T1]{fontenc}
\usepackage{hyperref}
\usepackage{url}
\usepackage{booktabs}
\usepackage{array}
\usepackage{amsfonts}
\usepackage{amsmath}
\usepackage{amssymb}
\usepackage{amsthm}
\usepackage{nicefrac}
\usepackage{microtype}
\usepackage{xcolor}
\usepackage{graphicx}
\graphicspath{{figs/}}
\usepackage{subcaption}
\usepackage{multirow}
\usepackage{algorithm}
\usepackage{algorithmic}
\usepackage{enumitem}
\setlist{nosep,leftmargin=*}

\definecolor{paperblue}{HTML}{1A42A6}
\hypersetup{colorlinks=true, linkcolor=paperblue, citecolor=paperblue,
            urlcolor=paperblue, filecolor=paperblue,
            pdftitle={Pattern-Based Matrix Analysis Under Permutations},
            pdfauthor={Andrei Ermakov}}

\usepackage{float}
\usepackage{tcolorbox}
\tcbuselibrary{skins,breakable}
\colorlet{thmback}{paperblue!8!white}
\tcbset{thmbox/.style={enhanced, frame hidden, boxrule=0pt,
    arc=3pt, colback=thmback, boxsep=0pt, left=6pt, right=6pt,
    top=4pt, bottom=4pt, parbox=false,
    before skip=7pt plus 2pt minus 2pt, after skip=7pt plus 2pt minus 2pt,
    before upper={\setlength{\partopsep}{0pt}}}}
\newtheoremstyle{boxedplain}{0pt}{0pt}{\itshape}{}{\bfseries}{.}{ }{}
\theoremstyle{boxedplain}
\newtheorem{theorem}{Theorem}
\newtheorem{lemma}{Lemma}
\newtheorem{proposition}{Proposition}

\tcolorboxenvironment{theorem}{thmbox}
\tcolorboxenvironment{lemma}{thmbox}
\tcolorboxenvironment{proposition}{thmbox}
\tcolorboxenvironment{corollary}{thmbox}
\newtheorem{definition}{Definition}
\tcolorboxenvironment{definition}{thmbox}
\theoremstyle{plain}

\newcommand{\dM}{d_M}
\newcommand{\PiM}{\Pi_M}

\newcommand{\norm}[1]{\left\|#1\right\|_F}
\newcommand{\fnorm}[1]{\left\|#1\right\|_F}
\newcommand{\R}{\mathbb{R}}
\newcommand{\Sn}{\mathcal{S}_n}
\DeclareMathOperator{\Aut}{Aut}

\DeclareMathOperator{\rank}{rank}
\DeclareMathOperator{\diag}{diag}

\DeclareMathOperator{\supp}{supp}
\DeclareMathOperator*{\argmin}{arg\,min}

\newcommand{\E}{\mathbb{E}}
\title{Pattern-Based Matrix Analysis Under Permutations}

\author{%
  Andrei Ermakov \\
  42 NL \\
  \texttt{ermakovhse@gmail.com}
}

\begin{document}

\maketitle

\begin{abstract}
Low-rank approximation captures one important notion of matrix simplicity: a
matrix is simple if it is well described by a low-dimensional linear model. This
paper studies a complementary notion. We consider matrices that become
structurally simple only after permuting rows and columns, so
that their mass concentrates in a prescribed geometric pattern such as a band,
block-diagonal, arrowhead, or related form.
We introduce a pattern distance that measures how close a matrix is to a given
pattern after optimal reordering, giving a common framework for
structural problems such as latent ordering, community detection, and
bandwidth minimization. The problem is NP-hard
in general; we derive an alternating algorithm based on Linear Assignment
subproblems. We further introduce a
pattern-plus-low-rank decomposition that combines permutation-structured fitting
with a low-rank residual, whose permutation step is pattern fitting on
the residual; its gains across pattern families persist at a matched
parameter budget and are substantially reduced under a density-matched
random-mask control.
We evaluate the framework on synthetic recovery tasks, real correlation and graph
adjacency matrices, and transformer attention heads,
showing that a single universal algorithm recovers qualitatively different
forms of matrix structure.
\end{abstract}

% =============================================================================
\section{Introduction}
\label{sec:intro}

A basic question in matrix analysis is what structure a matrix has.
The standard answer is low-rank: the SVD gives the best rank-$k$
approximation and reveals latent linear factors. But many matrices are
simple in another sense: after reordering rows and columns they show an
interpretable geometric form, blocks, bands, chains or hierarchies.
Seriation~\citep{atkins1998spectral,fogel2013seriation}, community
detection~\citep{newman2006modularity,holland1983stochastic} and
bandwidth minimization~\citep{cuthill1969reducing,george1973nested} are
usually studied separately yet share this intuition.
\citet{goldengorin2017linear} make it precise:
many combinatorial problems (LAP, TSP, linear ordering, matching,
cut) reduce to finding a permutation that aligns a matrix with a
geometric \emph{pattern}, and fixing one permutation reduces the
remaining search to a Linear Assignment Problem.

This paper turns that combinatorial idea into a matrix-analysis tool.
A binary \emph{pattern mask} $M$ specifies the target geometry; the
\emph{pattern distance} $d_M(A)$ is the off-pattern mass after the best
row/column permutations.
This viewpoint, which we call \emph{permutation-structural
simplicity}, complements low-rank approximation: a matrix may have
clear geometric organization after reordering even when no low-rank
model fits, and vice versa; the target geometry, the discrepancy
measure and the search over permutations become independent
ingredients of one formalism.

\subsection{Contributions}

\begin{enumerate}
\item \textbf{Pattern-based framework.}
We define the \emph{pattern distance} $d_M$, the off-pattern Frobenius
mass after optimal reordering, and use it to formalize
permutation-structural simplicity.

\item \textbf{Alternating LAP algorithm.}
An alternating procedure built on the auxiliary-matrix construction
of~\citet{goldengorin2017linear} performs exact one-sided updates of
$d_M$ in $O(n^3)$, applicable to any binary mask without changes to the
solver.

\item \textbf{Pattern-plus-low-rank (PatLR) decomposition.}
A hybrid model combining permutation-structured fitting with a
low-rank residual, with exact closed-form updates for both parts and a
monotone alternating algorithm.

\item \textbf{Adaptive pattern selection.}
When the pattern is unknown, a simulated-annealing search jointly
optimizes the pattern and the permutation against a
density-normalized score.

\item \textbf{Experimental validation.}
On synthetic and real data the same algorithm recovers qualitatively
different structures, scored by the metrics of latent ordering,
consecutive ones and block recovery against the methods developed for
them; on the attention heads of three transformers the same score gives
labels with a shared depth profile.
\end{enumerate}

\subsection{Related Work}

\textbf{Structure after reordering.}
Several literatures study matrices that become simple once rows and
columns are permuted, each around one geometry.
Seriation seeks a Robinson order by spectral~\citep{robinson1951method,atkins1998spectral}
or convex~\citep{fogel2013seriation} methods, with the consecutive-ones
property~\citep{booth1976c1p,kendall1971seriation} and circular
seriation~\citep{recanati2018circular} as its interval and cyclic forms.
Community detection seeks block-diagonal structure, by modularity and
spectral methods~\citep{girvan2002community,vonluxburg2007tutorial} or
under the stochastic block model, where belief propagation and its
spectral linearizations are optimal near the detectability
threshold~\citep{holland1983stochastic,decelle2011asymptotic,krzakala2013spectral,saade2014bethe}.
Bandwidth and profile reduction seek near-diagonal
sparsity~\citep{cuthill1969reducing,gibbs1976gps,george1973nested};
nestedness~\citep{brualdi1999nested}, bordered block-diagonal
forms~\citep{aykanat2004bordered}, chordal
sparsity~\citep{vandenberghe2015chordal}, core--periphery
structure~\citep{rombach2014core} and
biclustering~\citep{hartigan1972biclustering} are further such
geometries (Table~\ref{tab:patterns}). Robust PCA separates a low-rank
part from a sparse one~\citep{candes2011robust} with no permutation;
GoDec~\citep{zhou2011godec} alternates low-rank and sparse
approximation, whereas PatLR restricts the support of the second
component to a geometric mask under unknown permutations.

\textbf{A common optimization core.}
\citet{goldengorin2017linear} introduced patterns as a description of
the input matrices of combinatorial optimization problems. For a fixed
pattern the alignment is a Koopmans--Beckmann quadratic assignment
problem~\citep{koopmans1957assignment,lawler1963qap,cela1998qap},
with two permutations a bilinear assignment
problem~\citep{custic2017bilinear}, for which alternating LAP
heuristics are studied by \citet{sokol2020bilinear}. Generic solvers
are the FAQ heuristic~\citep{vogelstein2015faq} and Birkhoff-polytope
relaxations~\citep{fogel2013seriation,cuturi2013sinkhorn,mena2018gumbel}.
In this paper the pattern is the interface between these literatures:
the same solver serves every mask, and PatLR extends it with a low-rank
residual.

\begin{figure}[t]
  \centering
  \includegraphics[width=\linewidth]{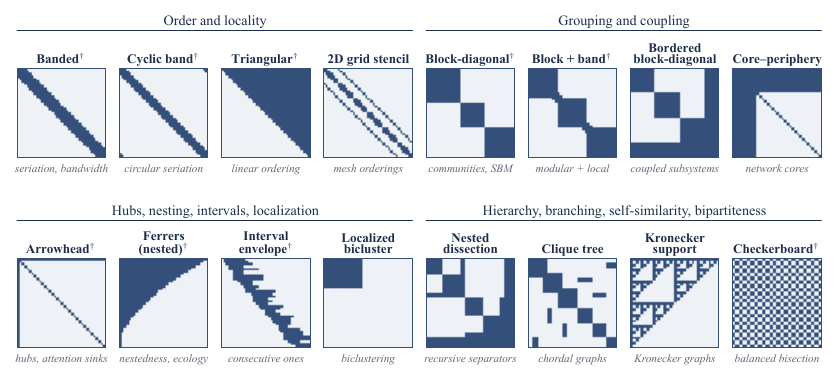}
  \caption{Gallery of patterns, one representative per structural
    hypothesis (each cell is the binary mask $M$, blue $=1$), grouped by
    the kind of structure they encode:
    order, grouping, nesting, separability, branching, self-similarity.
    Daggers mark the families used in our experiments;
    Table~\ref{tab:patterns} gives the classical problem behind each pattern.
    Masks are shown up to the permutations the framework optimizes over
    (an anti-diagonal band equals a band under column reversal). Any
    binary mask is admissible: $M$ encodes the structure hypothesized for
    an application or is selected from a vocabulary.}
  \label{fig:gallery}
\end{figure}
\section{Theory of Patterns}
\label{sec:prelim}

A \emph{pattern} is a target structure (a band, block-diagonal,
arrowhead, etc.), encoded by a binary \emph{mask}
$M\in\{0,1\}^{n\times n}$ marking the positions it allows; we use
``pattern'' for the structural hypothesis and ``mask'' for the matrix
$M$.
A permutation $\pi\in\Sn$ maps an original index to its new position,
$(P_\pi)_{ij}=\mathbf{1}\{i=\pi(j)\}$, so the $(\pi,\sigma)$-reordered
matrix $A^{\pi,\sigma}:=P_\pi A P_\sigma^\top$ has entries
$A_{\pi^{-1}(i),\sigma^{-1}(j)}$; $\Pi_M(X):=M\odot X$ is the masking
operator. For fixed row and column
permutations the \emph{off-pattern mass} is
\begin{equation}
d_M(A;\pi,\sigma) \;:=\; \bigl\|(I-\Pi_M)(A^{\pi,\sigma})\bigr\|_F^2
\;=\; \sum_{(i,j):\,M_{ij}=0} \bigl(A^{\pi,\sigma}_{ij}\bigr)^2 .
\label{eq:offmass}
\end{equation}

\begin{definition}[Pattern distance]
\label{def:pattern_dist}
The pattern distance of $A\in\R^{n\times n}$ from the mask
$M\in\{0,1\}^{n\times n}$ is the smallest off-pattern mass over row and
column permutations,
\begin{equation}
d_M(A) \;:=\; \min_{\pi,\sigma\in\Sn} d_M(A;\pi,\sigma).
\label{eq:pattern_dist}
\end{equation}
\end{definition}

For $A\neq0$ the \emph{explained pattern variance}
$\nu_M(A) := 1 - d_M(A)/\|A\|_F^2 \in [0,1]$ is the fraction of matrix
energy placed inside the pattern after optimal reordering.
\label{eq:nuM}
For symmetric matrices and for the attention matrices we also minimize
over a common permutation, $d_M^{\mathrm{com}}(A) := \min_{\pi}
d_M(A;\pi,\pi)$ (the case $\sigma=\pi$), writing $d_M(A;\pi) :=
d_M(A;\pi,\pi)$ at a fixed $\pi$; the restriction shrinks the feasible
set, so $d_M^{\mathrm{com}}(A)\ge d_M(A)$. The mask density is
$\rho_M := |M|_1/n^2$ ($|M|_1$ the number of ones). $d_M(A)=0$ iff $A$
fits inside $M$ after some reordering, and under independent uniform row
and column permutations a fraction $\rho_M$ of the energy lies inside
the mask in expectation (Appendix~\ref{app:basic}).

\textbf{Example (two notions of matrix simplicity).} Pattern
simplicity and low rank are independent: neither bounds the other. Let
$A = P S_0 P^\top$ with $S_0$ $k$-block-diagonal, constant blocks of size
$b=n/k$ and entries $\mu>0$. The $k$-block mask fits $A$ exactly,
$d_M(A)=0$, while $A$ has $k$ equal singular values $\mu b$, so its best
rank-$r$ approximation has squared Frobenius error $(k-r)\,b^2\mu^2$
for every $r<k$: what the mask describes by its support, low rank
describes only at rank $k$. Conversely a generic rank-one $uu^\top$,
the simplest matrix for the SVD, fits no mask that leaves an entry out,
$d_M(uu^\top)>0$. A matrix can thus be simple in one sense and not in
the other, which is why Section~\ref{sec:patlr} fits both parts jointly
instead of choosing one.

\paragraph{Hardness.}\mbox{}  % run-in label on its own line before the box (same layout under pdflatex and xelatex)
\begin{theorem}[NP-hardness of pattern distance]
\label{thm:nphard}
Computing $d_M(A)$ exactly is NP-hard for the $2$-block-diagonal mask,
for independent $(\pi,\sigma)$ as well as under $\sigma=\pi$.
\end{theorem}
\emph{Proof sketch.} Reduction from \textsc{Minimum
Bisection}~\citep{garey1979computers} through the adjacency matrix of
the graph with a heavy diagonal, which forces the row and column
bisections to coincide; the band of width $1$ under $\sigma=\pi$ is
NP-hard too, as a maximum-weight Hamiltonian path
(Appendix~\ref{app:hamiltonian}). Hardness is mask-specific: the
arrowhead under $\sigma=\pi$ and the diagonal and anti-diagonal masks
with two permutations are solved exactly (Appendix~\ref{app:special}).

\paragraph{Permutation ambiguities.}
An exact ordering need not be unique. If a noiseless signal is nonzero
at every position allowed by the mask and zero elsewhere, all zero-error
orderings differ only by symmetries of that mask
(Appendix~\ref{app:ambiguity}). Under a common row/column permutation
these include the reversal of a banded order and arbitrary reorderings
within blocks and among equal-sized blocks; for an arrowhead mask with
$n\ge3$ the hub is identifiable, but the remaining indices have no
identifiable order.

% =============================================================================
\section{Alternating LAP}
\label{sec:qap}

Computing $d_M(A)$ is a search over two permutations at once: since
permutations preserve $\|\cdot\|_F$, $d_M(A) = \|A\|_F^2 -
\max_{\pi,\sigma}\sum_{i,j}M_{ij}\, A_{\pi^{-1}(i),\sigma^{-1}(j)}^2$, a
bilinear assignment objective, a QAP under $\sigma=\pi$ (NP-hard,
Theorem~\ref{thm:nphard}).
Building on the alternating LAP schemes
of \citet{goldengorin2017linear} and \citet{sokol2020bilinear}, we
derive its assignment updates.

\begin{lemma}[Column-separability]
\label{lem:col_sep}
Fix $\pi_r$ and let $B = P_{\pi_r} A$. With
$Q_{p,c} := \sum_{i} (1-M_{i,p})\, B_{i,c}^2$,
\begin{equation}
\min_{\sigma\in\Sn}
\bigl\|(I-\Pi_M)(B P_\sigma^\top)\bigr\|_F^2
\;=\;
\min_{\sigma\in\Sn}\sum_c Q_{\sigma(c),c} ,
\label{eq:lap_reduction}
\end{equation}
a Linear Assignment Problem whose solution $\mathrm{LAP}(Q)$ assigns
each original column $c$ its position $\sigma(c)$. The row case is
symmetric (Appendix~\ref{app:proofs}).
\end{lemma}

\noindent
Here $Q$ is the auxiliary cost matrix of \citet{goldengorin2017linear}:
$Q_{p,c}$ is the off-pattern cost of placing original column $c$ at
position $p$ under the fixed row order; since these costs do not depend
on where the other columns go, choosing the column permutation is
exactly a linear assignment problem. Alternating the two one-sided
problems gives Algorithm~\ref{alg:alternating_lap}: from a spectral
start (the Fiedler order~\citep{barnard1995spectral}; Appendix~\ref{app:init}), fix the rows and
solve a column LAP, fix the columns and solve a row LAP, and repeat;
random restarts guard against poor local minima. A step costs $O(n^3)$
(auxiliary matrix and Hungarian solver), a run $O(RTn^3)$ for $R$
restarts of $T$ iterations. Below,
$\textsc{AltLAP}(A, M, T;\, (\pi_0,\sigma_0))$ denotes $T$ iterations
of Algorithm~\ref{alg:alternating_lap} from $(\pi_0,\sigma_0)$ (spectral
start if the pair is omitted).

\begin{algorithm}[tbp]
\caption{Alternating LAP (worked example in Appendix~\ref{app:example})}
\label{alg:alternating_lap}
\begin{algorithmic}[1]
\REQUIRE $A \in \R^{n\times n}$, mask $M$, iterations $T$, restarts $R$
\STATE $\hat\pi \leftarrow$ Fiedler ordering
\FOR{$r = 1, \ldots, R$}
\STATE $\pi_0 \leftarrow \hat\pi$ if $r{=}1$, else a random permutation; $(\pi,\sigma) \leftarrow (\pi_0,\pi_0)$
\FOR{$t = 1, \ldots, T$}
\STATE $Q_{pc} \leftarrow \sum_i (1{-}M_{ip})\, (P_\pi A)_{ic}^2$; \quad $\sigma \leftarrow \mathrm{LAP}(Q)$
\STATE $Q^{\mathrm{row}}_{pi} \leftarrow \sum_j (1{-}M_{pj})\, (A P_\sigma^\top)_{ij}^2$; \quad $\pi \leftarrow \mathrm{LAP}(Q^{\mathrm{row}})$
\ENDFOR
\STATE Update $(\pi^*, \sigma^*)$ if $F(\pi,\sigma)$ improved; $\hat\pi \leftarrow$ best $\tau\in\{\hat\pi,\pi_0,\pi,\sigma\}$ by $d_M(A;\tau,\tau)$
\ENDFOR
\STATE \textbf{for} $\sigma=\pi$: \textsc{Exchange}: apply pairwise swaps to $\hat\pi$ while they lower $d_M(A;\hat\pi,\hat\pi)$
\RETURN best $(\pi^*, \sigma^*)$, or $\hat\pi$
\end{algorithmic}
\end{algorithm}

\begin{proposition}[Monotone convergence]
\label{prop:convergence}
Each LAP sub-step is a global minimizer of the one-sided objective
(Lemma~\ref{lem:col_sep}), so the iterates satisfy
$F(\pi^{(t+1)},\sigma^{(t+1)}) \le F(\pi^{(t)},\sigma^{(t)})$ with
$F(\pi,\sigma) = \|(I{-}\Pi_M)(P_\pi A P_\sigma^\top)\|_F^2$, and
the $F$-values stabilize after finitely many steps; if ties in the LAP
are broken in favour of the current permutation, the iterates themselves
stop at a coordinate-wise local minimum. For a common permutation the
exchange phase strictly decreases $d_M(A;\pi,\pi)$ and stops at an
order that no pairwise exchange improves.
\end{proposition}
\paragraph{Common permutation.}
For $\sigma=\pi$ the two-sided iterate is rounded to one permutation
and finished by an exchange descent on $d_M(A;\pi,\pi)$: the change of
an exchange has a closed form, two matrix products score all $n^2$
exchanges, and improving ones are applied until none remains
(Appendix~\ref{app:init}). The LAP steps move whole rows and columns
and stall on the ties a $\{0,1\}$ input induces in $Q$; the exchanges
move one pair at a time and finish the descent
(Appendix~\ref{app:ablation}).

\paragraph{Variants.}
The lemma uses only that the cost is a sum of per-entry weights, so
Algorithm~\ref{alg:alternating_lap} runs on any separable score
$\sum_{ij}(1-M_{ij})\,W_{\pi^{-1}(i),\sigma^{-1}(j)}$: $W=A\odot A$ is
the pattern distance; $W=A$ (\textsc{AltLAP}-lin) is the likelihood
criterion for a mean shift rather than an energy contrast
(Section~\ref{sec:exp_blocks}).

% =============================================================================
\section{Pattern + Low-Rank Decomposition}
\label{sec:patlr}

\paragraph{Formulation.}
The \textbf{PatLR decomposition} approximates $P_\pi A P_\sigma^\top$
by a pattern-supported component $S$ ($S_{ij}=0$ when $M_{ij}=0$) and
a low-rank component $UV^\top$ of rank $\le k$. Writing
$X = P_\pi^\top UV^\top P_\sigma$ for the low-rank part in the original
coordinates, the objective is
\begin{equation}
  \inf_{\pi,\,\sigma,\, S,\, X}
  \norm{P_\pi(A-X)P_\sigma^T - S}^2
  \quad \text{s.t.} \quad S \odot (1-M) = 0,\;\; \rank(X) \le k.
  \label{eq:patlr_residual}
\end{equation}
For fixed $(\pi,\sigma,X)$ the optimal $S$ is the mask projection of
the permuted residual, and substituting it back gives
\begin{equation}
  \min_{S:\, S\odot(1-M)=0}\;\norm{P_\pi(A-X)P_\sigma^T - S}^2
  \;=\; d_M(A-X;\pi,\sigma),
  \label{eq:patlr_dM}
\end{equation}
so PatLR looks for a low-rank component after whose removal the residual
fits the pattern. Algorithm~\ref{alg:patlr} alternates the two
closed-form updates, $S$ by mask projection and $X$ by the rank-$k$ SVD
of $A$ minus the pattern part, with a \emph{safeguarded} permutation
update: $\textsc{AltLAP}(A-X, M, T_{\mathrm{LAP}};\,(\pi,\sigma))$ on
the residual from the current pair is accepted only if it lowers
$d_M(A-X;\cdot)$, and $S$ is then re-synchronized. For a common
permutation the update takes the common-permutation output of
Algorithm~\ref{alg:alternating_lap}, sets $\sigma=\pi$ and compares
$d_M(A-X;\pi,\pi)$; Proposition~\ref{prop:patlr_monotone} is unchanged.
The infimum need not be attained and the split between $S$ and $X$ need
not be unique: PatLR is a structured approximation model, scored by its
reconstruction. Since
$\pi=\sigma=\mathrm{id}$ with
$S=0$ is feasible, the infimum never exceeds the rank-$k$ SVD error,
and the example of Section~\ref{sec:prelim} shows that the gap can be
large.

\begin{algorithm}[tbp]
\caption{\textsc{PatLR}: Safeguarded Pattern + Low-Rank Decomposition}
\label{alg:patlr}
\begin{algorithmic}[1]
\REQUIRE $A \in \R^{n\times n}$, mask $M$, rank $k$, outer iterations $T$, LAP iterations $T_{\mathrm{LAP}}$
\STATE $(\pi,\sigma) \leftarrow \textsc{AltLAP}(A, M, T_{\mathrm{LAP}})$; \quad $X \leftarrow 0$; \quad $S \leftarrow \Pi_M(P_\pi A P_\sigma^\top)$
\FOR{$t = 1, \ldots, T$}
  \STATE $B \leftarrow P_\pi(A{-}X)P_\sigma^\top$; \quad $S \leftarrow \Pi_M(B)$
  \STATE $X \leftarrow \mathrm{SVD}_k(A - P_\pi^\top S P_\sigma)$
  \STATE $(\tilde\pi,\tilde\sigma) \leftarrow \textsc{AltLAP}(A - X,\; M,\; T_{\mathrm{LAP}};\, (\pi,\sigma))$
  \STATE \textbf{if}\ $d_M(A{-}X;\tilde\pi,\tilde\sigma) < d_M(A{-}X;\pi,\sigma)$ \ \textbf{then}\ $(\pi,\sigma) \leftarrow (\tilde\pi,\tilde\sigma)$
  \STATE $S \leftarrow \Pi_M\bigl(P_\pi(A{-}X)P_\sigma^\top\bigr)$ \quad\COMMENT{re-synchronize $S$ with the accepted $(\pi,\sigma)$}
\ENDFOR
\RETURN $(\pi,\sigma),\; S,\; X$
\end{algorithmic}
\end{algorithm}

\begin{proposition}[Monotone convergence of \textsc{PatLR}]
\label{prop:patlr_monotone}
For $F(\pi,\sigma,S,X) = \|P_\pi(A-X)P_\sigma^\top - S\|_F^2$ subject to
$S \odot (1{-}M) = 0$ and $\mathrm{rank}(X)\le k$, Algorithm~\ref{alg:patlr}
yields a non-increasing sequence of $F$-values, which therefore
converges (Appendix~\ref{app:patlr_proof}).
Because $X$ is continuous we do not claim finite termination or
convergence of the iterates $(S,X)$ themselves; the
algorithm stops at a tolerance $\varepsilon$ on the decrease of $F$
or a capped iteration count.
\end{proposition}

\subsection{Adaptive Pattern Selection}
\label{sec:adaptive}

Pattern $M$ is rarely known a priori. We select it jointly with the
permutations from a fixed vocabulary $\mathcal{V}=\{M_1,\dots,M_K\}$ by
the \emph{density-normalized score} (for $A\neq0$, $\rho_M<1$)
\begin{equation}
  \tilde{s}_M(A;\pi,\sigma) := \frac{d_M(A;\pi,\sigma)}
  {\|A\|_F^2 \cdot (1-\rho_M)^{3/2}}\,,
  \qquad
  \tilde{s}_M(A) := \min_{\pi,\sigma}\tilde{s}_M(A;\pi,\sigma).
  \label{eq:adaptive_score}
\end{equation}
The denominator is a reference against independent uniform row and
column permutations, which place a fraction $\rho_M$ of the energy
inside the mask in expectation (Section~\ref{sec:prelim}); its
superlinear exponent penalizes dense masks beyond this coverage
advantage. Under a common permutation the random-permutation reference
also depends on the split of the energy between the diagonal and the
off-diagonal entries (Appendix~\ref{app:basic}); the normalization is
used as stated in both settings. Without noise a mask and every mask
containing it score $0$
alike, so the identified geometry need not be unique
(Appendix~\ref{app:selection}). The default vocabulary has nine masks,
blocks $k\in\{2,3,4,6\}$, bands $w\in\{1,2,4\}$, arrowhead and
block+band ($k{=}3$, $w{=}2$); any binary mask can be added.
\paragraph{Adaptive PatLR.}
Pattern and permutation are searched jointly by simulated annealing over
$\mathcal{V}$ with Metropolis acceptance on $\tilde s_M$
(Algorithm~\ref{alg:adaptive}, Appendix~\ref{app:adaptive_search}). The
current $(\pi,\sigma)$ is a warm start for \emph{any} proposed pattern:
good permutations for similar patterns are correlated, so a candidate
$M_{j'}$ is evaluated with $\ell{=}3$ LAP steps instead of a full
\textsc{AltLAP} run (exchange phase and \textsc{PatLR} only on the
selected mask). After screening every mask with $\ell$ LAP steps
($O(K\ell n^3)$), the $N{=}80$ search iterations cost
$O(N(1{+}\ell)n^3)$ independently of $K$; in practice the selection
costs about one \textsc{PatLR} call (Appendix~\ref{app:scaling}).

% =============================================================================
\section{Experiments}
\label{sec:experiments}

Together, the experiments test whether one solver recovers different
geometric structures, whether these structures complement low-rank
approximation, and whether they can be selected and interpreted on real
data. Unless stated otherwise we use a common row/column permutation
($\sigma{=}\pi$). We report our objective where the planted order is
known and, where a task has its own literature, its metric alongside
the methods developed for it. Comparisons are paired over $20$ seeds with $95\%$ bootstrap
intervals; Figure~\ref{fig:showcase} illustrates the outputs; setups,
baselines and full grids are in Appendix~\ref{app:experiments}.

\begin{figure}[t]
  \centering
  \includegraphics[width=\linewidth]{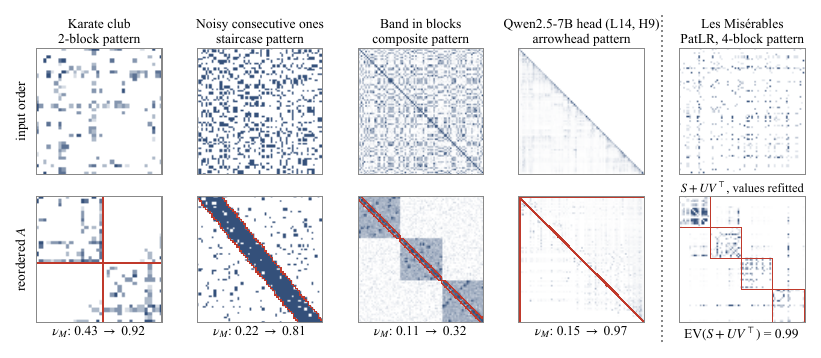}
  \caption{Outputs of the framework on one input of each kind (not a
    comparison). Top: the input in its given order; bottom: the order
    found by Alternating LAP for the named pattern (mask outlined in
    red), with the in-mask energy fraction $1-d_M(A;\pi)/\|A\|_F^2$
    before and after. The consecutive-ones column uses two permutations; the
    attention head is at $n{=}128$ on a square-root colour scale; the
    last column is the PatLR reconstruction $S + UV^\top$ (rank $3$) in
    the found order, with its explained variance.}
  \label{fig:showcase}
\end{figure}

\subsection{Ordering Structures}
\label{sec:exp_algo}
\label{sec:exp_ordering}

\textbf{Banded recovery (our objective).}
We first ask whether the generic solver optimizes the pattern distance
as well as a general QAP solver and classical ordering heuristics do.
On randomly permuted banded targets in a
Gaussian and a sparse-graph regime, Table~\ref{tab:baselines} reports
the error reduction (ER, the fraction of the input's off-pattern mass
removed by the found order) of Alternating LAP next to reverse
Cuthill--McKee~\citep{cuthill1969reducing} (RCM), Fiedler ordering plus
greedy 2-SUM swaps (F+greedy) and QAP-FAQ~\citep{vogelstein2015faq}.
Alternating LAP
gives the largest reduction on the Gaussian inputs at every $n$ and on
the sparse graphs at $w{=}5$; FAQ is ahead by $0.04$--$0.06$ on the
sparse $w{=}2$ graphs, where the ties induced by binary inputs stall
the LAP steps and the exchange phase carries the fit
(Appendix~\ref{app:ablation}). RCM and GPS reduce the bandwidth, a
criterion that saturates on noisy inputs, rather than the off-band mass
(Appendix~\ref{app:baselines}).

\begin{table}[t]
\caption{Error reduction (ER; higher = better) on banded patterns, $20$
  seeds. RCM (via KNN sparsification on dense inputs) and F+greedy
  (Fiedler order plus greedy 2-SUM swaps) optimize other criteria; FAQ
  (three starts) and AltLAP (five restarts) both minimize $d_M$.}
\label{tab:baselines}
\centering
\scriptsize
\renewcommand{\arraystretch}{0.92}
\setlength{\tabcolsep}{2.6pt}
\begin{tabular}{r cccc cccc cccc cccc}
\toprule
& \multicolumn{4}{c}{\textbf{Gauss}\,$w{=}2$}
& \multicolumn{4}{c}{\textbf{Gauss}\,$w{=}5$}
& \multicolumn{4}{c}{\textbf{Sparse}\,$w{=}2$}
& \multicolumn{4}{c}{\textbf{Sparse}\,$w{=}5$} \\
\cmidrule(lr){2-5} \cmidrule(lr){6-9} \cmidrule(lr){10-13} \cmidrule(lr){14-17}
& \multicolumn{2}{c}{\scriptsize other} & \multicolumn{2}{c}{\scriptsize $d_M$}
& \multicolumn{2}{c}{\scriptsize other} & \multicolumn{2}{c}{\scriptsize $d_M$}
& \multicolumn{2}{c}{\scriptsize other} & \multicolumn{2}{c}{\scriptsize $d_M$}
& \multicolumn{2}{c}{\scriptsize other} & \multicolumn{2}{c}{\scriptsize $d_M$} \\
\cmidrule(lr){2-3}\cmidrule(lr){4-5}\cmidrule(lr){6-7}\cmidrule(lr){8-9}\cmidrule(lr){10-11}\cmidrule(lr){12-13}\cmidrule(lr){14-15}\cmidrule(lr){16-17}
$n$ & RCM & F+gr. & FAQ & \textbf{AltLAP}
    & RCM & F+gr. & FAQ & \textbf{AltLAP}
    & RCM & F+gr. & FAQ & \textbf{AltLAP}
    & RCM & F+gr. & FAQ & \textbf{AltLAP} \\
\midrule
30 & 0.03 & 0.84 & 0.60 & \textbf{0.87} & 0.07 & \textbf{0.98} & 0.84 & \textbf{0.98} & 0.23 & 0.45 & \textbf{0.73} & 0.67 & 0.41 & 0.78 & 0.88 & \textbf{0.91} \\
100 & 0.05 & 0.37 & 0.51 & \textbf{0.69} & 0.09 & 0.72 & 0.66 & \textbf{0.87} & 0.06 & 0.09 & \textbf{0.49} & 0.45 & 0.16 & 0.29 & 0.66 & \textbf{0.69} \\
200 & 0.03 & 0.14 & 0.46 & \textbf{0.61} & 0.08 & 0.33 & 0.61 & \textbf{0.75} & 0.02 & 0.03 & \textbf{0.36} & 0.30 & 0.09 & 0.10 & 0.52 & \textbf{0.53} \\
500 & 0.03 & 0.03 & 0.40 & \textbf{0.52} & 0.06 & 0.08 & 0.57 & \textbf{0.61} & 0.00 & 0.00 & \textbf{0.20} & 0.15 & 0.03 & 0.02 & \textbf{0.33} & \textbf{0.34} \\
\bottomrule
\end{tabular}
\end{table}

\begin{figure}[t]
  \centering
  \includegraphics[width=\linewidth]{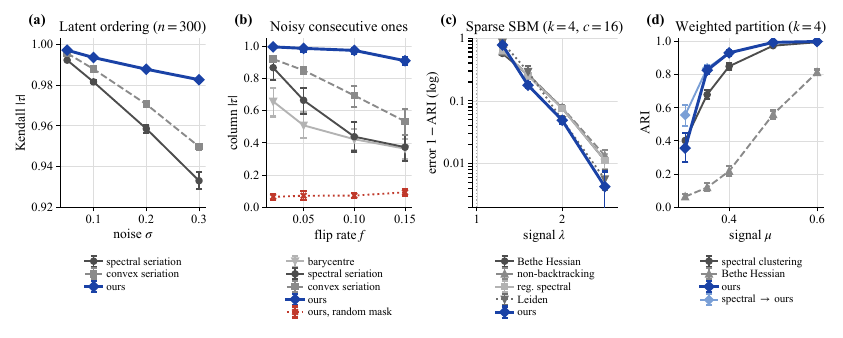}
  \caption{Four classical tasks scored by their own metrics (means,
    $95\%$ bootstrap intervals, $20$ seeds). ``Ours'' is Alternating LAP
    on the mask of the task: (a)~band of width
    $25$ from the spectral order; latent ordering, $n{=}300$, correlation
    length $25$ ($y$-axis from $0.92$). (b)~staircase of interval
    length $12$ from random restarts, with the same run on a random mask
    as control; noisy consecutive ones, $60\times60$. (c)~equal-size block
    mask; sparse SBM, $k{=}4$, mean degree $16$,
    drawn as the error $1-\mathrm{ARI}$ on a log axis (lower is better;
    dotted: detectability threshold). (d)~block mask on the linear score, from random restarts
    and from the spectral partition; weighted planted partition, $k{=}4$.}
  \label{fig:tasks}
\end{figure}

\textbf{Latent ordering (Figure~\ref{fig:tasks}a).}
A lower pattern distance need not mean a more accurate order, so the
remaining tasks are scored by their own metrics. Planted Robinson
matrices, scrambled and noised, scored by Kendall
$|\tau|$ against the planted order, next to spectral
seriation~\citep{atkins1998spectral} and the convex relaxation of
\citet{fogel2013seriation}. Over $24$ configurations one Alternating
LAP run from the spectral order, with the band width set to the
generator's correlation length, improves on spectral seriation in $18$
cells and matches or exceeds the convex relaxation in $19$, the gain
growing with noise and with $n$; the exceptions are the cells at
$\ell{=}25$, $n{=}100$, where every method is near $\tau{=}1$, and one
cell at $\ell{=}6$, $n{=}100$ where the relaxation leads by $0.04$
(Appendix~Table~\ref{tab:seriation_full}).

\textbf{Consecutive ones (Figure~\ref{fig:tasks}b).}
The consecutive-ones property (every row an interval after a column
order~\citep{booth1976c1p,kendall1971seriation}) is the staircase band in a
rectangular matrix, fitted with independent row and column permutations
(Appendix~\ref{app:notation}).
On planted instances with flip noise, Alternating LAP from random
restarts recovers the column order with a higher $\tau$ and fewer edits
than convex seriation on every seed at every noise level, while a random
or a block mask of the same density recovers almost nothing: the gain is
the geometry of the mask.

Circular seriation and the sensitivity of these fits to the mask width
and to the generator are in Appendix~\ref{app:specialists}, as are
nested matrices, a boundary case: where the marginals already determine
the order, the general optimizer does not improve over the specialized
one.

\subsection{Block Structures}
\label{sec:exp_specialists}
\label{sec:exp_blocks}

We turn from order to grouping, where the block-mask objective is the
likelihood criterion of the planted models; on the weighted matrices
the solver runs on the linear score (AltLAP-lin, Section~\ref{sec:qap}).

\textbf{Sparse and weighted block models (Figure~\ref{fig:tasks}c,d).}
On the sparse SBM at the detectability scale ($\lambda{=}1$ the
threshold~\citep{decelle2011asymptotic}) the block-mask objective is the
maximum-likelihood criterion~\citep{abbe2016exact}, and two regimes
appear: below $\lambda\approx1.5$ a better criterion value does not give
a better partition (Alternating LAP is below the Bethe Hessian in ARI,
above it when started from its partition); above this regime
Alternating LAP is competitive with the specialized methods, with the
highest ARI in most cells, and at $c\ge16$ from $\lambda{=}1.6$ on it
matches or exceeds the spectral baselines (Appendix~Table~\ref{tab:sbm}).
On a weighted planted partition the
linear score is the likelihood criterion and AltLAP-lin is above spectral
clustering from $\mu{=}0.35$ on (Appendix~Table~\ref{tab:weighted}). The
spectral baselines receive the true $k$; our mask also assumes equal
block sizes, and under unequal planted sizes, sizes estimated from a
spectral partition recover most of the gap
(Appendix~Table~\ref{tab:block_sizes}).

\textbf{Composite target: a band inside blocks.}
The band \emph{inside} the blocks (Figure~\ref{fig:showcase}, third
column) tests the joint refinement of communities and their internal
order: two structural hypotheses combined in one mask and refined by
one criterion. On $16$ planted configurations we compare the natural
pipeline, spectral clustering then seriation
inside each cluster, with the same pipeline followed by one Alternating
LAP pass on the composite mask, with block boundaries re-estimated from
the order. The pass raises the error reduction on
the target in every configuration, from a median $45\%$ of the planted
order's value to $94\%$, keeps the blocks exact and improves the
within-block order where the band is wide enough to determine it
($w{=}3$); similar ER gains hold with convex seriation inside the
clusters. From a cold start the composite mask alone recovers the blocks
only in part: the pipeline supplies the start and the mask the target
(Appendix~Table~\ref{tab:composite_full}).

\textbf{Labelled networks.}
On four networks with published communities no single panel method is
best; PatLR with rank $1$ recovers the karate-club split exactly
(Appendix~Table~\ref{tab:realnets}).

\subsection{Pattern + Low-Rank Decomposition}
\label{sec:exp_patlr}

The experiments above test whether permutations reveal structure; we
now ask whether representing it explicitly improves reconstruction
beyond low rank. We measure \textbf{denoising quality} (DQ)
$= 1 - \|\hat{A} - A_\text{true}\|_F^2 / \|A_\text{true}\|_F^2$ of
reconstructions $\hat A$ from $A_\text{true} + E$ ($E$ symmetric
Gaussian), against SVD and Robust PCA, on planted matrices whose family
is known (PatLR receives its mask).

\begin{figure}[t]
  \centering
  \includegraphics[width=\linewidth]{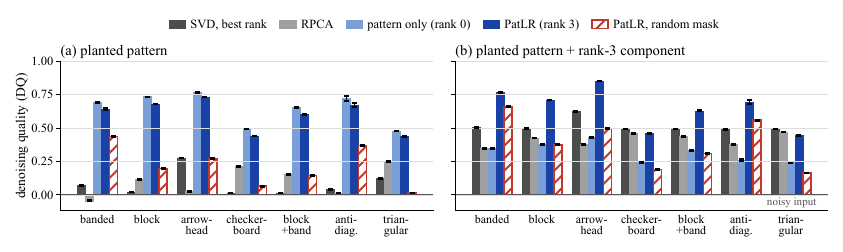}
  \caption{Denoising quality at SNR~$1$ ($n{=}200$, $20$ seeds, $95\%$
    bootstrap intervals; the noisy input has DQ $0$). (a)~Planted pattern.
    (b)~Planted pattern plus a rank-$3$ component of equal energy. SVD at
    the rank that maximizes DQ; RPCA; the pattern alone (PatLR at rank
    $0$); PatLR at rank $3$; and PatLR with a random mask of the same
    density as control.}
  \label{fig:denoise}
\end{figure}

\textbf{Planted structure at SNR~1 (Figure~\ref{fig:denoise}).}
Seven pattern families at $n{=}200$, each planted alone and together
with a rank-$3$ component of equal energy, scrambled, with noise at
$\mathrm{SNR} = \|A_\text{true}\|_F^2/\|E\|_F^2 = 1$, so that the noisy
input has DQ $0$. On the pattern alone the mask removes most of the
noise and the low-rank estimates little. On pattern plus low rank no
single component suffices, and PatLR reconstructs the combined signal
more accurately than either estimate alone on the masks of density
$\le0.25$. On the two half-dense masks every estimate that copies the
in-mask entries also copies their noise, and PatLR ends $0.03$--$0.04$
below SVD (bound in Appendix~\ref{app:capacity}).

\textbf{Two controls.} A random mask of the same density, solver and rank stays
$0.2$--$0.5$ DQ below the planted mask ($0.1$--$0.4$ with the rank-$3$
component), but above the SVD level on the band and anti-diagonal
masks. Part of the gain is therefore mass that a free permutation packs
into $|M|$ entries; the rest is the geometry. At matched nominal
parameter budgets PatLR has a higher mean DQ than SVD and spectral
co-clustering on all seven families
(Appendix~Table~\ref{tab:capacity_full}).

\textbf{Real data.}
Where no truth is planted, Figure~\ref{fig:showcase} shows the
decomposition of Les
Mis\'erables~\citep{knuth1993graphbase} under a four-block template;
Appendix~Figure~\ref{fig:patlr_real} follows it step by step on this
network, on the karate club~\citep{zachary1977information} and on
three feature-correlation matrices: the pattern part takes the
community blocks, the band and the feature groups, and the low-rank
part the hubs and the general correlation factor.

\subsection{Adaptive Pattern Selection}
\label{sec:exp_adaptive}

\begin{figure}[t]
  \centering
  \includegraphics[width=\linewidth]{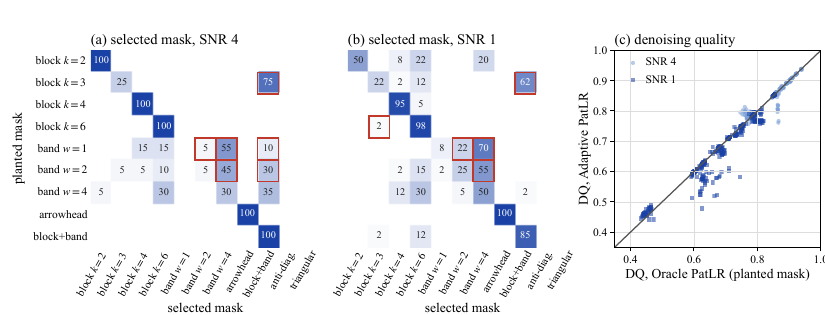}
  \caption{Adaptive pattern selection ($n{=}200$, $540$ instances).
    (a,\,b)~Mask selected by Adaptive PatLR (columns; the checkerboard
    is counted with block $k{=}2$) for each planted mask (rows) at
    SNR~$4$ and SNR~$1$, in \% of instances over both generators; red
    frames: the selected mask contains the planted one (block+band
    covers block $k{=}3$ up to two rows). (c)~DQ of Adaptive against
    Oracle PatLR per instance. Per-family numbers:
    Appendix Table~\ref{tab:adaptive_selection}.}
  \label{fig:adaptive_main}
\end{figure}

The experiments above give PatLR its mask; Algorithm~\ref{alg:adaptive}
selects it jointly with the permutation, and the experiment asks how
much reconstruction quality this costs and whether the planted geometry
is identified. We plant each of the nine
masks of the default vocabulary ($n{=}200$, with and without a rank-3
component, SNR $4$ and $1$; $540$ instances) and compare
Adaptive PatLR, run over these nine masks plus three gallery
distractors, with Oracle PatLR (the planted mask), brute force over the
same vocabulary and SVD at its best rank
(Figure~\ref{fig:adaptive_main}; Appendix~\ref{app:adaptive_oracle}).

\textbf{Quality (Figure~\ref{fig:adaptive_main}c).} Adaptive PatLR reaches the oracle's
denoising quality on essentially every instance (median ratio $1.00$,
$5$th percentile $0.92$), where SVD at its best rank averages less than
half of the oracle's value at SNR~$1$.

\textbf{Identification (Figure~\ref{fig:adaptive_main}a,b).} The outcome varies by family: arrowhead,
block+band and the block templates $k\ge4$ are recovered reliably,
whereas thin bands and block $k{=}3$ often select a broader or
alternative mask (a wider band, a block template or block+band) within
$0.07$ of the oracle's DQ; good reconstruction therefore does not imply
exact recovery of the geometry. The anti-diagonal and triangular
distractors are never selected.

\textbf{Brute force.} An exhaustive scan of the vocabulary with the LAP phase gives the same reconstruction
(within $0.01$) and similar selections at a cost growing with
$|\mathcal{V}|$; the search costs one PatLR call.

\subsection{Attention Heads}
\label{sec:exp_attention}

We use attention matrices to test whether pattern labels are stable
across prompts and context lengths, whether the reordering adds
information beyond the native token order, in which the literature
describes head types, and whether the layerwise profile of labels is
shared across three models.

\paragraph{Setup.}
We score every head of GPT-2, Qwen2.5-0.5B and \textbf{Qwen2.5-7B}
against a vocabulary of nine geometric masks (bands, blocks,
arrowhead, block+band, diagonal), applied to the causal attention
matrices without requiring the
reordered matrix to stay causal, with the density-normalized
score~\eqref{eq:adaptive_score}, on $20$ WikiText-2 passages at
$n\in\{128, 512\}$ tokens; the native token order is a candidate for
every mask, and the label of a (head, prompt) pair is the mask with the
lowest score (Appendix~\ref{app:attention_details}).

\begin{figure}[t]
  \centering
  \includegraphics[width=\linewidth]{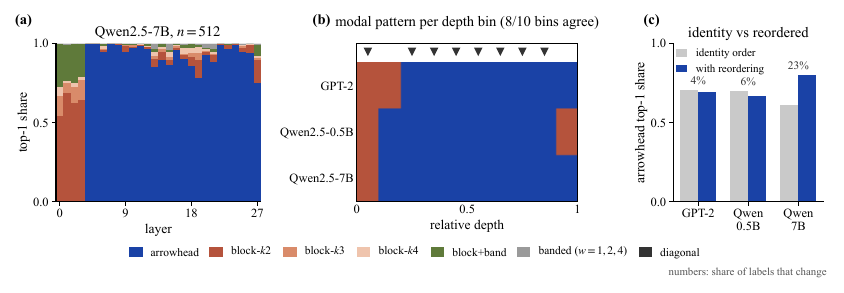}
  \caption{Attention heads at $n{=}512$, unified scoring. (a)~Qwen2.5-7B:
    share of the top-$1$ label per layer over $20$ prompts. (b)~Modal label
    per relative-depth bin for the three models; markers show the bins
    where all three agree. (c)~Arrowhead share under the native token
    order and with reordering, with the share of (head, prompt) labels
    that change (Appendix~\ref{app:identity}).}
  \label{fig:attention_main}
\end{figure}

\paragraph{Labels and layer organization.}
Arrowhead, the geometry of the attention
sink~\citep{xiao2023efficient,zhai2023stabilizing}, is the label of $69\%$ / $67\%$ / $80\%$ of the
(head, prompt) pairs of GPT-2 / Qwen2.5-0.5B / Qwen2.5-7B at $n{=}512$;
block masks take most of the rest, thin bands and the diagonal a few
percent. The label is a fit, not the best of a poor set: the selected
mask's score is about an order of magnitude below its
random-permutation level. By layer (Figure~\ref{fig:attention_main}a), the
first three to four layers of every model carry block labels and the
later layers arrowhead; on Qwen2.5-0.5B the last two layers return to
block-$k$2. Most block-labelled pairs are wide local heads in the
native token order, labelled by the block mask containing their window
(Appendix~Figure~\ref{fig:permuted_detail}).

\paragraph{Are the permutations needed?}
Scoring the same pairs in their native token order
(Figure~\ref{fig:attention_main}c) changes a few percent of the labels
on GPT-2 and Qwen2.5-0.5B; on Qwen2.5-7B the arrowhead share rises from
$0.61$ to $0.80$ and $23\%$ of the labels change. The reason is that on
a quarter of the passages the sink is not the first token: there the
native-order arrowhead share is below $2\%$, and the reordering
restores the label (Appendix~\ref{app:identity}). The reordering thus
buys a label independent of where the sink sits.

\paragraph{Stability.}
The mean agreement of a head's label with its modal label over the
$20$ prompts is $0.96$--$0.99$ for arrowhead heads and $0.64$--$0.74$
for the others. Between $n{=}128$ and $n{=}512$ the modal label of a
head is unchanged for more than nine heads in ten. On a shared $10$-bin
relative-depth grid the per-layer modal label agrees across the three
models in $8/10$ bins (Figure~\ref{fig:attention_main}b): blocks in the
first tenth and arrowhead thereafter
(Appendix~\ref{app:attention_experiments}).

% =============================================================================
\section{Discussion and Limitations}
\label{sec:discussion}

\paragraph{Limitations.}
The dense implementation costs $O(n^3)$ per LAP step and per exchange
round, which makes large-matrix and long-context analyses costly
(Appendix~\ref{app:scaling}). The permutation search is heuristic
and does not guarantee global optimality (Section~\ref{sec:qap}).
Adaptive selection operates within a supplied mask vocabulary: a
selected label is the best fit among the candidates, not an
identification of the generating geometry (Section~\ref{sec:exp_adaptive}). Our theoretical guarantees concern
optimization, while recovery under noise is evaluated empirically. The
attention study covers three decoder-only models at $128$ and $512$
tokens; other architectures and longer contexts remain future work.

% =============================================================================
\section{Conclusion}
\label{sec:conclusion}

We introduced \emph{permutation-structural simplicity} as an axis of
matrix structure complementary to low rank: the pattern distance relates
a dozen classical reordering problems to one objective
(Table~\ref{tab:patterns}), Alternating LAP is one solver for any
binary mask, PatLR adds a low-rank residual, Adaptive PatLR selects
the pattern from a vocabulary, and on attention heads the same score
gives stable labels with one depth profile across three models.

\begin{ack}
We thank the anonymous reviewers and the area chair for suggestions that
led to the specialized baselines, the capacity controls, and the
hardness statement.
This work received no third-party funding. The authors declare no
competing interests.
\end{ack}

% =============================================================================
\bibliographystyle{plainnat}
\bibliography{neurips}

\newpage

%%%%%%%%%%%%%%%%%%%%%%%%%%%%%%%%%%%%%%%%%%%%%%%%%%%%%%%%%%%%

\appendix
\newpage

% ===========================================================================
\section{Pattern Dictionary, Worked Example and Notation}
\label{app:example_notation}
% ===========================================================================

\subsection{Pattern dictionary}
\label{app:dictionary}

Table~\ref{tab:patterns} lists the pattern families of
Figure~\ref{fig:gallery} together with the classical problems they encode;
families used in our experiments are marked with a dagger, the others are
included to delineate the scope of the formulation and are not analysed
here.
Table~\ref{tab:correspondence} relates the masks to the established problems
and to the specialized methods used as baselines.

\begin{table}[htp]
\caption{Pattern dictionary: mask families, the structure they encode and
  the classical problem behind each ($\dagger$: used in our experiments).
  The last column records only what this paper establishes about
  computing $d_M$ (Theorem~\ref{thm:nphard} and exact special cases);
  dashes: not analysed here.}
\label{tab:patterns}
\centering
\scriptsize
\setlength{\tabcolsep}{3pt}
\begin{tabular}{l l >{\raggedright\arraybackslash}p{0.26\linewidth} l}
\toprule
Pattern $M$ & Shape & Classical problem & $d_M$ in this paper \\
\midrule
\multicolumn{4}{l}{\emph{Order and locality}} \\
Diagonal$^\dagger$ & $\delta_{ij}$ & Linear assignment (LAP) & one LAP$^{\S\S}$ \\
Banded ($w$)$^\dagger$ & $|i-j| \le w$ & Seriation, bandwidth min.~\citep{atkins1998spectral,cuthill1969reducing} & NP-hard at $w{=}1$, $\sigma{=}\pi$ (App.~\ref{app:hamiltonian}) \\
Anti-diagonal$^\dagger$ & $i+j=n+1$ & Reverse ordering (diagonal under column reversal) & one LAP$^{\S\S}$ \\
Cyclic band$^\dagger$ & $\min(|i-j|, n-|i-j|)\le w$ & Circular seriation~\citep{recanati2018circular} & -- \\
Upper triangular$^\dagger$ & $i \le j$ & Linear ordering / feedback arc set & --$^{\P}$ \\
2D grid stencil & $\|u-v\|_1\le 1$ on a grid & Mesh / PDE orderings~\citep{george1973nested} & -- \\
\midrule
\multicolumn{4}{l}{\emph{Grouping and coupling}} \\
Block-diagonal ($k$)$^\dagger$ & $k$ diagonal blocks & Community detection, SBM~\citep{holland1983stochastic} & NP-hard, Thm.~\ref{thm:nphard} \\
Block + band$^\dagger$ & blocks $\cup$ band & Modular networks with local structure & -- \\
Band-in-block$^\dagger$ & blocks $\cap$ band & Composite target (Sec.~\ref{sec:exp_blocks}) & -- \\
Bordered block-diagonal & blocks + coupling border & Decomposition for parallel LA / LP~\citep{aykanat2004bordered} & -- \\
Core--periphery & $J_{cc}, J_{cp}, J_{pc}$, $0_{pp}$ & Core--periphery structure~\citep{rombach2014core} & -- \\
Block-triangular & $k$ blocks, upper part & Dulmage--Mendelsohn decomposition~\citep{pothen1990dm} & --$^{\dagger\dagger}$ \\
\midrule
\multicolumn{4}{l}{\emph{Hubs, nesting, intervals, localization}} \\
Arrowhead$^\dagger$ & row/col $0$ + diagonal & Hub / sink structure & $O(n^2)$ under $\sigma{=}\pi$ (App.~\ref{app:special}) \\
Ferrers (nested)$^\dagger$ & $\mathbf{1}[j\le r_i]$, $r$ non-incr. & Nestedness~\citep{brualdi1999nested,berger2017nestedness} & --$^{\natural}$ \\
Interval envelope$^\dagger$ & $\mathbf{1}[\ell_i\le j\le r_i]$ & Consecutive ones, interval graphs~\citep{booth1976c1p,mcconnell2004c1p} & --$^{\natural}$ \\
Localized bicluster & $k\times\ell$ rectangle & Submatrix localization, biclustering~\citep{hartigan1972biclustering,hajek2018submatrix} & -- \\
\midrule
\multicolumn{4}{l}{\emph{Hierarchy, branching, self-similarity, parity}} \\
Nested dissection & recursive separators & Fill-reducing orderings~\citep{george1973nested} & -- \\
Clique tree & overlapping cliques on a tree & Chordal sparsity, GGMs~\citep{vandenberghe2015chordal} & -- \\
Kronecker support & $K^{\otimes q}$ & Kronecker graphs~\citep{leskovec2010kronecker} & -- \\
Checkerboard$^\dagger$ & $i+j$ even & Balanced bisection (the $2$-block mask under the even/odd relabeling) & as $2$ blocks$^{\ddagger}$ \\
Local + strided & band $\cup$ periodic columns & Sparse attention layouts~\citep{child2019sparse} & -- \\
\bottomrule
\multicolumn{4}{p{0.97\linewidth}}{\scriptsize
  $^{\S\S}$With two permutations any permutation-matrix mask is a single
  LAP (Lemma~\ref{lem:col_sep}); under $\sigma=\pi$ the diagonal objective
  is constant and the anti-diagonal is a maximum-weight matching. The
  anti-diagonal \emph{band} of Section~\ref{sec:exp_patlr} is the banded
  mask under column reversal.
  $^{\P}$Under $\sigma=\pi$ the off-pattern mass is the weight of the
  backward arcs, the linear-ordering objective (constant for symmetric
  $A$).
  $^{\dagger\dagger}$Dulmage--Mendelsohn computes the finest
  block-triangular form in polynomial time; fitting a prescribed block
  profile with given sizes is a different problem.
  $^{\natural}$Exact recognition of the noiseless structure is polynomial
  (\citealp{berger2017nestedness}; \citealp{booth1976c1p}); our experiments
  concern the noisy fixed-profile versions (Appendix~\ref{app:c1p}).
  $^\ddagger$The same-parity mask (ones where $i+j$ is even, including
  the diagonal) is the $2$-block mask under the even/odd relabeling, so
  $d_M$ coincides with the $2$-block case; the opposite-parity mask
  (zero diagonal) would correspond to maximum bisection instead and is
  not used.}
\end{tabular}
\end{table}

\begin{table}[htp]
\caption{Correspondence between pattern masks and established problems,
  with the specialized methods used as baselines in this paper.}
\label{tab:correspondence}
\centering
\footnotesize
\setlength{\tabcolsep}{4pt}
\begin{tabular}{p{0.21\linewidth} p{0.27\linewidth} p{0.34\linewidth} l}
\toprule
Mask / setting & Classical problem & Specialized methods compared & Where \\
\midrule
Fixed $M$, one permutation & Koopmans--Beckmann QAP & QAP-FAQ (as alternative solver) & \S\ref{sec:exp_algo} \\
Fixed $M$, two permutations & Bilinear assignment, graph matching & Birkhoff relaxations (discussed) & \S\ref{sec:qap} \\
Banded, exact support & Bandwidth / profile reduction & RCM, Gibbs--Poole--Stockmeyer & \S\ref{sec:exp_algo} \\
Banded / Robinson, noisy & Seriation, latent ordering & Spectral and convex seriation & \S\ref{sec:exp_ordering} \\
Staircase band (rectangular) & Consecutive-ones, interval seriation & Barycentre, spectral, convex seriation & \S\ref{sec:exp_ordering} \\
Cyclic band & Circular seriation & Circular spectral embedding & App.~\ref{app:c1p} \\
Block-diagonal, $\{0,1\}$ & Community detection, SBM & Bethe Hessian, non-backtracking, regularized spectral, Leiden & \S\ref{sec:exp_specialists} \\
Block-diagonal, weighted & Planted partition (likelihood) & Spectral clustering & \S\ref{sec:exp_specialists} \\
Band inside blocks & Composite (no dedicated method) & Cluster-then-order pipeline & \S\ref{sec:exp_blocks} \\
Any $M$ + low rank (PatLR) & Structured plus low-rank approximation & SVD at matched nominal budget, RPCA, spectral co-clustering & \S\ref{sec:exp_patlr} \\
\bottomrule
\end{tabular}
\end{table}

\subsection{A $6\times 6$ example}
\label{app:example}

Take $n=6$, the band mask $M_{ij}=1$ iff $|i-j|\le 1$, and the reference
matrix $B_{ij}=\max(0,\,5-2|i-j|)$. Scramble it by
$p=(0,2,4,1,3,5)$ (zero-based), i.e.\ $A_{ij}=B_{p_i,p_j}$:
\begin{center}
\footnotesize
\begin{tabular}{@{}c@{\qquad}c@{}}
$B$ (reference) & $A$ (scrambled) \\[2pt]
$\begin{pmatrix}
5&3&1&0&0&0\\ 3&5&3&1&0&0\\ 1&3&5&3&1&0\\ 0&1&3&5&3&1\\ 0&0&1&3&5&3\\ 0&0&0&1&3&5
\end{pmatrix}$ &
$\begin{pmatrix}
5&1&0&3&0&0\\ 1&5&1&3&3&0\\ 0&1&5&0&3&3\\ 3&3&0&5&1&0\\ 0&3&3&1&5&1\\ 0&0&3&0&1&5
\end{pmatrix}$
\end{tabular}
\end{center}
$B$ is concentrated near the diagonal but not exactly supported on the
mask: its eight entries at $|i-j|=2$ have magnitude one, so the optimum
is $d_M(A)=d_M(B)=8$, while the objective at the identity ordering is
$\|(1-M)\odot A\|_F^2 = 90$.
\emph{One step.} With rows fixed, placing column $c$ at position $q$
costs $Q_{q,c}=\sum_i (1-M_{iq})\,A_{ic}^2$, the off-pattern mass this
placement would create (Lemma~\ref{lem:col_sep}):
\begin{center}
\footnotesize
\begin{tabular}{@{}c@{\qquad}c@{}}
$M$ (band, $w=1$) & $Q$ (rows $=$ positions $q$, columns $=c$) \\[2pt]
$\begin{pmatrix}
1&1&0&0&0&0\\ 1&1&1&0&0&0\\ 0&1&1&1&0&0\\ 0&0&1&1&1&0\\ 0&0&0&1&1&1\\ 0&0&0&0&1&1
\end{pmatrix}$ &
$\begin{pmatrix}
9&19&43&26&36&35\\ 9&18&18&26&27&26\\ 25&10&18&10&26&26\\ 26&26&10&18&10&25\\ 26&27&26&18&18&9\\ 35&36&26&43&19&9
\end{pmatrix}$
\end{tabular}
\end{center}
An optimal assignment on $Q$ orders the columns $(0,1,3,2,4,5)$ at cost
$74$, exactly the off-pattern mass after the move; the row step is
symmetric.
\emph{Run.} The successive exact one-sided minimizations give
$90\to 74\to 40\to 8$, and both final orders equal
$(0,3,1,4,2,5)=p^{-1}$, so the reordered matrix is exactly $B$.
Each step is the global optimum of its subproblem, so the objective
never increases (Proposition~\ref{prop:convergence}); this does not
imply global convergence in general, since other initializations or
tied assignments can end at coordinatewise local minima, which is what
the spectral initialization and the restarts address.

\subsection{Notation and terminology}
\label{app:glossary}

\begin{table}[H]
\caption{Symbols and terms used throughout; standard notions (LAP, QAP,
  Fiedler vector, ARI, Kendall $\tau$) are used in their usual sense.}
\label{tab:glossary}
\centering
\footnotesize
\setlength{\tabcolsep}{4pt}
\begin{tabular}{l p{0.68\linewidth}}
\toprule
Symbol / term & Meaning \\
\midrule
$A^{\pi,\sigma} = P_\pi A P_\sigma^\top$ & rows reordered by $\pi$, columns by $\sigma$; $(A^{\pi,\sigma})_{ij} = A_{\pi^{-1}(i),\sigma^{-1}(j)}$, a permutation mapping an original index to its position \\
$\Pi_M(X) = M\odot X$ & masking operator; $(I-\Pi_M)(X)$ is the off-pattern part \\
$d_M(A;\pi,\sigma)$ & off-pattern squared Frobenius mass at fixed permutations~\eqref{eq:offmass} \\
$d_M(A)$ & its minimum over independent row and column permutations~\eqref{eq:pattern_dist} \\
$d_M^{\mathrm{com}}(A)$, $d_M(A;\pi)$ & minimum under a common permutation ($\sigma=\pi$); the value at a fixed $\pi$ \\
$\nu_M(A)$ & explained pattern variance $1-d_M(A)/\|A\|_F^2$ \\
$\rho_M = |M|_1/n^2$ & mask density ($|M|_1$ the number of ones) \\
$\tilde s_M(A)$ & density-normalized score $d_M(A)/\bigl(\|A\|_F^2(1-\rho_M)^{3/2}\bigr)$, equation~\eqref{eq:adaptive_score} \\
$S$, $X=P_\pi^\top UV^\top P_\sigma$ & PatLR pattern part (reordered frame) and low-rank part (original frame), Section~\ref{sec:patlr} \\
2-SUM & $\sum_{ij}|A_{ij}|(\pi(i)-\pi(j))^2$, the seriation objective whose spectral relaxation is the Fiedler order \\
ER & error reduction $1-d_M(A;\hat\pi)/d_M(A;\mathrm{id})$ of the found order over the input order \\
BE & band energy, the in-mask energy fraction after reordering \\
DQ & denoising quality $1-\|\hat A-A_{\mathrm{true}}\|_F^2/\|A_{\mathrm{true}}\|_F^2$, against the clean matrix \\
EV & explained variance $1-\|A-\hat A\|_F^2/\|A\|_F^2$, against the observed matrix \\
block+band ($k$, $w$) & \emph{union} of $k$ equal diagonal blocks and the band $|i-j|\le w$ (vocabulary, Table~\ref{tab:capacity_full}, attention) \\
band-in-block & \emph{intersection}, a band inside each block: the composite target of Section~\ref{sec:exp_blocks} and Appendix~\ref{app:composite} \\
anti-diagonal ($w$) & anti-diagonal band $|i+j-(n+1)|\le w$ ($w{=}10$ at $n{=}200$); Table~\ref{tab:patterns} lists $w{=}0$ \\
\textsc{AltLAP}-lin & Alternating LAP on the separable score with $W=A$, the linear criterion (Section~\ref{sec:qap}) \\
\bottomrule
\end{tabular}
\end{table}

% ===========================================================================
\section{Deferred Proofs}
\label{app:proofs}
% ===========================================================================

\subsection{Basic identities and scope of the formulation}
\label{app:notation}

We use the conventions $A^{\pi,\sigma} := P_\pi A P_\sigma^\top$,
$(A^{\pi,\sigma})_{ij} = A_{\pi^{-1}(i),\sigma^{-1}(j)}$;
$\Pi_M(X) := M\odot X$ is the masking operator and
$(I-\Pi_M)(X) = (1-M)\odot X$ extracts the off-pattern part. For
rectangular $A, M\in\R^{m\times n}$ the same definitions apply with
$\pi\in S_m$, $\sigma\in S_n$ and $\rho_M = |M|_1/(mn)$; Lemma~\ref{lem:col_sep}
and Algorithm~\ref{alg:alternating_lap} are unchanged, the two LAPs
having sizes $n$ and $m$.
The restriction $\sigma=\pi$ gives the common-permutation distance
$d_M^{\mathrm{com}}(A)=\min_\pi d_M(A;\pi)$ with
$d_M(A;\pi) := \|(I-\Pi_M)(A^\pi)\|_F^2$; it is natural for symmetric
$A=A^\top$, $M=M^\top$ and is also imposed on the asymmetric attention
matrices (Section~\ref{sec:exp_attention}).
Because $\|A^{\pi,\sigma}\|_F = \|A\|_F$, the partition of entries by
$M$ yields the in-pattern reformulation
$d_M(A)= \|A\|_F^2 - \max_{\pi,\sigma}\|\Pi_M(A^{\pi,\sigma})\|_F^2$
(Section~\ref{sec:prelim}), which we use freely in the proofs below.

\paragraph{The boundary of the formulation.}
Masks whose objective does not depend on the permutation (the full
mask; under $\sigma=\pi$ the diagonal and, for symmetric $A$, the
triangle) are collected in Appendix~\ref{app:special}.
Some classical notions of structure after reordering are not fixed
binary supports and therefore fall outside Definition~\ref{def:pattern_dist}
as stated: Monge and Robinson matrices are ordering constraints on the
\emph{values} of $A$ (a Robinson matrix is banded only approximately,
which is why Section~\ref{sec:exp_ordering} fits a band of the
correlation length to it); a nested hierarchy of blocks coincides, as a
binary mask, with its coarsest level, so the hierarchy is visible in a
support only through separators (nested dissection); and $N{:}M$
sparsity~\citep{pool2021nm} is a \emph{family} of masks whose allowed
positions differ between rows
and groups, fitted by alternating the permutation with a per-group
selection of the support rather than with one mask.

\subsection{Basic properties of $d_M$}
\label{app:basic}

\textbf{(i) Nonnegativity} follows from the Frobenius norm being nonneg\-ative.
Equality $\dM(A)=0$ iff some $(\pi,\sigma)$ places every off-pattern
entry of $A^{\pi,\sigma}$ at zero, i.e.\ $A^{\pi,\sigma}$ is $M$-supported.

\textbf{(ii) Upper bound.}
By the in-pattern reformulation of~\eqref{eq:pattern_dist},
$\dM(A) = \|A\|_F^2 - \max_{\pi,\sigma}\|\PiM(A^{\pi,\sigma})\|_F^2 \le \|A\|_F^2$.
For the tighter bound, average over $(\pi,\sigma) \in \Sn \times \Sn$:
each entry $A_{kl}^2$ appears at each position $(i,j)$ with equal probability,
so $\E_{\pi,\sigma}[\|\PiM(A^{\pi,\sigma})\|_F^2] = \rho_M \|A\|_F^2$.
Hence $\max_{\pi,\sigma} \|\PiM(A^{\pi,\sigma})\|_F^2 \ge \rho_M\|A\|_F^2$, giving
$\dM(A) \le (1-\rho_M)\|A\|_F^2$.
The averaging argument uses independent $\pi$ and $\sigma$. Under the
restriction $\sigma=\pi$ a common permutation never moves a diagonal
entry off the diagonal, so with $D_A=\sum_i A_{ii}^2$,
$O_A=\sum_{i\ne j}A_{ij}^2$ and $m_{\mathrm{diag}}$, $m_{\mathrm{off}}$
the numbers of diagonal and off-diagonal positions of $M$,
$\E_\pi\|\PiM(A^{\pi,\pi})\|_F^2 = \tfrac{m_{\mathrm{diag}}}{n}D_A +
\tfrac{m_{\mathrm{off}}}{n(n-1)}O_A$, and the bound
$(1-\rho_M)\|A\|_F^2$ can fail: for $A=\bigl(\begin{smallmatrix}0&1\\1&0\end{smallmatrix}\bigr)$
and $M=I_2$, $\min_\pi d_M(A;\pi,\pi)=2>(1-\rho_M)\|A\|_F^2=1$.
The same remark applies to the expectation $\rho_M\|A\|_F^2$ used to
motivate the density normalization in Section~\ref{sec:adaptive}. More
generally the common-permutation distance is a restriction of the
feasible set: it is never below $d_M(A)$ and can exceed it even for
symmetric $A$ and $M$ (the same example has $d_M(A)=0$ with two
permutations and $2$ under one), so statements proved for independent
permutations transfer to the case $\sigma=\pi$ only where this is said.

\textbf{(iii) Permutation invariance.}
$\dM(P_\alpha A P_\beta^T)
= \min_{\pi,\sigma} \|(I{-}\PiM)(P_\pi P_\alpha A P_\beta^T P_\sigma^T)\|_F^2
= \min_{\pi,\sigma} \|(I{-}\PiM)(P_{\pi\alpha}\, A\, P_{\sigma\beta}^T)\|_F^2
= \min_{\pi',\sigma'} d_M(A;\pi',\sigma') = \dM(A)$,
reparametrising $\pi' = \pi\alpha$, $\sigma' = \sigma\beta$. \hfill$\square$

\subsection{Proof of Lemma~\ref{lem:col_sep} (Column-separability)}

Expand the off-pattern mass with rows fixed at $\pi_r$ and columns at $\sigma$:
\begin{align}
  &\sum_{i,j} (1-M_{ij})\,(P_{\pi_r} A P_\sigma^T)_{ij}^2 \notag\\
  &= \sum_{i,j} (1-M_{ij})\, A_{\pi_r^{-1}(i),\,\sigma^{-1}(j)}^2
  = \sum_{i,j} (1-M_{ij})\, B_{i,\,\sigma^{-1}(j)}^2.
  \label{eq:expand_app}
\end{align}
Substituting $p = j$ (target position) and $c = \sigma^{-1}(j)$
(original column placed at position $j$):
\begin{align}
  \eqref{eq:expand_app}
  &= \sum_{p=1}^{n} \sum_{i=1}^{n} (1-M_{ip})\, B_{i,\,\sigma^{-1}(p)}^2
  = \sum_{c=1}^{n} \sum_{i=1}^{n} (1-M_{i,\sigma(c)})\, B_{i,c}^2
  = \sum_{c=1}^{n} Q_{\sigma(c),\, c}\,,
\end{align}
using $Q_{p,c} = \sum_i (1-M_{ip}) B_{ic}^2$.
The expression $\sum_c Q_{\sigma(c), c}$ over bijections
$\sigma\colon [n] \to [n]$ is exactly a Linear Assignment Problem.
The row case follows identically with rows and columns exchanged. \hfill$\square$

\subsection{Proof of Proposition~\ref{prop:convergence} (Monotone convergence)}

Define $F(\pi_r, \sigma) := \fnorm{(I-\PiM)(P_{\pi_r} A P_\sigma^T)}^2$.
At each half-step, one permutation is fixed and the other is set to the
\emph{global} minimizer of the resulting LAP (Lemma~\ref{lem:col_sep}):
\[
  \sigma^{(t+1)} = \arg\min_{\sigma} F(\pi_r^{(t)}, \sigma)
  \;\;\Rightarrow\;\;
  F(\pi_r^{(t)}, \sigma^{(t+1)}) \le F(\pi_r^{(t)}, \sigma^{(t)}).
\]
Similarly, $\pi_r^{(t+1)} = \arg\min_{\pi_r} F(\pi_r, \sigma^{(t+1)})$
gives $F(\pi_r^{(t+1)}, \sigma^{(t+1)}) \le F(\pi_r^{(t)}, \sigma^{(t+1)})$.
Each full iteration reduces (or maintains) $F$. Since $F$ takes values in the
finite set $\{F(\pi,\sigma) : \pi,\sigma \in \Sn\}$,
the sequence of values is eventually constant. Stabilization of the
values alone does not stop the iterates: when the LAP has several
optimal assignments the permutations can keep moving on a plateau of
constant $F$. If ties are broken by keeping the current permutation
whenever it already attains the LAP optimum, every change of a
permutation strictly decreases $F$, so only finitely many changes occur
and the final pair is a coordinate-wise local minimum (neither one-sided
LAP can improve it). \hfill$\square$

\subsection{Proof of Theorem~\ref{thm:nphard} (NP-hardness)}

Throughout, $G=(V,E)$ is a simple graph with $|V|=n$ and $|E|=m$, and
$A_G$ its $\{0,1\}$ adjacency matrix. Restricting the feasible set to
$\sigma=\pi$ changes the problem, so hardness of the symmetric problem
does not by itself imply hardness of the two-permutation problem; the
reduction treats both cases. For a pair $(\pi,\sigma)$
the entry $A_{uv}$ is placed at position $(\pi(u),\sigma(v))$.

We reduce from \textsc{Minimum Bisection}~\citep{garey1979computers}:
given $G$ with $n$ even, find $|S|=n/2$ minimizing $\mathrm{cut}(S,\bar S)$.
Let $M$ be the $2$-block mask with blocks of size $n/2$ and set
\[
A \;=\; A_G + (m+1)\,I .
\]
For $(\pi,\sigma)$ let $S$ be the set of vertices that $\pi$ places in
the first $n/2$ row positions and $T$ the set that $\sigma$ places in the
first $n/2$ column positions. An entry is off-pattern iff its row and
column positions lie in different blocks, so
\[
d_M(A;\pi,\sigma)
\;=\; (m+1)^2\,|S\,\triangle\,T|
\;+\; \sum_{u\in S,\,v\notin T} (A_G)_{uv}
\;+\; \sum_{u\notin S,\,v\in T} (A_G)_{uv} .
\]
The two edge sums together are at most $2m$. If $S\neq T$ then, because
$|S|=|T|$, $|S\triangle T|\ge 2$ and $d_M(A;\pi,\sigma)\ge 2(m+1)^2 > 2m$,
whereas any pair with $S=T$ costs exactly $2\,\mathrm{cut}(S,\bar S)\le 2m$.
Hence every minimizer has $S=T$ and
$d_M(A) = 2\min_{|S|=n/2}\mathrm{cut}(S,\bar S)$. The same identity holds
under $\sigma=\pi$ (where $S=T$ automatically), which is the original
symmetric reduction. \hfill$\square$

\paragraph{The band of width $1$ under $\sigma=\pi$.}
\label{app:hamiltonian}
Let $A$ be symmetric and $M$ the banded mask with $w=1$. Under
$\sigma=\pi$ the in-band mass is
\[
\sum_{|i-j|\le 1} A^2_{\pi^{-1}(i),\pi^{-1}(j)}
\;=\; \sum_u A_{uu}^2 \;+\; 2\sum_{i=1}^{n-1} A^2_{\pi^{-1}(i),\pi^{-1}(i+1)} ,
\]
so by the in-pattern reformulation (Section~\ref{sec:prelim}) minimizing $d_M$ is equivalent to
maximizing $\sum_i A^2_{\pi^{-1}(i),\pi^{-1}(i+1)}$ over orderings, i.e.\
finding a maximum-weight Hamiltonian path in the complete graph with
edge weights $A_{uv}^2$. For $A=A_G$ this sum equals $n-1$ iff $G$ has a
Hamiltonian path, so even $\{0,1\}$ weights suffice and computing the
value of $d_M$ is NP-hard~\citep{garey1979computers}. Deciding whether
$d_M=0$ (bandwidth $1$) is polynomial~\citep{saxe1980bandwidth}; the
hardness concerns the value of $d_M$ when it is positive, and we do not
claim it for independent permutations. \hfill$\square$

\subsection{Permutation ambiguities: automorphisms of a mask}
\label{app:ambiguity}

In words: a planted instance has zero error at the planted
permutations; any other zero-error pair must place the whole support of
the signal inside the mask; and when the signal fills the mask, such a
pair can only be a symmetry of the mask. Formally, let
$\Aut(M) := \{(\alpha,\beta)\in\Sn^2 : M_{\alpha(i),\beta(j)} =
M_{ij}\;\forall i,j\}$ be the joint automorphism group of $M$ (the pairs
of row and column permutations that map the mask onto itself), and
$\Aut^{\mathrm{sym}}(M) := \{\sigma\in\Sn : M_{\sigma(i),\sigma(j)} =
M_{ij}\}$ its single-permutation version for $M=M^\top$.

\begin{proposition}[Zero-error permutations]
\label{prop:ambiguity}
Let $S_0$ be $M$-supported and $A = P_{\pi^*}^\top S_0\, P_{\sigma^*}$.
Then $d_M(A)=0$, attained at $(\pi^*,\sigma^*)$; every pair with
$d_M(A;\hat\pi,\hat\sigma)=0$ has relative permutations
$(\alpha,\beta) = (\hat\pi\circ(\pi^*)^{-1}, \hat\sigma\circ(\sigma^*)^{-1})$,
the found pair expressed relative to the planted one, for which
$P_\alpha S_0 P_\beta^\top$ is $M$-supported; and if
$\supp(S_0)=\supp(M)$, then $(\alpha,\beta)\in\Aut(M)$. Under
$\sigma=\pi$ the same holds for instances planted by one permutation,
$A = P_{\pi^*}^\top S_0 P_{\pi^*}$, with $\Aut^{\mathrm{sym}}(M)$ in
place of $\Aut(M)$ (an instance planted by two different permutations
need not have $d_M(A;\pi)=0$ for any $\pi$).
\end{proposition}
\begin{proof}
$d_M(A;\pi^*,\sigma^*) = \fnorm{(I-\PiM)(S_0)}^2 = 0$ by the support
condition. If $d_M(A;\hat\pi,\hat\sigma)=0$ then
$P_{\hat\pi} A P_{\hat\sigma}^\top = P_\alpha S_0 P_\beta^\top$ is
$M$-supported. If moreover $\supp(S_0)=\supp(M)$, then
$\supp(P_\alpha S_0 P_\beta^\top) = \{(\alpha(i),\beta(j)) : M_{ij}=1\}
\subseteq \supp(M)$, and since permutations preserve cardinality the
containment is an equality, i.e.\ $M_{\alpha(i),\beta(j)}=M_{ij}$ for
all $i,j$. The symmetric case is the same argument with
$\alpha=\beta$. Note that $\Aut(M)$ is the joint row\,$\times$\,column
action, in general strictly larger than the product of the row and
column stabilizers: for $M=I_n$ it is the diagonal copy of $\Sn$, while
the only purely row-preserving permutation is the identity.
\end{proof}

\begin{proposition}[Automorphism groups of symmetric patterns]
\label{cor:aut}
In the symmetric case ($\sigma = \pi$, $M = M^\top$) the ambiguities of
the three main families are the following.
\begin{enumerate}[label=\textnormal{(\roman*)}]
\item \textbf{Banded mask} ($M_{ij} = \mathbf{1}[|i{-}j| \le w]$, $w\ge1$,
  $n > 2w{+}1$): the only symmetries are the identity and the reversal
  $i\mapsto n{+}1{-}i$, so an order is recoverable up to global
  reversal; $\Aut^{\mathrm{sym}}(M) = \{\mathrm{id},\, \mathrm{rev}\}$.
\item \textbf{$k$-block mask} (equal blocks of size $b = n/k$): the
  symmetries are arbitrary permutations of the $k$ blocks combined with
  arbitrary permutations within each block, so the partition into
  blocks is recoverable and the order within a block is not;
  $\Aut^{\mathrm{sym}}(M) \cong (S_b)^k \rtimes S_k$, the wreath product
  $S_b \wr S_k$.
\item \textbf{Arrowhead mask} ($n\ge3$): every permutation fixing the
  hub is a symmetry and no other is, so only the hub is identifiable;
  $\Aut^{\mathrm{sym}}(M) \cong S_{n-1}$.
\end{enumerate}
These ambiguities fix what recovery can mean, and the experiments score
accordingly: an order by $|\tau|$, i.e.\ up to reversal
(Section~\ref{sec:exp_ordering}), a partition by ARI
(Section~\ref{sec:exp_blocks}), and an arrowhead fit as the
identification of a hub.
\end{proposition}
\begin{proof}
For~(i): in the simple graph on $\{1,\dots,n\}$ defined by the
off-diagonal support of $M$ ($i\sim j$ iff $1\le|i-j|\le w$; the
diagonal is preserved by every permutation), the vertices of minimum
degree $w$ are exactly $1$ and $n$ (for $w\ge1$ and $n>2w{+}1$ every
interior vertex has degree $\ge w{+}1$), so an automorphism fixes or
swaps them; composing
with $\mathrm{rev}$ we may assume $\sigma(1)=1$. The neighbours
$2,\dots,w{+}1$ of vertex $1$ have the distinct degrees $w{+}1,\dots,2w$,
so each is fixed, and for $j=2,\dots,n{-}w$ the closed neighbourhood of
the fixed vertex $j$ contains exactly one vertex not yet fixed, $j{+}w$,
which is therefore fixed as well. Hence $\sigma=\mathrm{id}$. (At $w=0$
the mask is $I_n$ and every permutation is an automorphism.)
For~(ii): $M_{ij}=1$ iff $\lceil i/b \rceil = \lceil j/b \rceil$, so the
off-diagonal support is a disjoint union of $k$ cliques of size $b$; an
automorphism maps cliques to cliques, hence permutes the blocks and acts
by an arbitrary permutation within each.
For~(iii): any $\sigma$ with $\sigma(1)=1$ preserves $M$, and for
$n\ge3$ the hub is the only vertex adjacent to all others, so every
automorphism fixes it (at $n=2$ the arrowhead is the full mask).
\end{proof}

% ===========================================================================
\section{Further Properties of the Framework}
\label{app:theory}
% ===========================================================================

\subsection{PatLR updates and proof of monotone convergence}
\label{app:patlr_proof}

The infimum in~\eqref{eq:patlr_residual} need not be attained: the set
of sums $S+UV^\top$ is not closed, as for low-rank plus sparse
models~\citep{tanner2019rigidity}; the algorithm uses only the
conditional minimizers in $S$ and $X$, which exist, and a safeguarded
proposal in the permutations. Nor is the split unique: for an in-mask
position $(i,j)$ and $H = e_ie_j^\top$, every pair $S = tH$,
$X = (1-t)H$ reconstructs $A = H$ exactly, so without further
constraints the components are a decomposition of the fit, not
identified parts, and the experiments score the reconstruction
$S + UV^\top$. For fixed permutations, both inner updates are exact.
Writing
\[
F(\pi,\sigma,S,X) := \norm{P_\pi(A-X)P_\sigma^T - S}^2
\]
with $S \odot (1-M)=0$ and $\rank(X)\le k$:
\begin{enumerate}[label=\textnormal{(\roman*)}]
\item the pattern step is the projection
  $S^* = \Pi_M(P_\pi(A-X)P_\sigma^T)$;
\item the low-rank step is the rank-$k$ truncation
  $X^* = \mathrm{SVD}_k(A - P_\pi^T S P_\sigma)$ by Eckart--Young
  (using $\fnorm{P_\pi(A-X)P_\sigma^\top - S}^2 = \fnorm{(A-X) - P_\pi^\top S P_\sigma}^2$).
\end{enumerate}

\paragraph{Useful identity.}
For any $(\pi,\sigma,X)$ the optimal $S$ given the rest is $S^* = \Pi_M(P_\pi(A-X)P_\sigma^\top)$, and substituting back yields
equation~\eqref{eq:patlr_dM} of Section~\ref{sec:patlr}, restated here for
the proof:
\[
\min_{S:\; S\odot(1-M)=0} F(\pi,\sigma,S,X)
= \fnorm{(I-\Pi_M)(P_\pi(A-X)P_\sigma^\top)}^2
= d_M(A-X;\pi,\sigma).
\]

\paragraph{Proof of Proposition~\ref{prop:patlr_monotone}.}
Index iterations by $t$ and write the state at the start of iteration $t$ as
$(\pi_t,\sigma_t,S_t,X_t)$. The body of the loop performs three updates.
\begin{enumerate}[label=\textnormal{(\arabic*)}]
\item \emph{Pattern step.} $S_{t+1} = \Pi_M(P_{\pi_t}(A-X_t)P_{\sigma_t}^\top)$ is
the exact minimizer of $F$ over $M$-supported $S$, so
$F(\pi_t,\sigma_t,S_{t+1},X_t) \le F(\pi_t,\sigma_t,S_t,X_t)$.
\item \emph{Low-rank step.} $X_{t+1} = \mathrm{SVD}_k(A - P_{\pi_t}^\top S_{t+1} P_{\sigma_t})$
is the global minimizer over rank-$k$ $X$ (Eckart--Young), so
$F(\pi_t,\sigma_t,S_{t+1},X_{t+1}) \le F(\pi_t,\sigma_t,S_{t+1},X_t)$.
\item \emph{Safeguarded permutation step, followed by re-synchronization of $S$.}
A candidate $(\tilde\pi,\tilde\sigma)$ is
produced by \textsc{AltLAP} on $A-X_{t+1}$ and accepted only if
$d_M(A-X_{t+1};\tilde\pi,\tilde\sigma) < d_M(A-X_{t+1};\pi_t,\sigma_t)$.
By~\eqref{eq:patlr_dM} this compares the reduced objectives
$\min_S F(\tilde\pi,\tilde\sigma,S,X_{t+1})$ and
$\min_S F(\pi_t,\sigma_t,S,X_{t+1})$. The algorithm then recomputes
$S_{t+1}' = \Pi_M(P_{\pi_{t+1}}(A-X_{t+1})P_{\sigma_{t+1}}^\top)$ for the
accepted pair, which is the minimizer in~\eqref{eq:patlr_dM}, so
\begin{align*}
F(\pi_{t+1},\sigma_{t+1},S_{t+1}',X_{t+1})
&= \min_S F(\pi_{t+1},\sigma_{t+1},S,X_{t+1}) \\
&\le \min_S F(\pi_t,\sigma_t,S,X_{t+1})
\le F(\pi_t,\sigma_t,S_{t+1},X_{t+1}).
\end{align*}
If the proposal is rejected, $(\pi_{t+1},\sigma_{t+1}) = (\pi_t,\sigma_t)$ and
the same chain holds with equality. The recomputation matters: without
it the stored $S_{t+1}$ belongs to the old pair and
$F(\pi_{t+1},\sigma_{t+1},S_{t+1},X_{t+1})$ can exceed the previous
value even though the reduced objective decreased.
\end{enumerate}
Combining the three inequalities,
$F(\pi_{t+1},\sigma_{t+1},S_{t+1}',X_{t+1}) \le F(\pi_t,\sigma_t,S_t,X_t)$,
where the left-hand side is exactly the state the loop carries into the
next iteration and returns at termination.
The sequence of $F$-values is therefore non-increasing and bounded below
by $0$, hence convergent.\hfill$\square$

Unlike Proposition~\ref{prop:convergence}, this does not give finite
termination: the permutations range over the finite set $\Sn\times\Sn$,
but $X$ is continuous, and for fixed $(\pi,\sigma)$ the alternating
$S$/$X$ updates are alternating projections that converge in value
without necessarily reaching a fixed point in finitely many steps. The
algorithm stops when the decrease of $F$ falls below a tolerance
$\varepsilon$ or after a capped number of iterations.

\paragraph{Coordinates of $X$.}
The low-rank factor $X$ is kept in the original (unpermuted) coordinates
across iterations: when $(\pi,\sigma)$ changes, $X$ does not need to be
relabeled, whereas a copy stored in the permuted frame would become inconsistent
with the new $(\pi,\sigma)$ and break monotonicity.

\subsection{Identifiability limits of the pattern score}
\label{app:selection}

The score $\tilde{s}_M(A;\pi,\sigma)$ and its minimised form
$\tilde{s}_M(A)$ are those of equation~\eqref{eq:adaptive_score}.
Let $A = P_{\pi^*}^\top S_0\, P_{\sigma^*}$ with
$\supp(S_0)=\supp(M^*)$. Then $d_{M^*}(A)=0$ (the planted pair places
every entry of $S_0$ inside $M^*$), so $\tilde s_{M^*}(A)=0$. For a
candidate $M'$, $P_{\hat\pi} A P_{\hat\sigma}^\top = P_\alpha S_0
P_\beta^\top$ with $\alpha=\hat\pi(\pi^*)^{-1}$,
$\beta=\hat\sigma(\sigma^*)^{-1}$, so $d_{M'}(A;\hat\pi,\hat\sigma)=0$
exactly when $P_\alpha M^* P_\beta^\top \le M'$ entrywise. Hence every
mask that contains a row/column-permuted copy of $M^*$ also scores
exactly $0$, and every mask that does not scores strictly more, for any
strictly positive density penalty $g(\rho_M)$ in place of
$(1-\rho_M)^{3/2}$; a sufficient condition for the latter is
$|M'|_1 < |M^*|_1$. Among the zero-score masks the score does not
distinguish at all: $0$ divided by any penalty is $0$, so the planted
mask and its containers tie, and preferring the sparsest of them is a
separate convention, not a property of the score. The remark concerns
exact support only: a zero residual cannot distinguish a mask from its
containers, whatever the density factor.

\textbf{Role of the exponent.} Under noise the containers acquire
positive scores and the density factor becomes active. The exponent
$3/2$ is a heuristic over-correction of coverage, not a derived
constant: under independent uniform row and column permutations
$\E[d_M(A;\pi,\sigma)] = (1-\rho_M)\|A\|_F^2$
(Appendix~\ref{app:basic}), so the normalized null level is
$(1-\rho_M)^{-1/2}$ and denser masks receive a larger null score,
whereas the exponent $1$ would equalize the null level across
densities. Whether this is enough once the permutations are optimized
is an empirical question, measured in Appendix~\ref{app:adaptive_oracle}.

\subsection{Special masks: permutation-invariant objectives and exact solutions}
\label{app:special}

Table~\ref{tab:special} summarizes the cases; the derivations follow.

\begin{table}[H]
\caption{Masks for which the objective is permutation-invariant or
  exactly solvable.}
\label{tab:special}
\centering
\footnotesize
\begin{tabular}{l l}
\toprule
Mask, setting & Objective \\
\midrule
full mask $M=\mathbf{1}$ & identically $0$ \\
diagonal, $\sigma=\pi$ & constant $\sum_{a\ne b}A_{ab}^2$ (permutation-invariant) \\
upper triangle, $\sigma=\pi$, symmetric $A$ & constant $\tfrac12\sum_{a\ne b}A_{ab}^2$ \\
upper triangle, $\sigma=\pi$, asymmetric $A$ & weight of the backward arcs (linear ordering) \\
diagonal or anti-diagonal, independent $(\pi,\sigma)$ & one LAP (Lemma~\ref{lem:col_sep}) \\
arrowhead, $\sigma=\pi$ & exact: choice of the hub, $O(n^2)$ \\
\bottomrule
\end{tabular}
\end{table}

\textbf{Objectives that do not depend on the permutation.}
For $M=\mathbf{1}$, $d_M(A;\pi,\sigma)=0$ for every $A$, $\pi$, $\sigma$.
Under a common permutation ($\sigma=\pi$) and for any $A$, the diagonal
mask gives $d_M(A;\pi)=\sum_{i\ne j}A_{\pi^{-1}(i),\pi^{-1}(j)}^2
=\sum_{a\ne b}A_{ab}^2$, since the pairs $(\pi^{-1}(i),\pi^{-1}(j))$
with $i\ne j$ range over all ordered pairs $a\ne b$. Under $\sigma=\pi$
and for \emph{symmetric} $A$, the upper-triangular mask gives
$d_M(A;\pi)=\sum_{i>j}A_{\pi^{-1}(i),\pi^{-1}(j)}^2
=\tfrac12\sum_{a\ne b}A_{ab}^2$, because $A_{ab}^2=A_{ba}^2$ and the
pairs $i>j$ cover every unordered pair once; for an asymmetric $A$ the
same quantity is the weight of the backward arcs of the ordering, the
linear-ordering objective. With independent permutations the diagonal
and triangular masks are no longer trivial: the diagonal and the
anti-diagonal are single LAPs
(Lemma~\ref{lem:col_sep}), and the triangle is not constant.

\textbf{Exact solution for the arrowhead mask under $\sigma=\pi$.}
The diagonal is invariant under a common permutation, so the only
choice is which index becomes the hub $h$:
\[
\max_\pi \|\Pi_M(A^{\pi,\pi})\|_F^2 \;=\; \sum_i A_{ii}^2 +
\max_h \sum_{j\ne h}\bigl(A_{hj}^2 + A_{jh}^2\bigr),
\]
computable in $O(n^2)$. The attention pipeline of
Appendix~\ref{app:attention_details} uses this value as the arrowhead
score and as a check on the solver.

% ===========================================================================
\section{Algorithmic Details}
\label{app:tech_alg}
% ===========================================================================

\subsection{Initialization and the exchange phase}
\label{app:init}
\textbf{Initialization.} The default start for every mask is the
Fiedler vector of the graph Laplacian $L = D - |A|$,
$D=\diag(|A|\mathbf{1})$: ordering vertices by their spectral embedding
aligns with bandwidth-minimal orderings~\citep{barnard1995spectral}.
For the anti-diagonal band the Fiedler order of the folded path is
unfolded from both ends and the two starts are kept as candidates;
experiments that start a block fit from a spectral or Bethe-Hessian
partition say so where they are reported
(Appendix~\ref{app:blockmodels}). Random restarts mitigate the residual
local-minimum issue.

\textbf{Common permutation.} For $\sigma=\pi$ the two-sided iterate of
a restart is rounded to one permutation, $\hat\pi \in
\argmin_{\pi\in\{\pi_0,\pi^*,\sigma^*\}} d_M(A;\pi)$ with $\pi_0$ the
start (rounding alone can lose ground, so the start is kept as a
candidate), and the best $\hat\pi$ over restarts is finished by the
exchange descent. With $C$ the permuted weight matrix ($C = B\odot B$
for the pattern distance, $C=B$ for the linear score, $B = P_{\hat\pi} A
P_{\hat\pi}^\top$) and $D = 1-M$, exchanging positions $p$ and $q$
changes the objective $\sum_{ij}D_{ij}C_{ij}$ by
\begin{align*}
\Delta_{pq} \;=\; \sum_{j\ne p,q}\bigl[(D_{pj}-D_{qj})(C_{qj}-C_{pj})
+ (D_{jp}-D_{jq})(C_{jq}-C_{jp})\bigr]\\
+\; (D_{pq}-D_{qp})(C_{qp}-C_{pq}) + (D_{pp}-D_{qq})(C_{qq}-C_{pp}),
\end{align*}
valid for asymmetric $A$ as well; the matrix $(\Delta_{pq})$ is two
products $DC^\top$ and $D^\top C$ plus entrywise terms, $O(n^3)$ per
evaluation. A round evaluates $\Delta$, then applies the improving
exchanges in order of gain on disjoint positions, at most $n/32$ per
round, each re-checked exactly in $O(n)$ under the current order before
it is applied; rounds continue while an exchange improves. The output
is never worse than the start and no exchange of two positions improves
it (Proposition~\ref{prop:convergence}). With several starts the LAP
phase runs from each start and the exchange phase from the best rounded
permutation; on the sparse graphs of Appendix~\ref{app:blockmodels},
where the LAP steps stall early, the two phases are interleaved per
start (three rounds). The exchange phase costs about $1$\,s at
$n{=}500$ and $7$\,s at $n{=}1000$ on one core (Appendix~\ref{app:scaling}).

\subsection{Adaptive PatLR: joint pattern--permutation search}
\label{app:adaptive_search}
\begin{algorithm}[t]
\caption{\textsc{Adaptive PatLR}}
\label{alg:adaptive}
\begin{algorithmic}[1]
\REQUIRE $A$, rank $k$, vocabulary $\mathcal{V}{=}\{M_1,\dots,M_K\}$, budget $N$, cooling $\gamma$, exploration floor $p_{\min}$, proposal steps $\ell$; write $f(j,\pi,\sigma) := \tilde{s}_{M_j}(A;\pi,\sigma)$
\STATE $\pi_0 \leftarrow \textsc{FiedlerInit}(A)$; for every $j'$: $(\pi_{j'},\sigma_{j'}) \leftarrow \textsc{AltLAP}(A, M_{j'}, \ell;\,(\pi_0,\pi_0))$ \COMMENT{screening}
\STATE $j \leftarrow \argmin_{j'} f(j',\pi_{j'},\sigma_{j'})$; $(\pi,\sigma) \leftarrow (\pi_j,\sigma_j)$; $T \leftarrow T_0 > 0$
\STATE $(j^*,\pi^*,\sigma^*) \leftarrow (j,\pi,\sigma)$ \COMMENT{incumbent: best visited state}
\FOR{$t = 1, \dots, N$}
\STATE $(\pi,\sigma) \leftarrow \textsc{AltLAP}(A, M_j, 1;\, (\pi,\sigma))$
\STATE \textbf{if} $f(j,\pi,\sigma) < f(j^*,\pi^*,\sigma^*)$ \textbf{then} $(j^*,\pi^*,\sigma^*) \leftarrow (j,\pi,\sigma)$ \COMMENT{update incumbent}
\STATE \textbf{with prob.}\ $p(T){=}\max(p_{\min}, T/T_0)$:
\STATE \quad $j' \sim \mathrm{Uniform}[K]$; $(\pi',\sigma') \leftarrow \textsc{AltLAP}(A, M_{j'}, \ell;\, (\pi,\sigma))$
\STATE \quad $\Delta \leftarrow f(j',\pi',\sigma') - f(j,\pi,\sigma)$
\STATE \quad \textbf{if} $\Delta < 0$ \textbf{or} $\mathrm{rand}() < e^{-\Delta/T}$ \textbf{then} $(j,\pi,\sigma) \leftarrow (j',\pi',\sigma')$
\STATE \textbf{if} $f(j,\pi,\sigma) < f(j^*,\pi^*,\sigma^*)$ \textbf{then} $(j^*,\pi^*,\sigma^*) \leftarrow (j,\pi,\sigma)$ \COMMENT{update incumbent}
\STATE $T \leftarrow \gamma\, T$
\ENDFOR
\RETURN $\textsc{PatLR}(A, M_{j^*}, k)$ started from $(\pi^*,\sigma^*)$ \COMMENT{Alg.~\ref{alg:patlr}; a fresh \textsc{AltLAP} start on $M_{j^*}$ is also tried and the better kept}
\end{algorithmic}
\end{algorithm}

Settings not stated in Section~\ref{sec:adaptive}: the default
vocabulary is blocks $k\in\{2,3,4,6\}$, bands $w\in\{1,2,4\}$, arrowhead
and block+band ($k{=}3$, $w{=}2$), nine masks with fixed parameters (the
attention study uses the nine-mask vocabulary of Section~\ref{sec:exp_attention});
the exploration probability is $p(T) = \max(p_{\min},\,T/T_0)$ with
$p_{\min}{=}0.15$, $T_0$ is calibrated from
the score spread at the Fiedler initialization (floored at a small
positive constant), $\gamma{=}0.95$, and a proposal is evaluated with
$\ell{=}3$ LAP steps from the current pair (a permutation pair
optimized for block $k{=}6$ is already a reasonable start for block
$k{=}4$ or block+band). The \textsc{AltLAP} calls inside the search are
the LAP phase of Algorithm~\ref{alg:alternating_lap}, without the
exchange phase: the search compares many masks under a small budget,
and finishing every proposal at a local minimum would turn each cheap
evaluation into a full fit. The exchange phase, a fresh three-restart
fit and \textsc{PatLR} are applied to the selected mask only. The
incumbent updates make the output the best
visited state; whether the finite search selects the planted mask is the
empirical question of Section~\ref{sec:exp_adaptive} and
Appendix~\ref{app:adaptive_oracle}.

\paragraph{Out-of-vocabulary structure.}
The selector never rejects: it returns the best-scoring visited state,
so a structure outside the vocabulary (unequal or nested
blocks, a hybrid shape) is approximated by the closest available mask;
in our planted experiments the nearest single-family mask is selected
rather than a noise-level answer. Two practical remedies follow. A best
score close to the random-permutation level indicates that no mask in
the vocabulary explains the matrix. That level is not a universal
constant: for independent random permutations and no optimization,
$\E[\tilde s_M] = (1-\rho_M)/(1-\rho_M)^{3/2} = (1-\rho_M)^{-1/2}$,
under $\sigma=\pi$ it depends on the split of the mass between the
diagonal and the off-diagonal (Appendix~\ref{app:proofs}), and
optimizing over permutations lowers it further, so a threshold on it is a
heuristic calibrated per mask rather than a derived rejection rule;
$A=0$ and $\rho_M=1$ are excluded by convention. The vocabulary is
extended by adding a mask, at a cost linear in $|\mathcal{V}|$.
Estimating mask parameters from the data instead of enumerating fixed
masks (e.g.\ block boundaries by change-point segmentation of the
reordered matrix, alternated with the permutation step), and calibrating
each mask's score against its own random-permutation null rather than a
density correction, are left for future work.

% ===========================================================================
\section{Experimental Details and Additional Results}
\label{app:experiments}
% ===========================================================================

For the experiments summarized in Section~\ref{sec:experiments} this
appendix gives the generators, the information available to each
method, the baseline settings, the full grids and the paired intervals;
it also reports the experiments that appear only here (the compute-budget
ablation, the circular and nested ordering geometries, block-size
estimation, the long-context control and wall-clock scaling), each with
the question it answers. Unless stated otherwise, comparisons are paired
on identical instances over $20$ seeds with $95\%$ bootstrap intervals
of the paired difference.

\subsection{Ablation: Initialization, Restarts, and Compute Budget}
\label{app:ablation}

\begin{figure}[h]
  \centering
  \includegraphics[width=\linewidth]{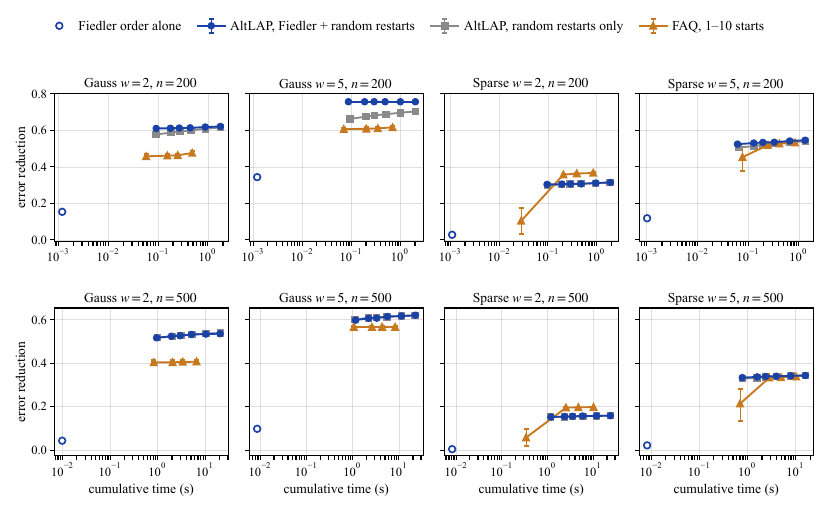}
  \caption{Error reduction against cumulative wall-clock on the instances
    of Table~\ref{tab:baselines} ($20$ seeds, $95\%$ bootstrap intervals;
    single CPU core). Markers: AltLAP after $1,2,3,5,10,20$ restarts
    (Fiedler start plus random, or random only), FAQ after
    $1,3,5,10$ starts; the open circle is the Fiedler order without any
    LAP step.}
  \label{fig:ablation}
\end{figure}

Figure~\ref{fig:ablation} runs the solvers of Table~\ref{tab:baselines}
on the same instances with increasing budgets and records the running
best error reduction against cumulative wall-clock: Alternating LAP from
the Fiedler order plus random restarts (the protocol of
Table~\ref{tab:baselines}, five restarts there) and from random restarts
only, up to $20$ restarts, and FAQ with $1$ to $10$ starts. Three
things are visible. The first run accounts for almost all of the gain
over the Fiedler order alone; additional restarts add at most $0.04$
over $20$ restarts, and random starts reach the values of the Fiedler
start on all but one panel (Gauss $w{=}5$, $n{=}200$, where the Fiedler
start stays $0.05$ ahead), so on these planted instances the descent,
not the spectral start, does the work. On the sparse $\{0,1\}$ graphs
the restart budget adds about $0.01$: the LAP steps stall on the many
ties induced by sparse binary inputs (the same stalling appears without any
noise on the $w{=}1$ band, the Hamiltonian-path case of
Appendix~\ref{app:hamiltonian}) and the exchange phase supplies the
remaining descent (next paragraph). FAQ with one start is weak there
and owes its final value to its randomized starts; on the Gaussian
inputs Alternating LAP is above FAQ at every budget.

\paragraph{Without the exchange phase.}
Running Algorithm~\ref{alg:alternating_lap} with the exchange phase
disabled, on the instances and budget of Table~\ref{tab:baselines},
separates the two phases. The LAP phase alone reaches ER
$0.61/0.50/0.41$ (Gauss $w{=}2$), $0.78/0.65/0.55$ (Gauss $w{=}5$),
$0.15/0.06/0.01$ (sparse $w{=}2$) and $0.63/0.48/0.29$ (sparse $w{=}5$)
at $n{=}100/200/500$; the exchange phase raises these to
$0.69/0.61/0.52$, $0.87/0.75/0.61$, $0.45/0.30/0.15$ and
$0.69/0.53/0.34$, the values of Table~\ref{tab:baselines}. On the
Gaussian inputs it adds $0.06$--$0.11$ and on the sparse graphs it
supplies most of the reduction, since the LAP steps stall on the ties
induced by binary inputs, the same stalling as on the sparse SBM
graphs of Section~\ref{sec:exp_blocks}. At $n{=}500$ the phase takes about a third of the run time on the
Gaussian instances and about a tenth on the sparse graphs (runtimes in
Appendix~\ref{app:baselines}).

\subsection{Banded Baselines from the Seriation, QAP, and Bandwidth Literatures}
\label{app:baselines}

Table~\ref{tab:baselines_full} extends Table~\ref{tab:baselines} with all
methods and both of our metrics (ER\,/\,BE), and
Table~\ref{tab:bandwidth} scores the same orderings by the bandwidth
family's own criteria. Methods: RCM~\citep{cuthill1969reducing} and
Gibbs--Poole--Stockmeyer (GPS)~\citep{gibbs1976gps} on the $\{0,1\}$
support (the graph itself in the sparse regime; the KNN-$20$
sparsification in the Gaussian regime); F+greedy (Fiedler order plus
greedy adjacent swaps on 2-SUM); the plain spectral 2-SUM order of
\citet{atkins1998spectral} (the Fiedler ordering itself); convex seriation~\citep{fogel2013seriation}
(Frank--Wolfe over the permutahedron with the concave vertex-pulling
term, penalty weight and start chosen by the 2-SUM value); QAP-FAQ
with three starts; FAQ followed by AltLAP (keeping the better of the
two); AltLAP with one restart (equal-budget variant) and with five.
Our objective and FAQ's are the same QAP, but the two heuristics solve
different inner problems: FAQ relaxes the single-permutation problem,
whereas AltLAP descends on the two-sided objective, rounds to one
permutation and finishes with the exchange descent; the rounding can
lose ground, which is why the start is kept as a candidate. The two
are closely related: at a permutation $P_\pi$ the FAQ gradient
$\nabla_P\,\mathrm{tr}(M^\top P W P^\top) = 2MP_\pi W$ equals
$-2Q$ of Lemma~\ref{lem:col_sep} up to a constant per column, so the first
FAQ direction-finding step from a permutation is the LAP step of
Algorithm~\ref{alg:alternating_lap}; FAQ then takes a fractional step
inside the Birkhoff polytope, while AltLAP moves to the vertex and
alternates sides. Warm-starting AltLAP from the FAQ order (the
FAQ+AltLAP column of Table~\ref{tab:baselines_full}) adds $0.02$--$0.12$
on the Gaussian inputs, where the LAP steps rarely move the FAQ order
and the gain is the exchange phase, and at most $0.01$ on the sparse
graphs.

\begin{table}[h]
\caption{Banded recovery, ER\,/\,BE (higher = better), $20$ seeds.
  Bold marks the best ER per row (ties within $0.005$).}
\label{tab:baselines_full}
\centering
\scriptsize
\setlength{\tabcolsep}{2.5pt}
\begin{tabular}{l r ccccccccc}
\toprule
Regime & $n$ & RCM & GPS & F+greedy & Spectral & Convex & FAQ & FAQ+AltLAP & AltLAP-r1 & AltLAP \\
\midrule
Gauss $w{=}2$ & 30 & 0.03/0.39 & 0.05/0.41 & 0.84/0.90 & 0.76/0.85 & 0.85/0.91 & 0.60/0.75 & 0.68/0.80 & \textbf{0.87/0.92} & \textbf{0.87/0.92} \\
 & 100 & 0.05/0.34 & 0.05/0.34 & 0.37/0.56 & 0.32/0.52 & 0.36/0.56 & 0.51/0.66 & 0.62/0.73 & \textbf{0.69/0.78} & \textbf{0.69/0.78} \\
 & 200 & 0.03/0.30 & 0.04/0.31 & 0.14/0.38 & 0.15/0.39 & 0.14/0.38 & 0.46/0.61 & 0.58/0.70 & \textbf{0.61/0.72} & \textbf{0.61/0.72} \\
 & 500 & 0.03/0.27 & 0.03/0.27 & 0.03/0.27 & 0.04/0.28 & 0.04/0.28 & 0.40/0.55 & \textbf{0.52/0.64} & 0.52/0.64 & \textbf{0.52/0.64} \\
\midrule
Gauss $w{=}5$ & 30 & 0.07/0.54 & 0.09/0.55 & \textbf{0.98/0.99} & 0.97/0.98 & \textbf{0.98/0.99} & 0.84/0.92 & 0.86/0.93 & \textbf{0.98/0.99} & \textbf{0.98/0.99} \\
 & 100 & 0.09/0.41 & 0.09/0.41 & 0.72/0.82 & 0.61/0.75 & 0.73/0.82 & 0.66/0.78 & 0.72/0.82 & \textbf{0.87/0.91} & \textbf{0.87/0.91} \\
 & 200 & 0.08/0.35 & 0.08/0.36 & 0.33/0.53 & 0.34/0.54 & 0.34/0.54 & 0.61/0.73 & 0.66/0.76 & \textbf{0.76/0.83} & 0.75/0.82 \\
 & 500 & 0.06/0.30 & 0.05/0.29 & 0.08/0.32 & 0.10/0.33 & 0.09/0.32 & 0.57/0.68 & 0.60/0.70 & 0.60/0.70 & \textbf{0.61/0.71} \\
\midrule
Sparse $w{=}2$ & 30 & 0.23/0.33 & 0.24/0.33 & 0.45/0.51 & 0.39/0.47 & 0.46/0.52 & 0.73/0.76 & \textbf{0.74/0.77} & 0.67/0.71 & 0.67/0.71 \\
 & 100 & 0.06/0.10 & 0.06/0.10 & 0.09/0.13 & 0.09/0.13 & 0.09/0.13 & \textbf{0.49/0.51} & \textbf{0.50/0.52} & 0.44/0.46 & 0.45/0.47 \\
 & 200 & 0.02/0.05 & 0.02/0.04 & 0.03/0.05 & 0.03/0.05 & 0.03/0.05 & \textbf{0.36/0.37} & \textbf{0.36/0.37} & 0.30/0.32 & 0.30/0.31 \\
 & 500 & 0.00/0.01 & 0.00/0.01 & 0.00/0.01 & 0.00/0.01 & 0.01/0.01 & \textbf{0.20/0.20} & \textbf{0.20/0.20} & 0.15/0.16 & 0.15/0.16 \\
\midrule
Sparse $w{=}5$ & 30 & 0.41/0.59 & 0.04/0.34 & 0.78/0.85 & 0.76/0.83 & 0.78/0.85 & 0.88/0.92 & 0.89/0.93 & \textbf{0.91/0.94} & \textbf{0.91/0.94} \\
 & 100 & 0.16/0.24 & 0.15/0.23 & 0.29/0.36 & 0.29/0.36 & 0.30/0.37 & 0.66/0.69 & 0.66/0.69 & 0.68/0.71 & \textbf{0.69/0.72} \\
 & 200 & 0.09/0.14 & 0.08/0.13 & 0.10/0.14 & 0.12/0.16 & 0.10/0.15 & 0.52/0.54 & 0.52/0.55 & 0.53/0.55 & \textbf{0.53/0.56} \\
 & 500 & 0.03/0.05 & 0.02/0.04 & 0.02/0.04 & 0.02/0.04 & 0.02/0.04 & \textbf{0.33/0.35} & \textbf{0.34/0.35} & \textbf{0.33/0.35} & \textbf{0.34/0.35} \\
\bottomrule
\end{tabular}
\end{table}

\textbf{Two metric families, one support.}
Bandwidth $\max\{|i-j| : S_{ij}=1\}$ and profile
$\sum_i (i-\min\{j\le i : S_{ij}=1\})$ need a discrete support, so every
ordering is scored on the same $\{0,1\}$ support that RCM and GPS are
given. GPS is the better bandwidth reducer, as expected, and neither
bandwidth method reduces off-band mass. Conversely our methods barely
reduce bandwidth: a few small entries far from the diagonal cost almost
no mass but fix $\max|i-j|$. On noisy inputs the classical criterion
saturates (at $n{=}500$ in the Gaussian regime every ordering, including
the identity, has a bandwidth close to the maximum), because the signal
falls below the noise at $|i-j|\approx 7$ and the support beyond is
noise, so the criterion is set by the farthest noise entries while the
pattern distance measures the mass concentration.

\begin{table}[h]
\caption{The same orderings scored by the bandwidth family: bandwidth
  (lower = better; identity order and six methods) and profile for RCM,
  GPS and AltLAP, means over $20$ seeds on the shared $\{0,1\}$ support.}
\label{tab:bandwidth}
\centering
\scriptsize
\setlength{\tabcolsep}{3pt}
\begin{tabular}{l r r rrrrrr rrr}
\toprule
 & & & \multicolumn{6}{c}{Bandwidth} & \multicolumn{3}{c}{Profile} \\
\cmidrule(lr){4-9} \cmidrule(lr){10-12}
Regime & $n$ & identity & RCM & GPS & F+greedy & Convex & FAQ & AltLAP & RCM & GPS & AltLAP \\
\midrule
Gauss $w{=}2$ & 30 & 29 & \textbf{26} & \textbf{26} & 28 & 29 & 29 & 29 & \textbf{416} & \textbf{417} & 422 \\
 & 100 & 97 & \textbf{82} & 83 & 96 & 95 & 97 & 96 & \textbf{4251} & 4515 & 4466 \\
 & 200 & 196 & 162 & \textbf{159} & 192 & 192 & 197 & 195 & \textbf{16809} & 17712 & 17592 \\
 & 500 & 494 & 405 & \textbf{399} & 489 & 490 & 496 & 492 & \textbf{104377} & 109910 & 110834 \\
\midrule
Gauss $w{=}5$ & 30 & 29 & \textbf{26} & \textbf{26} & 28 & 29 & 29 & 29 & \textbf{416} & \textbf{417} & 423 \\
 & 100 & 97 & \textbf{82} & 83 & 96 & 95 & 97 & 97 & \textbf{4251} & 4515 & 4476 \\
 & 200 & 196 & 162 & \textbf{159} & 192 & 192 & 196 & 194 & \textbf{16809} & 17712 & 17598 \\
 & 500 & 494 & 405 & \textbf{399} & 489 & 490 & 493 & 494 & \textbf{104377} & 109910 & 110756 \\
\midrule
Sparse $w{=}2$ & 30 & 26 & 10 & \textbf{8} & 10 & 10 & 25 & 16 & 135 & 142 & \textbf{120} \\
 & 100 & 94 & 47 & \textbf{45} & 54 & 52 & 92 & 87 & \textbf{2298} & 2634 & 2363 \\
 & 200 & 192 & 117 & \textbf{111} & 139 & 133 & 190 & 192 & \textbf{11442} & 12589 & 13083 \\
 & 500 & 492 & 365 & \textbf{358} & 455 & 441 & 491 & 491 & \textbf{93432} & 101738 & 103868 \\
\midrule
Sparse $w{=}5$ & 30 & 29 & 29 & \textbf{24} & 29 & 29 & 29 & 29 & 430 & \textbf{407} & 428 \\
 & 100 & 96 & 58 & \textbf{55} & 70 & 71 & 93 & 91 & 2823 & 3165 & \textbf{2483} \\
 & 200 & 195 & 131 & \textbf{128} & 164 & 159 & 192 & 189 & \textbf{13100} & 14894 & 13319 \\
 & 500 & 493 & \textbf{392} & 393 & 456 & 460 & 493 & 490 & \textbf{98342} & 106402 & 103963 \\
\bottomrule
\end{tabular}
\end{table}

Runtimes at $n{=}500$ (one EPYC core): RCM, GPS and the spectral order
$0.01$--$0.02$\,s; convex seriation $0.45$--$0.55$\,s; F+greedy
$0.7$\,s; FAQ (three starts) $2.0$--$2.8$\,s; AltLAP with one restart
$0.8$--$1.2$\,s and with five $1.7$--$4.1$\,s (the exchange phase runs
once, from the best restart).

\subsection{PatLR Decomposition Visualizations}
\label{app:patlr_visual}

\begin{figure}[h]
  \centering
  \includegraphics[width=\linewidth]{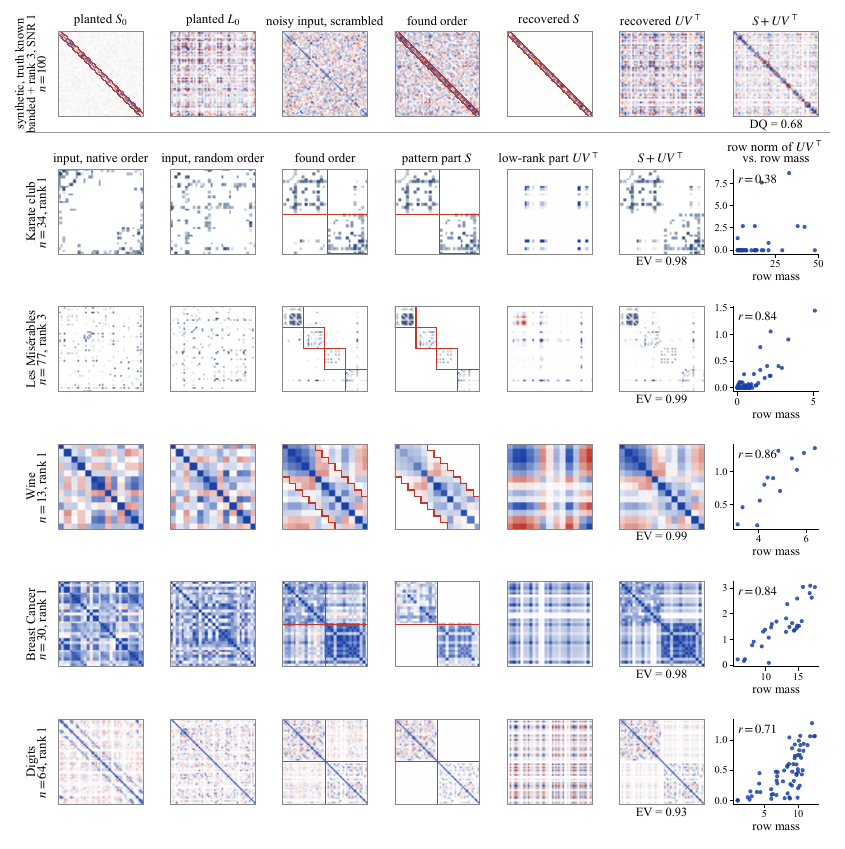}
  \caption{PatLR step by step. Top row: a synthetic instance with known
    truth (banded $w{=}3$ pattern plus a rank-$3$ component of equal
    energy, $n{=}100$, scrambled, SNR~$1$): the planted $S_0$ and $L_0$,
    the noisy input, the input in the found order, the recovered $S$ and
    $UV^\top$ and their sum with its DQ. Remaining rows: real matrices
    without added noise,
    karate club (weighted adjacency, $\mathrm{block}(18,16)$ from the two
    factions, rank $1$); Les Mis\'erables (co-occurrence weights scaled to
    $[0,1]$, $\mathrm{block}(19,19,19,20)$ template, rank $3$); and the
    signed feature-correlation matrices of Wine (band $w{=}4$), Breast
    Cancer ($\mathrm{block}(15,15)$) and Digits (pixel correlations,
    $\mathrm{block}(32,32)$), rank $1$; left to right: the input in its
    native order; the input in the random order given to the solver; the
    input in the order found by PatLR, mask outlined;
    the pattern part $S$; the low-rank part $UV^\top$; their sum with its
    explained variance; the row norm of $UV^\top$ against the row mass
    $\sum_j|A_{ij}|$ (weighted degree for the graphs, total absolute
    correlation for the features). Graphs on a white-to-blue scale, signed
    matrices and $UV^\top$ on a diverging scale (blue positive); $r$ is
    the correlation in the last column.}
  \label{fig:patlr_real}
\end{figure}

Figure~\ref{fig:patlr_real} follows the decomposition on one synthetic
instance of the experiment of Figure~\ref{fig:denoise}, where the planted
parts are known and can be compared with the recovered ones, and, without
added noise, on the two real networks of Section~\ref{sec:exp_patlr} and
on three feature-correlation matrices from the UCI
repository~\citep[Wine, Breast Cancer Wisconsin (Diagnostic) and Optical
Recognition of Handwritten Digits;][]{aeberhard1992wine,wolberg1993wdbc,alpaydin1998optdigits}, each with
the mask of the default vocabulary that fits it best at rank $1$. The first
column shows each matrix in its native order (node numbering of the graph,
feature order of the dataset), the second the random order given to the
solver. The row norm of the low-rank part follows the row mass on Les
Mis\'erables, Wine and Breast Cancer; on the karate club the rank-$1$
part concentrates on the two highest-degree nodes, the two faction
leaders.

\subsection{PatLR: Full Denoising Grid, Ablations and Capacity Control}
\label{app:capacity}

\paragraph{Setup of Figure~\ref{fig:denoise}, remaining details.}
The pattern part is $S_0 = 2\,M\odot R + 0.15\,(1-M)\odot R$ for one
symmetric Gaussian matrix $R$, with the diagonal set to $|R_{ii}|+0.3$
(the generator of the capacity control below); $S_0$ is symmetric for
the six symmetric masks, while for the upper-triangular mask it is
asymmetric (upper triangle at scale $2$, lower at $0.15$), so under the
common permutation its objective is the linear-ordering problem of
Table~\ref{tab:patterns}, not the constant it would be for a symmetric
matrix. The low-rank part is $L_0 = UU^\top$
with $U\in\mathbb{R}^{n\times3}$ Gaussian, scaled to a fraction $f$ of
the clean energy ($f{=}0.5$ in Figure~\ref{fig:denoise}); the noise is
scaled per instance to $\mathrm{SNR}=\|A\|_F^2/\|E\|_F^2$, so the noisy
input has DQ $1-1/\mathrm{SNR}$. RPCA is principal component pursuit
with its low-rank part as the estimate; SVD-dof is the SVD at the rank
with the parameter count of PatLR (rule below); PatLR runs from the Fiedler order
and from its unfolding (one random restart each) and keeps the lower
objective; PatLR-rnd uses a random mask of the same density; the oracle
column fixes the permutation to the planted one and alternates $S$ and
$X$ only.

Table~\ref{tab:denoise_full} gives the full grid. Three regularities hold
across it. An estimate that copies the in-mask entries, i.e.\
$\hat C = M\odot(C+E) + (1-M)\odot X$ in the coordinates of the mask
(pattern alone with $X=0$, PatLR and the oracle with the rank-$3$ $X$),
has error $\|M\odot E\|_F^2 + \|(1-M)\odot(X-C)\|_F^2$ exactly, the two
terms having disjoint supports, so its DQ is at most
$1-\|M\odot E\|_F^2/\|C\|_F^2$. For a mask placed independently of the
noise the in-mask noise energy is $\rho_M\|E\|_F^2$ in expectation,
which gives the value $1-\rho_M/\mathrm{SNR}$; a permutation chosen from the data tends
to place more noise inside the mask, not less. The off-mask term is the
clean energy outside the mask for the pattern-only estimate ($X=0$) and
$\|(1-M)\odot(X-C)\|_F^2$ for PatLR, which is smaller when $C$ has
low-rank structure outside the mask and larger when $X$ fits noise. On
the pattern-only families the off-mask clean values are i.i.d.\ and
carry no low-rank structure, so on the five families whose order the
solver recovers the rank-$3$ oracle lands below the pattern-only value
by $0.01$ at SNR~$4$, $0.05$ at SNR~$1$ and $0.07$--$0.11$ at SNR~$0.5$
(on the band and the anti-diagonal the oracle's planted order puts it
above the pattern-only estimate, whose order is the solver's). The gap between PatLR and the
oracle is concentrated on the banded and anti-diagonal families, i.e.\ in
the permutation search on sparse masks, and grows as the SNR falls. The
random-mask control PatLR-rnd lies $0.2$--$0.5$ below PatLR on the
planted pattern and $0.1$--$0.4$ below it with the rank-3 component at
SNR~$1$; it stays above the SVD level where random positions can absorb
signal mass under a free permutation (the band, anti-diagonal, block
and block+band families on the planted pattern, the band and
anti-diagonal with the rank-3 component) and within $0.05$ of it or
below elsewhere.
With $80\%$ of the clean energy in the low-rank part, SVD-best is the better
estimate on the dense masks, and with $20\%$ PatLR is the best estimate
on every family: the mask pays for the in-mask noise it copies only while
the pattern carries enough of the signal.

\begin{table}[p]
\caption{Denoising quality on the seven families, all settings of the
  experiment of Figure~\ref{fig:denoise} ($n{=}200$; $20$ seeds unless
  stated; best of SVD-dof, SVD-best, RPCA, pattern only, PatLR and
  PatLR-rnd per row in bold; ``oracle'' is PatLR at the planted
  permutation).}
\label{tab:denoise_full}
\centering
\scriptsize
\setlength{\tabcolsep}{3.5pt}
\begin{tabular}{l r r r r r r r r r}
\toprule
Pattern & $\rho_M$ & input & SVD-dof & SVD-best & RPCA & pattern only & \textbf{PatLR} & PatLR-rnd & oracle \\
\midrule
\multicolumn{10}{l}{\emph{planted pattern, SNR 4.0}} \\
Banded & 0.035 & 0.75 & 0.12 & 0.75 & 0.10 & 0.76 & \textbf{0.76} & 0.55 & 0.88 \\
Block & 0.250 & 0.75 & 0.34 & 0.75 & 0.21 & \textbf{0.92} & 0.91 & 0.54 & 0.91 \\
Arrowhead & 0.015 & 0.75 & 0.34 & 0.75 & 0.19 & \textbf{0.78} & \textbf{0.78} & 0.36 & 0.78 \\
Checkerboard & 0.500 & 0.75 & 0.53 & 0.75 & 0.34 & \textbf{0.87} & 0.86 & 0.60 & 0.86 \\
Block+band & 0.327 & 0.75 & 0.41 & 0.75 & 0.25 & \textbf{0.91} & 0.89 & 0.55 & 0.89 \\
Anti-diagonal & 0.102 & 0.75 & 0.21 & 0.75 & 0.12 & \textbf{0.86} & \textbf{0.86} & 0.55 & 0.91 \\
Triangular & 0.502 & 0.75 & 0.66 & 0.75 & 0.41 & \textbf{0.87} & 0.86 & 0.53 & 0.86 \\
\midrule
\multicolumn{10}{l}{\emph{planted pattern, SNR 1.0}} \\
Banded & 0.035 & 0.00 & 0.03 & 0.07 & -0.05 & \textbf{0.69} & 0.64 & 0.44 & 0.81 \\
Block & 0.250 & -0.00 & 0.01 & 0.02 & 0.11 & \textbf{0.73} & 0.68 & 0.20 & 0.68 \\
Arrowhead & 0.015 & 0.00 & 0.27 & 0.28 & 0.03 & \textbf{0.77} & 0.73 & 0.27 & 0.72 \\
Checkerboard & 0.500 & -0.00 & -0.00 & 0.01 & 0.21 & \textbf{0.49} & 0.44 & 0.07 & 0.44 \\
Block+band & 0.327 & -0.00 & 0.01 & 0.01 & 0.15 & \textbf{0.66} & 0.60 & 0.14 & 0.60 \\
Anti-diagonal & 0.102 & 0.00 & 0.03 & 0.04 & 0.01 & \textbf{0.72} & 0.67 & 0.37 & 0.80 \\
Triangular & 0.502 & -0.00 & 0.11 & 0.12 & 0.25 & \textbf{0.48} & 0.44 & 0.01 & 0.44 \\
\midrule
\multicolumn{10}{l}{\emph{planted pattern, SNR 0.5}} \\
Banded & 0.035 & -1.00 & -0.10 & -0.02 & -0.28 & \textbf{0.59} & 0.49 & 0.28 & 0.71 \\
Block & 0.250 & -1.00 & -0.42 & -0.02 & -0.11 & \textbf{0.48} & 0.37 & -0.28 & 0.37 \\
Arrowhead & 0.015 & -1.00 & 0.19 & 0.25 & -0.22 & \textbf{0.75} & 0.67 & 0.14 & 0.65 \\
Checkerboard & 0.500 & -1.00 & -0.72 & -0.02 & -0.04 & \textbf{-0.01} & -0.12 & -0.63 & -0.12 \\
Block+band & 0.327 & -1.00 & -0.53 & -0.02 & -0.08 & \textbf{0.32} & 0.21 & -0.40 & 0.22 \\
Anti-diagonal & 0.102 & -1.00 & -0.20 & -0.01 & -0.21 & \textbf{0.50} & 0.36 & 0.08 & 0.64 \\
Triangular & 0.502 & -1.00 & -0.63 & \textbf{-0.01} & \textbf{-0.01} & -0.04 & -0.13 & -0.68 & -0.11 \\
\midrule
\multicolumn{10}{l}{\emph{pattern + rank-3 component (equal energy), SNR 4.0}} \\
Banded & 0.035 & 0.75 & 0.55 & 0.75 & 0.54 & 0.42 & \textbf{0.86} & 0.76 & 0.94 \\
Block & 0.250 & 0.75 & 0.62 & 0.75 & 0.60 & 0.57 & \textbf{0.92} & 0.71 & 0.92 \\
Arrowhead & 0.015 & 0.75 & 0.67 & 0.76 & 0.58 & 0.44 & \textbf{0.89} & 0.58 & 0.89 \\
Checkerboard & 0.500 & 0.75 & 0.68 & 0.75 & 0.65 & 0.63 & \textbf{0.86} & 0.71 & 0.85 \\
Block+band & 0.327 & 0.75 & 0.64 & 0.75 & 0.62 & 0.59 & \textbf{0.90} & 0.70 & 0.90 \\
Anti-diagonal & 0.102 & 0.75 & 0.57 & 0.75 & 0.55 & 0.43 & \textbf{0.88} & 0.75 & 0.94 \\
Triangular & 0.502 & 0.75 & 0.74 & 0.76 & 0.67 & 0.63 & \textbf{0.86} & 0.67 & 0.86 \\
\midrule
\multicolumn{10}{l}{\emph{pattern + rank-3 component (equal energy), SNR 1.0}} \\
Banded & 0.035 & 0.00 & 0.47 & 0.51 & 0.35 & 0.34 & \textbf{0.77} & 0.66 & 0.88 \\
Block & 0.250 & -0.00 & 0.30 & 0.49 & 0.43 & 0.38 & \textbf{0.71} & 0.37 & 0.71 \\
Arrowhead & 0.015 & 0.00 & 0.63 & 0.63 & 0.37 & 0.43 & \textbf{0.85} & 0.50 & 0.85 \\
Checkerboard & 0.500 & 0.00 & 0.15 & \textbf{0.49} & 0.46 & 0.24 & 0.46 & 0.19 & 0.45 \\
Block+band & 0.327 & -0.00 & 0.24 & 0.49 & 0.44 & 0.33 & \textbf{0.63} & 0.31 & 0.63 \\
Anti-diagonal & 0.102 & -0.00 & 0.41 & 0.49 & 0.38 & 0.26 & \textbf{0.69} & 0.56 & 0.84 \\
Triangular & 0.502 & 0.00 & 0.19 & \textbf{0.49} & 0.47 & 0.24 & 0.45 & 0.16 & 0.46 \\
\midrule
\multicolumn{10}{l}{\emph{pattern + rank-3 component (equal energy), SNR 0.5}} \\
Banded & 0.035 & -1.00 & 0.37 & 0.47 & 0.08 & 0.22 & \textbf{0.61} & 0.50 & 0.81 \\
Block & 0.250 & -1.00 & -0.12 & \textbf{0.46} & 0.15 & -0.09 & 0.36 & -0.08 & 0.42 \\
Arrowhead & 0.015 & -1.00 & 0.56 & 0.56 & 0.09 & 0.41 & \textbf{0.80} & 0.38 & 0.80 \\
Checkerboard & 0.500 & -1.00 & -0.57 & \textbf{0.46} & 0.17 & -0.39 & -0.22 & -0.50 & -0.08 \\
Block+band & 0.327 & -1.00 & -0.28 & \textbf{0.46} & 0.16 & -0.19 & 0.18 & -0.22 & 0.27 \\
Anti-diagonal & 0.102 & -1.00 & 0.19 & \textbf{0.45} & 0.11 & -0.04 & 0.35 & 0.27 & 0.71 \\
Triangular & 0.502 & -1.00 & -0.54 & \textbf{0.46} & 0.18 & -0.28 & -0.12 & -0.52 & -0.08 \\
\midrule
\multicolumn{10}{l}{\emph{pattern + rank-3 component carrying 20\% of the clean energy, SNR 1 ($10$ seeds)}} \\
Banded & 0.035 & 0.00 & 0.21 & 0.22 & 0.11 & 0.55 & \textbf{0.71} & 0.53 & 0.85 \\
Block & 0.250 & -0.00 & 0.13 & 0.20 & 0.24 & 0.59 & \textbf{0.70} & 0.27 & 0.70 \\
Arrowhead & 0.015 & 0.00 & 0.42 & 0.42 & 0.16 & 0.64 & \textbf{0.70} & 0.29 & 0.79 \\
Checkerboard & 0.500 & -0.00 & 0.06 & 0.20 & 0.31 & 0.40 & \textbf{0.46} & 0.11 & 0.45 \\
Block+band & 0.327 & 0.00 & 0.10 & 0.20 & 0.26 & 0.53 & \textbf{0.62} & 0.21 & 0.63 \\
Anti-diagonal & 0.102 & 0.00 & 0.18 & 0.20 & 0.16 & 0.55 & \textbf{0.67} & 0.45 & 0.83 \\
Triangular & 0.502 & -0.00 & 0.14 & 0.21 & 0.33 & 0.38 & \textbf{0.45} & 0.06 & 0.46 \\
\midrule
\multicolumn{10}{l}{\emph{pattern + rank-3 component carrying 80\% of the clean energy, SNR 1 ($10$ seeds)}} \\
Banded & 0.035 & 0.00 & 0.73 & 0.78 & 0.57 & 0.14 & \textbf{0.81} & 0.77 & 0.91 \\
Block & 0.250 & -0.00 & 0.47 & \textbf{0.78} & 0.60 & 0.16 & 0.57 & 0.48 & 0.72 \\
Arrowhead & 0.015 & 0.00 & 0.81 & 0.81 & 0.57 & 0.21 & \textbf{0.91} & 0.73 & 0.91 \\
Checkerboard & 0.500 & -0.00 & 0.23 & \textbf{0.78} & 0.60 & 0.12 & 0.35 & 0.27 & 0.43 \\
Block+band & 0.327 & -0.00 & 0.38 & \textbf{0.78} & 0.60 & 0.15 & 0.49 & 0.41 & 0.64 \\
Anti-diagonal & 0.102 & 0.00 & 0.63 & \textbf{0.78} & 0.58 & 0.07 & 0.67 & 0.66 & 0.86 \\
Triangular & 0.502 & -0.00 & 0.24 & \textbf{0.78} & 0.60 & 0.10 & 0.43 & 0.26 & 0.46 \\
\bottomrule
\end{tabular}
\end{table}

\paragraph{Ablation of the permutation.}
Table~\ref{tab:patlr_ablation} fixes the permutation instead of searching
it, on the planted-pattern matrices at $\sigma{=}0.3$ (scrambled). With
the permutation fixed to the input order the effect differs between
families: block, checkerboard, block+band, anti-diagonal and triangular
fall below the SVD level, whereas the masks that contain the diagonal
keep what a symmetric scrambling leaves in place; the band stays above
SVD on the diagonal entries alone, and the arrowhead reaches the
searched value because its hub row and column form a rank-$2$ matrix
that the low-rank term absorbs at any order. With the permutation fixed
to the planted order PatLR is high on every family. The joint solver
attains the planted-order value on five families and stays below it on
banded and anti-diagonal, the two sparse masks whose order is hardest
to recover.

\begin{table}[h]
\caption{PatLR with the permutation searched, fixed to the input order, or
  fixed to the planted order (DQ, $n{=}200$, $\sigma{=}0.3$, scrambled
  inputs, $20$ seeds).}
\label{tab:patlr_ablation}
\centering
\footnotesize
\setlength{\tabcolsep}{5pt}
\begin{tabular}{l r r r r r}
\toprule
Pattern & SVD-dof & pattern only & PatLR, input order & \textbf{PatLR} & PatLR, planted order \\
\midrule
Banded & 0.10 & 0.73 & 0.29 & 0.73 & 0.86 \\
Block & 0.41 & 0.96 & 0.31 & 0.96 & 0.96 \\
Arrowhead & 0.28 & 0.48 & 0.74 & 0.74 & 0.73 \\
Checkerboard & 0.68 & 0.97 & 0.54 & 0.97 & 0.97 \\
Block+band & 0.50 & 0.96 & 0.38 & 0.97 & 0.96 \\
Anti-diagonal & 0.21 & 0.81 & 0.14 & 0.83 & 0.92 \\
Triangular & 0.79 & 0.97 & 0.55 & 0.97 & 0.97 \\
\bottomrule
\end{tabular}
\end{table}

\paragraph{Capacity control at a matched parameter count.}
Table~\ref{tab:capacity_full} compares the models at an equal nominal
parameter count on the planted-pattern matrices at $\sigma{=}0.3$ in
their planted order (the scrambled version changes no entry by more than
$0.01$).
PatLR with mask $M$ and residual rank $r$ has
$\mathrm{df} = |M| + r(2n-r)$ free parameters (the in-mask entries plus a
rank-$r$ factorization). SVD-dof receives the rank
$r'=\mathrm{round}\bigl(n-\sqrt{\max(n^2-\mathrm{df},0)}\bigr)$ clipped to
$[1,n]$, which has the same count; SVD-$3$ is the fixed-rank estimate.
The blockmodel baseline is spectral co-clustering~\citep{dhillon2001coclustering}:
$k$-means on the leading singular vectors of rows and of columns, then
block means. ``BM'' chooses $k\in\{2,3,4,5,6,8\}$ by the best DQ (an
oracle choice, generous to the baseline); ``BM-dof'' uses
$k=\mathrm{round}(\sqrt{\mathrm{df}})$, the same nominal budget as PatLR.
PatLR-rnd replaces the planted mask by a random mask of the same density,
redrawn per seed, with an otherwise identical pipeline. At $\sigma{=}0.3$
the noise carries up to half of the energy of these matrices (column
``input''), so the table measures which model class
retains a full-rank in-mask structure at a given parameter count rather
than noise removal; the i.i.d.\ in-mask values are a model mismatch for
the blockmodel, which is why it sits at the SVD-dof level at matched
budget. The hierarchical pattern of Table~\ref{tab:patterns} is omitted: as a
binary mask a two-level hierarchy coincides with its coarsest level.

\begin{table}[h]
\caption{Capacity control, DQ at $n{=}200$, $\sigma{=}0.3$, $20$ seeds:
  $\rho_M$ is the mask density, $r'$ the matched SVD rank, $k'$ the
  matched number of co-clusters, ``input'' the DQ of the noisy matrix
  itself.}
\label{tab:capacity_full}
\centering
\footnotesize
\setlength{\tabcolsep}{3pt}
\begin{tabular}{l r r r r r r r r r r r}
\toprule
Pattern & $\rho_M$ & $r'$ & $k'$ & input & SVD-$3$ & SVD-dof & RPCA & BM & BM-dof & \textbf{PatLR} & PatLR-rnd \\
\midrule
Banded        & 0.035 & 7  & 51  & 0.57 & 0.05 & 0.10 & 0.07 & 0.04 & 0.27 & \textbf{0.73} & 0.52 \\
Block         & 0.250 & 30 & 106 & 0.92 & 0.06 & 0.41 & 0.18 & 0.04 & 0.42 & \textbf{0.96} & 0.61 \\
Arrowhead     & 0.015 & 5  & 42  & 0.13 & 0.28 & 0.28 & 0.05 & 0.27 & 0.43 & \textbf{0.74} & 0.29 \\
Checkerboard  & 0.500 & 63 & 146 & 0.96 & 0.06 & 0.68 & 0.31 & 0.04 & 0.62 & \textbf{0.97} & 0.73 \\
Block+band    & 0.327 & 40 & 120 & 0.93 & 0.06 & 0.51 & 0.22 & 0.04 & 0.42 & \textbf{0.97} & 0.64 \\
Anti-diagonal & 0.102 & 14 & 73  & 0.79 & 0.06 & 0.21 & 0.13 & 0.04 & 0.26 & \textbf{0.83} & 0.57 \\
Triangular    & 0.502 & 63 & 146 & 0.96 & 0.07 & 0.79 & 0.40 & 0.02 & 0.81 & \textbf{0.97} & 0.67 \\
\bottomrule
\end{tabular}
\end{table}

\subsection{Adaptive PatLR: Additional Analysis}
\label{app:adaptive_details}

\subsubsection{Selection, Identification and Cost on Planted Matrices}
\label{app:adaptive_oracle}

\paragraph{Setup.}
Each of the nine masks of the default vocabulary is planted at
$n{=}200$ with the generator of Figure~\ref{fig:denoise} (i.i.d.\
Gaussian values of scale $2$ on the mask and $0.15$ off it), either
alone (\emph{pattern only}) or plus a rank-3 PSD component of equal
energy (\emph{pattern + rank-3}); the matrix is scrambled by a random
permutation and symmetric Gaussian noise is added at SNR $4$ and
SNR $1$ ($10$ and $20$ seeds; $540$ instances).
The selector's vocabulary has $12$ masks: the nine planted families and
three gallery masks (parity checkerboard, anti-diagonal band of width
$10$, upper triangular). The parity checkerboard is a row/column-permuted
copy of the two-block mask (reorder even and odd indices), so the
density-normalized score cannot and should not separate the two; it is
counted with block $k{=}2$ throughout.
Methods: Oracle PatLR (rank $3$, planted mask; $12$ iterations, $3$
restarts); Adaptive PatLR (Algorithm~\ref{alg:adaptive}, $N{=}80$,
$\ell{=}3$, rank $3$); brute force (every mask fitted by the LAP phase
of Alternating LAP, $15$ iterations from $3$ restarts, i.e.\ the
search's LAP-only evaluation at a full budget per mask; the mask with
the lowest density-normalized score kept, then PatLR with it); SVD at
the rank with the highest DQ. DQ is
measured against the clean matrix in its planted order. Times are
single-core seconds on an Apple M3 (six instances at SNR~$1$).

\paragraph{Results (Table~\ref{tab:adaptive_selection},
Figure~\ref{fig:adaptive_main}).}
Across the $540$ instances the ratio
$\mathrm{DQ}_{\text{adapt}}/\mathrm{DQ}_{\text{oracle}}$ has median
$1.00$, $5$th percentile $0.92$ and minimum $0.76$ (a band $w{=}4$
instance at SNR~$1$, planted alone, for which block+band was
selected); it exceeds $1$ in a third of the instances. SVD at its best
rank averages DQ $0.29$ against the oracle's $0.68$ at SNR~$1$ and
$0.75$ (the noisy input itself) against $0.87$ at SNR~$4$. Brute force
has median ratio $1.00$ and minimum $0.74$; its DQ is within $0.01$ of
the search's in every (family, generator, SNR) cell, and its selection
frequencies agree with the search's within $10\%$ in $29$ of the $36$
cells, the larger differences being between masks of equal DQ (block
$k{=}3$ against block+band; on block $k{=}2$ with the rank-3 component
at SNR~$1$, block $k{=}6$ against band $w{=}4$). It takes $3.7$\,s
against $2.9$\,s for the search and $2.5$\,s for the oracle; its cost
grows with $|\mathcal{V}|$, whereas after the initial vocabulary
screening the search uses a fixed budget of $N$ iterations rather than
fully fitting every mask. Fitting every mask with the full
solver (exchange phase included) before comparing scores is not better:
the wide band then attains the lowest normalized score more often, and
on block $k{=}3$, arrowhead and block+band with the rank-3 component at
SNR~$1$ the selected mask has a lower DQ ($0.53$, $0.79$, $0.53$
against $0.61$, $0.85$, $0.62$), the exact minimizer of the heuristic
score not being the best-reconstructing mask
(Appendix~\ref{app:selection}). On six real
matrices (the feature correlations of Iris, Wine, Breast Cancer and
Digits, the karate club and Les Mis\'erables), where no planted mask
exists, the adaptively selected mask matches the best fixed mask of the
vocabulary in explained variance within $0.025$ in $24$ of $29$
(matrix, rank) cells, the exceptions being a sparser choice at rank
$\le3$ on Digits and rank $\le2$ on Les Mis\'erables.
Identification (Figure~\ref{fig:adaptive_main}a,b and the last
columns of Table~\ref{tab:adaptive_selection}) splits by the geometry of
the vocabulary. Arrowhead, block+band and the block templates $k{=}4$
and $k{=}6$ are identified in nearly every instance, although block
$k{=}4$ fits inside block $k{=}2$ and block $k{=}6$ inside block $k{=}3$:
under noise the denser container's smaller residual does not pay back
its density penalty. The few misses go to a neighbouring block size and
all occur at SNR~$1$. Block $k{=}2$ is identified
whenever the pattern is planted alone and at SNR~$4$ with the rank-3
component; at SNR~$1$ with the rank-3 component the selector prefers a
sparser mask (block $k{=}6$ or band $w{=}4$) at the oracle's DQ, the
low-rank term absorbing the two dense blocks so that the density penalty
of the half-dense mask is no longer paid back. The nested families go
to a larger mask in most instances (narrow bands to the wide band;
block $k{=}3$ to block+band, whose blocks cover those of block $k{=}3$
up to two rows), within $0.04$ of the oracle's DQ. The remaining
non-exact choices are the thin bands without a low-rank part: their
order is recovered only partially at this budget, the planted mask then
copies a misaligned band, and the selector prefers wider masks (block
$k{=}6$, block+band), which tolerate the misalignment; at SNR~$4$ they
reach a higher DQ than the planted mask and at SNR~$1$ a DQ
$0.02$--$0.07$ below it. With the rank-3 component the wide band is
identified in most instances.

\begin{table}[h]
\caption{Adaptive pattern selection on planted matrices ($n{=}200$).
  DQ of Oracle PatLR (planted mask), Adaptive PatLR and SVD at its best
  rank, and the share of instances (\%) in which Adaptive PatLR selects
  the planted mask (second number where present: a mask containing the
  planted one, a wider band for the bands and block $k{=}3$ for block
  $k{=}6$, or block+band for block $k{=}3$, whose blocks it covers up to
  two rows). At SNR~$4$ the noisy input itself has
  $\mathrm{DQ}=1-1/\mathrm{SNR}=0.75$. $10$
  seeds at SNR~$4$, $20$ at SNR~$1$.}
\label{tab:adaptive_selection}
\centering
\footnotesize
\setlength{\tabcolsep}{4pt}
\begin{tabular}{l rrrr rrrr}
\toprule
 & \multicolumn{4}{c}{SNR 4} & \multicolumn{4}{c}{SNR 1} \\
\cmidrule(lr){2-5}\cmidrule(lr){6-9}
planted mask & Oracle & Adaptive & SVD & exact\,/\,cont. & Oracle & Adaptive & SVD & exact\,/\,cont. \\
\midrule
\multicolumn{9}{l}{\emph{pattern only}}\\
block $k{=}2$ & 0.86 & 0.86 & 0.75 & 100 & 0.44 & 0.46 & 0.01 & 100 \\
block $k{=}3$ & 0.89 & 0.89 & 0.75 & 40\,/\,60 & 0.60 & 0.60 & 0.01 & 45\,/\,55 \\
block $k{=}4$ & 0.91 & 0.91 & 0.75 & 100 & 0.68 & 0.68 & 0.02 & 100 \\
block $k{=}6$ & 0.92 & 0.92 & 0.75 & 100 & 0.75 & 0.74 & 0.03 & 95\,/\,5 \\
band $w{=}1$ & 0.77 & 0.78 & 0.75 & 0\,/\,40 & 0.69 & 0.67 & 0.12 & 5\,/\,95 \\
band $w{=}2$ & 0.77 & 0.79 & 0.75 & 0\,/\,60 & 0.67 & 0.63 & 0.09 & 10\,/\,55 \\
band $w{=}4$ & 0.76 & 0.79 & 0.75 & 0 & 0.63 & 0.56 & 0.05 & 10 \\
arrowhead & 0.78 & 0.78 & 0.75 & 100 & 0.73 & 0.72 & 0.28 & 100 \\
block+band & 0.89 & 0.89 & 0.75 & 100 & 0.60 & 0.60 & 0.01 & 100 \\
\midrule
\multicolumn{9}{l}{\emph{pattern + rank-3 component}}\\
block $k{=}2$ & 0.86 & 0.80 & 0.75 & 100 & 0.46 & 0.47 & 0.49 & 0 \\
block $k{=}3$ & 0.90 & 0.90 & 0.75 & 10\,/\,90 & 0.63 & 0.61 & 0.49 & 0\,/\,70 \\
block $k{=}4$ & 0.92 & 0.92 & 0.75 & 100 & 0.71 & 0.70 & 0.49 & 90 \\
block $k{=}6$ & 0.94 & 0.94 & 0.75 & 100 & 0.79 & 0.79 & 0.50 & 100 \\
band $w{=}1$ & 0.88 & 0.88 & 0.75 & 0\,/\,100 & 0.81 & 0.77 & 0.51 & 10\,/\,90 \\
band $w{=}2$ & 0.87 & 0.88 & 0.75 & 10\,/\,90 & 0.79 & 0.78 & 0.51 & 40\,/\,55 \\
band $w{=}4$ & 0.86 & 0.85 & 0.75 & 60 & 0.75 & 0.74 & 0.50 & 90 \\
arrowhead & 0.89 & 0.89 & 0.76 & 100 & 0.85 & 0.85 & 0.63 & 100 \\
block+band & 0.90 & 0.90 & 0.75 & 100 & 0.63 & 0.62 & 0.49 & 70 \\
\bottomrule
\end{tabular}
\end{table}

\subsubsection{Sensitivity to the Residual Rank}
\label{app:rank_sensitivity}

\begin{figure}[h]
  \centering
  \includegraphics[width=\linewidth]{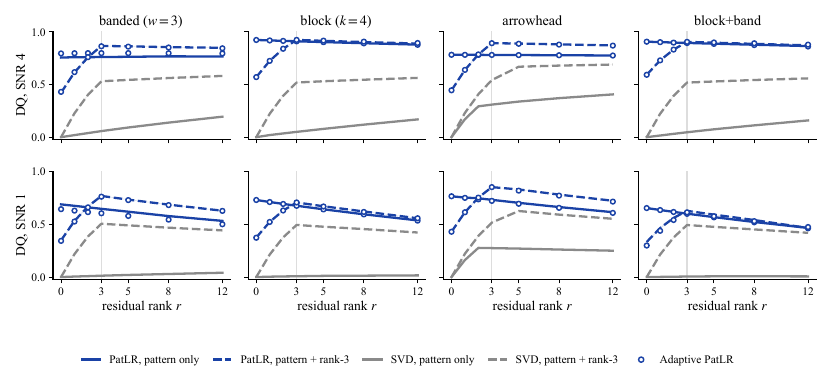}
  \caption{Denoising quality against the residual rank $r$ on the setup of
    Figure~\ref{fig:denoise} ($20$ seeds, $95\%$ intervals shaded):
    PatLR with the planted mask and SVD at the same rank, for the planted
    pattern alone (solid) and with a rank-3 component (dashed); markers
    are Adaptive PatLR. The vertical line marks the planted rank $3$.}
  \label{fig:rank_sensitivity}
\end{figure}

Figure~\ref{fig:rank_sensitivity} shows DQ against the residual rank $r$
on the setup of Figure~\ref{fig:denoise} ($n{=}200$, the planted pattern
alone and with a rank-3 component of equal energy, SNR $4$ and $1$,
$20$ seeds) for PatLR with the planted mask, SVD at the same rank and
Adaptive PatLR. With a rank-3 component in the data, DQ peaks at the
true rank for PatLR and declines slowly beyond it, while SVD needs a
higher rank to reach a lower plateau. With the pattern alone every unit
of rank costs DQ at SNR~$1$ (the low-rank term fits noise) and little at
SNR~$4$. Adaptive PatLR follows PatLR within $0.05$ at every rank; the
largest differences are on the band planted alone, where it selects a
wider mask (Appendix~\ref{app:adaptive_oracle}) and lies $0.04$ above
PatLR at SNR~$4$ and $0.04$--$0.05$ below it at SNR~$1$. The rank is a
budget to be chosen, not a free improvement.

\subsection{Ordering and Block-Structure Benchmarks: Full Results}
\label{app:specialists}

This appendix gives the full grids behind Figure~\ref{fig:tasks} and
Sections~\ref{sec:exp_ordering}--\ref{sec:exp_blocks}, and two ordering
geometries reported only here, circular seriation and nestedness
(Appendix~\ref{app:c1p}). Counting the cells of each grid by the sign of the paired difference to the named
reference (positive / interval containing zero / negative): latent
ordering $18/2/4$ against spectral and $9/10/5$ against convex seriation
($24$ cells); consecutive ones $4/0/0$; circular seriation $2/2/0$;
sparse SBM $8/1/3$ against the Bethe Hessian ($12$ cells, the negative
ones at $\lambda{=}1.3$); weighted partition $4/1/0$ against
spectral clustering; band-in-block $16/0/0$ for the refining pass
against the pipeline. In the tables, ``seeds $+$/$-$'' counts the seeds
on which the paired difference is positive / negative.

\subsubsection{Latent-order recovery}
\label{app:seriation}

\textbf{Setup.} $B_{ij}=\exp(-(i-j)^2/2\ell^2)$ is a Robinson matrix with
correlation length $\ell$; it is scrambled by a uniformly random
permutation $p$ and symmetric Gaussian noise of standard deviation
$\sigma$ is added. The target is the banded mask of width $w=\ell$. The
metric is $|\tau|$, the absolute Kendall correlation between the
recovered and the planted latent positions (absolute because reversing
an order preserves banded structure). Baselines: spectral seriation
(Fiedler order of the Laplacian of $|A|$~\citep{atkins1998spectral});
convex seriation~\citep{fogel2013seriation}, implemented as Frank--Wolfe
on the 2-SUM objective over the permutahedron with the concave
vertex-pulling term $-\mu\,\mathrm{var}(y)$, the penalty weight
$\mu\in\{0.1,0.3,0.5,1,2\}$ and the start (spectral or two random)
selected by the 2-SUM value itself, never by $\tau$ (on noiseless
inputs it recovers the planted order exactly); QAP-FAQ with three
starts; and a single run of Alternating LAP initialized from the
spectral order (AltLAP-r1; five restarts change no conclusion).
Grid: $\ell\in\{6,12,25\}$, $n\in\{100,300\}$,
$\sigma\in\{0.05,0.1,0.2,0.3\}$.

\textbf{Results.} Table~\ref{tab:seriation_full} lists the $24$ cells
behind the counts of Section~\ref{sec:exp_ordering}. The four negative
cells are $\ell=w=25$ at $n=100$, where the band covers $44.5\%$ of the
matrix; the two ties are $\ell=6$, $n=300$ at $\sigma\ge0.2$, where the
correlation length is short relative to $n$, the global order is weakly
identified and all methods are poor ($\tau$ between $0.25$ and $0.34$).
Runtimes at $n=300$ (one EPYC core): spectral $0.002$\,s, convex
$\approx 0.25$\,s, FAQ $0.6$--$1.1$\,s, AltLAP-r1 $0.1$--$0.2$\,s, of
which the exchange phase is the larger part.

\begin{table}[h]
\caption{Latent-order recovery, all $24$ cells: $|\tau|$ against the planted
  order (mean over $20$ seeds; best of spectral, convex and ours in bold)
  and paired differences of AltLAP-r1 against spectral and convex
  seriation. FAQ optimizes the same pattern objective from its default
  starts.}
\label{tab:seriation_full}
\centering
\scriptsize
\setlength{\tabcolsep}{2.5pt}
\begin{tabular}{l l r r r r l l}
\toprule
$\ell$ & $n$, $\sigma$ & Spectral & Convex & FAQ & \textbf{AltLAP-r1} & $\Delta$ vs spectral [CI], seeds $+$/$-$ & $\Delta$ vs convex [CI], seeds $+$/$-$ \\
\midrule
6 & 100, 0.05 & 0.981 & 0.992 & 0.269 & \textbf{1.000} & $+0.019$ [$+0.017,+0.021$], 20/0 & $+0.008$ [$+0.007,+0.008$], 20/0 \\
6 & 100, 0.1 & 0.943 & 0.975 & 0.357 & \textbf{0.986} & $+0.043$ [$+0.036,+0.051$], 20/0 & $+0.011$ [$-0.001,+0.021$], 17/3 \\
6 & 100, 0.2 & 0.824 & \textbf{0.905} & 0.293 & 0.861 & $+0.037$ [$+0.020,+0.053$], 17/3 & $-0.044$ [$-0.084,-0.003$], 9/11 \\
6 & 100, 0.3 & 0.673 & 0.700 & 0.225 & \textbf{0.722} & $+0.049$ [$+0.029,+0.072$], 18/2 & $+0.022$ [$-0.007,+0.047$], 15/5 \\
6 & 300, 0.05 & 0.811 & \textbf{0.861} & 0.159 & 0.828 & $+0.017$ [$+0.011,+0.022$], 18/2 & $-0.033$ [$-0.083,+0.015$], 8/12 \\
6 & 300, 0.1 & 0.543 & 0.549 & 0.165 & \textbf{0.554} & $+0.011$ [$+0.001,+0.020$], 14/6 & $+0.004$ [$-0.072,+0.078$], 8/12 \\
6 & 300, 0.2 & 0.336 & 0.334 & 0.117 & \textbf{0.337} & $+0.002$ [$-0.006,+0.009$], 9/11 & $+0.003$ [$-0.088,+0.080$], 10/10 \\
6 & 300, 0.3 & 0.250 & \textbf{0.284} & 0.109 & 0.256 & $+0.006$ [$-0.006,+0.017$], 12/8 & $-0.029$ [$-0.097,+0.033$], 10/10 \\
\midrule
12 & 100, 0.05 & 0.995 & \textbf{0.997} & 0.634 & \textbf{0.997} & $+0.002$ [$+0.002,+0.003$], 17/1 & $-0.000$ [$-0.001,+0.001$], 9/7 \\
12 & 100, 0.1 & 0.983 & 0.988 & 0.592 & \textbf{0.991} & $+0.008$ [$+0.007,+0.009$], 20/0 & $+0.003$ [$+0.002,+0.004$], 18/2 \\
12 & 100, 0.2 & 0.954 & 0.966 & 0.662 & \textbf{0.981} & $+0.027$ [$+0.023,+0.030$], 20/0 & $+0.015$ [$+0.013,+0.017$], 20/0 \\
12 & 100, 0.3 & 0.921 & 0.939 & 0.658 & \textbf{0.972} & $+0.051$ [$+0.045,+0.058$], 20/0 & $+0.033$ [$+0.030,+0.036$], 20/0 \\
12 & 300, 0.05 & 0.980 & 0.990 & 0.241 & \textbf{0.999} & $+0.019$ [$+0.018,+0.020$], 20/0 & $+0.010$ [$+0.009,+0.010$], 20/0 \\
12 & 300, 0.1 & 0.943 & 0.966 & 0.326 & \textbf{0.977} & $+0.034$ [$+0.026,+0.041$], 19/1 & $+0.011$ [$-0.001,+0.020$], 16/4 \\
12 & 300, 0.2 & 0.839 & 0.861 & 0.264 & \textbf{0.889} & $+0.050$ [$+0.032,+0.065$], 18/2 & $+0.028$ [$-0.010,+0.052$], 17/3 \\
12 & 300, 0.3 & 0.709 & \textbf{0.787} & 0.169 & 0.750 & $+0.041$ [$+0.018,+0.062$], 17/3 & $-0.037$ [$-0.133,+0.026$], 14/6 \\
\midrule
25 & 100, 0.05 & \textbf{0.999} & \textbf{0.999} & 0.990 & 0.991 & $-0.008$ [$-0.008,-0.007$], 0/20 & $-0.008$ [$-0.008,-0.007$], 0/20 \\
25 & 100, 0.1 & \textbf{0.993} & \textbf{0.993} & 0.979 & 0.980 & $-0.012$ [$-0.013,-0.011$], 0/20 & $-0.012$ [$-0.013,-0.011$], 0/20 \\
25 & 100, 0.2 & \textbf{0.978} & \textbf{0.978} & 0.960 & 0.962 & $-0.016$ [$-0.018,-0.013$], 0/20 & $-0.016$ [$-0.018,-0.014$], 0/20 \\
25 & 100, 0.3 & 0.960 & \textbf{0.961} & 0.942 & 0.945 & $-0.015$ [$-0.018,-0.012$], 0/20 & $-0.016$ [$-0.018,-0.013$], 0/20 \\
25 & 300, 0.05 & 0.992 & 0.996 & 0.497 & \textbf{0.997} & $+0.005$ [$+0.004,+0.005$], 20/0 & $+0.001$ [$+0.001,+0.002$], 20/0 \\
25 & 300, 0.1 & 0.982 & 0.988 & 0.560 & \textbf{0.993} & $+0.012$ [$+0.011,+0.013$], 20/0 & $+0.006$ [$+0.005,+0.006$], 20/0 \\
25 & 300, 0.2 & 0.959 & 0.971 & 0.516 & \textbf{0.988} & $+0.029$ [$+0.027,+0.031$], 20/0 & $+0.017$ [$+0.017,+0.018$], 20/0 \\
25 & 300, 0.3 & 0.933 & 0.950 & 0.492 & \textbf{0.983} & $+0.050$ [$+0.045,+0.054$], 20/0 & $+0.033$ [$+0.031,+0.034$], 20/0 \\
\bottomrule
\end{tabular}
\end{table}

\paragraph{Mask width.}
The band width given to the solver is the correlation length of the
generator. Table~\ref{tab:seriation_width} repeats AltLAP-r1 with the
width halved and doubled: a factor-two error in either direction
changes $\tau$ by less than $0.01$ in most cells and by up to $0.06$
(doubled width at $\ell{=}12$, $n{=}100$, $\sigma{=}0.3$), where the lead
over convex seriation is lost; a four-fold overestimate ($w{=}25$ at
$\ell{=}6$, $n{=}100$) falls below the spectral order. Choosing the
width by the density-normalized score of Section~\ref{sec:adaptive} is
not an option on these smooth kernels: the score never selects a width
below the correlation length and picks a wider candidate whenever one
is available, so $w$ is a user input here.

\begin{table}[h]
\caption{Latent-order recovery with the band width halved and doubled
  ($|\tau|$, mean over $20$ seeds; the paper's width $w{=}\ell$ in bold).
  At $n{=}100$ the doubled width $50$ is omitted.}
\label{tab:seriation_width}
\centering
\footnotesize
\setlength{\tabcolsep}{4pt}
\begin{tabular}{l l r r r r r}
\toprule
$\ell$ & $n$, $\sigma$ & Spectral & Convex & $w{=}\ell/2$ & $w{=}\ell$ & $w{=}2\ell$ \\
\midrule
6 & 100, 0.1 & 0.943 & 0.975 & 0.981 & \textbf{0.986} & 0.983 \\
6 & 100, 0.3 & 0.673 & 0.700 & 0.710 & \textbf{0.722} & 0.726 \\
6 & 300, 0.1 & 0.543 & 0.549 & 0.547 & \textbf{0.554} & 0.553 \\
6 & 300, 0.3 & 0.250 & 0.284 & 0.255 & \textbf{0.256} & 0.261 \\
\midrule
12 & 100, 0.1 & 0.983 & 0.988 & 0.989 & \textbf{0.991} & 0.972 \\
12 & 100, 0.3 & 0.921 & 0.939 & 0.967 & \textbf{0.972} & 0.913 \\
12 & 300, 0.1 & 0.943 & 0.966 & 0.975 & \textbf{0.977} & 0.971 \\
12 & 300, 0.3 & 0.709 & 0.787 & 0.748 & \textbf{0.750} & 0.759 \\
\midrule
25 & 100, 0.1 & 0.993 & 0.993 & 0.977 & \textbf{0.980} & -- \\
25 & 100, 0.3 & 0.960 & 0.961 & 0.939 & \textbf{0.945} & -- \\
25 & 300, 0.1 & 0.982 & 0.988 & 0.992 & \textbf{0.993} & 0.986 \\
25 & 300, 0.3 & 0.933 & 0.950 & 0.978 & \textbf{0.983} & 0.958 \\
\bottomrule
\end{tabular}
\end{table}

\subsubsection{Consecutive ones, circular seriation, and nestedness}
\label{app:c1p}

\textbf{Consecutive ones (fixed-width staircase recovery).}
The consecutive-ones property (Section~\ref{sec:exp_ordering}) is
decided in linear time by PQ-trees~\citep{booth1976c1p}, but recognition
is brittle (one corrupted entry can destroy the property), so the applied
literature, from physical mapping to archaeological
seriation~\citep{kendall1971seriation}, orders the matrix with
heuristics: barycentre reordering, spectral~\citep{atkins1998spectral}
and convex~\citep{fogel2013seriation} seriation. Our target is the
staircase band $M_{ij}=\mathbf{1}[s_i\le j<s_i+w]$ with non-decreasing
starts $s_i$, fitted with independent row and column permutations; for a
matrix whose rows all carry $w$ ones, $A$ has the C1P iff $d_M(A)=0$ for
some mask of this family (ordering rows by the start of their interval
makes the starts non-decreasing). General C1P with varying interval
lengths is a larger class and would require a search over the start and
length profiles; what we test is the noisy recovery of the fixed-width
model, with the evenly spaced profile $s_i$ and its sensitivity below.
Instances: planted $60\times 60$ staircases with $w=12$ and independent
flip noise at rate $f$, $20$ seeds. Our solver runs from random restarts
only and keeps the candidate with the best mask fit, which uses no
labels. Metrics: $|\tau|$ of the recovered row and column orders against
the planted ones, and \emph{edits}, the minimum number of entries that
must be changed to make every row contiguous under the returned
ordering.

\begin{table}[h]
\caption{Noisy consecutive ones ($60\times 60$, $w=12$, $20$ seeds):
  row $|\tau|$ / column $|\tau|$ / edits. Controls use the identical
  protocol and restart budget with a random or a block mask of the same
  density.}
\label{tab:c1p}
\centering
\footnotesize
\setlength{\tabcolsep}{3pt}
\begin{tabular}{l c c c c}
\toprule
Method & $f{=}0.02$ & $f{=}0.05$ & $f{=}0.10$ & $f{=}0.15$ \\
\midrule
Sort by marginals & 0.05 / 0.04 / 582 & 0.05 / 0.05 / 638 & 0.06 / 0.06 / 734 & 0.07 / 0.06 / 837 \\
Barycentre reordering & 0.73 / 0.66 / 295 & 0.61 / 0.51 / 448 & 0.53 / 0.42 / 608 & 0.46 / 0.36 / 732 \\
Spectral seriation & 0.90 / 0.87 / 203 & 0.84 / 0.66 / 447 & 0.67 / 0.44 / 659 & 0.49 / 0.37 / 790 \\
Convex seriation & 0.92 / 0.92 / 163 & 0.85 / 0.85 / 312 & 0.71 / 0.69 / 550 & 0.50 / 0.53 / 718 \\
Spectral + 2-SUM local search & 0.92 / 0.88 / 181 & 0.85 / 0.83 / 332 & 0.71 / 0.61 / 570 & 0.49 / 0.53 / 721 \\
\textbf{Staircase mask, AltLAP (cold)} & \textbf{0.99 / 1.00 / 66} & \textbf{0.99 / 0.99 / 164} & \textbf{0.98 / 0.97 / 335} & \textbf{0.95 / 0.91 / 502} \\
Spectral, then AltLAP & 0.99 / 1.00 / 67 & 0.98 / 0.97 / 168 & 0.83 / 0.78 / 368 & 0.56 / 0.53 / 550 \\
\midrule
Control: random mask (cold) & 0.06 / 0.06 / 617 & 0.07 / 0.07 / 672 & 0.05 / 0.07 / 765 & 0.08 / 0.09 / 859 \\
Control: block mask (cold) & 0.28 / 0.25 / 322 & 0.31 / 0.29 / 402 & 0.34 / 0.31 / 537 & 0.24 / 0.21 / 666 \\
\bottomrule
\end{tabular}
\end{table}

Paired against convex seriation, the column $\tau$ of the staircase
fit is higher by $+0.075$ [$+0.063,+0.088$], $+0.136$ [$+0.118,+0.157$],
$+0.280$ [$+0.220,+0.358$] and $+0.377$ [$+0.300,+0.469$] at $f=0.02$,
$0.05$, $0.10$, $0.15$, and the edit count lower by $97$ [$89,106$],
$148$ [$137,160$], $215$ [$194,238$] and $216$ [$202,232$], on $20$ of
$20$ seeds at every level. An insertion local search on the 2-SUM
criterion needs $114$--$201$ more edits than the staircase fit on $20$ of
$20$ seeds, so the gain is not produced by refinement as such. Row
totals are constant by construction and column totals vary only near the
boundary, which is why sorting by marginals carries almost no
information. Runtimes: $0.03$\,s spectral, $0.18$\,s convex,
$0.3$--$0.5$\,s local search, $0.5$--$1$\,s ours.

Table~\ref{tab:c1p_robust} repeats the comparison with the shape of the
matrix, the start profile and the interval lengths varied. The fixed
staircase keeps
the higher $\tau$ and the lower edit count for rectangular matrices, for
an irregular start profile and for interval lengths jittered by up to
$\pm8$, with the smallest margins under the irregular profile and the
largest jitter. Re-estimating the mask from the data after the cold fit,
one interval per row (the interval-envelope pattern of
Figure~\ref{fig:gallery}), lowers the edit count slightly and restores
the row order under the irregular profile.

\begin{table}[h]
\caption{Noisy consecutive ones under changes of the generator ($20$
  seeds): column $|\tau|$ / edits for convex seriation, the fixed
  staircase mask and the mask re-estimated as one interval per row;
  ``noise'' is the number of entries flipped during corruption.}
\label{tab:c1p_robust}
\centering
\scriptsize
\setlength{\tabcolsep}{3pt}
\begin{tabular}{l r l l l r l l l}
\toprule
& \multicolumn{4}{c}{$f{=}0.05$} & \multicolumn{4}{c}{$f{=}0.15$} \\
\cmidrule(lr){2-5}\cmidrule(lr){6-9}
Setting & noise & convex & fixed mask & re-estimated & noise & convex & fixed mask & re-estimated \\
\midrule
$60\times60$, $w{=}12$ (as Table~\ref{tab:c1p}) & 179 & 0.86 / 317 & 1.00 / 166 & 1.00 / 165 & 540 & 0.51 / 716 & 0.92 / 496 & 0.93 / 494 \\
$40\times80$, $w{=}16$ & 161 & 0.80 / 328 & 0.98 / 155 & 0.98 / 153 & 483 & 0.49 / 669 & 0.77 / 461 & 0.77 / 451 \\
$80\times40$, $w{=}8$ & 161 & 0.88 / 247 & 1.00 / 145 & 1.00 / 145 & 483 & 0.61 / 574 & 0.97 / 430 & 0.97 / 430 \\
irregular start profile & 179 & 0.71 / 334 & 0.85 / 188 & 0.85 / 184 & 540 & 0.51 / 695 & 0.80 / 510 & 0.80 / 507 \\
interval lengths $12\pm4$ & 179 & 0.86 / 304 & 0.96 / 189 & 0.97 / 182 & 540 & 0.53 / 704 & 0.86 / 515 & 0.85 / 510 \\
interval lengths $12\pm8$ & 179 & 0.87 / 291 & 0.91 / 205 & 0.93 / 193 & 540 & 0.65 / 664 & 0.78 / 527 & 0.80 / 519 \\
\bottomrule
\end{tabular}
\end{table}

\textbf{Circular seriation.} A latent order can be cyclic rather than
linear: similarities between points on a closed curve are banded modulo
$n$, and spectral seriation needs a dedicated circular embedding for
this case~\citep{recanati2018circular}. The question is whether the same
solver covers the cyclic geometry by changing only the mask. For $n=120$
points on a circle with a decaying kernel and Gaussian noise, the target
is the cyclic band mask of width $10$ and the reference is the spectral
circular embedding; the metric is circular Kendall $\tau$ up to rotation
and reflection ($20$ seeds). Ours is the better of $128$ cold restarts
and a coarse-to-fine run, chosen by the objective. In
Table~\ref{tab:circular} both methods recover the order at low noise
(the difference is within $\pm0.01$); at noise $0.5$ and $0.7$ the cyclic
band is ahead ($+0.035$ and $+0.103$, on $13$ and $18$ of $20$ seeds),
and at noise $1.0$ neither recovers the order and the interval contains
zero. The linear spectral order is far
below throughout, so the cyclic mask, not a generic ordering, carries
the result; the gain over the circular embedding is concentrated at
moderate noise.

\begin{table}[h]
\caption{Circular seriation ($n{=}120$, cyclic band $w{=}10$): circular
  Kendall $\tau$, mean over $20$ seeds, and the paired difference of ours
  against the circular spectral embedding.}
\label{tab:circular}
\centering
\scriptsize
\setlength{\tabcolsep}{3.5pt}
\begin{tabular}{l r r r r l r}
\toprule
Noise & Spectral (linear) & Circ.\ spectral & Circ.\ spectral $\to$ AltLAP & \textbf{AltLAP, cyclic (cold)} & $\Delta$ vs circ.\ spectral [CI] & seeds $+$/$-$ \\
\midrule
0.3 & 0.372 & 0.963 & \textbf{0.965} & 0.958 & $-0.005$ [$-0.015,+0.006$] & 6/13 \\
0.5 & 0.393 & 0.903 & 0.934 & \textbf{0.938} & $+0.035$ [$+0.014,+0.055$] & 13/7 \\
0.7 & 0.332 & 0.782 & 0.879 & \textbf{0.885} & $+0.103$ [$+0.065,+0.138$] & 18/2 \\
1.0 & 0.203 & 0.431 & \textbf{0.471} & 0.469 & $+0.039$ [$-0.072,+0.142$] & 11/9 \\
\bottomrule
\end{tabular}
\end{table}

\textbf{Nestedness: a boundary case.} In a nested (Ferrers) matrix
the ones of each row contain those of every row with a smaller degree,
the structure of species-by-site and similar ecological
matrices~\citep{brualdi1999nested}. The question here is whether a
general reordering objective adds anything where a one-pass specialized
method exists: sorting rows and columns by their marginals recovers a
noiseless nested matrix exactly and, although under flip noise it need
not minimize our objective, it is the reference. For planted
$60\times 80$ Ferrers matrices with flip rate $f$, the two-sided AltLAP
uses the nested mask with its profile estimated from the row degrees (no
oracle information); Table~\ref{tab:nested} gives the column $|\tau|$
and the flips to a nested matrix ($20$ seeds). The mask fit reproduces
the degree order at $f\le0.02$ and falls below it from $f=0.05$ on, by
$0.06$ in $\tau$ at $f=0.15$ with a similar flip count; warm-starting
from the degree order does not help, and a larger restart budget does
not change the values. Where degree sorting already recovers the
structure, the mask objective adds nothing; this is the boundary case
noted in Section~\ref{sec:exp_ordering}.

\begin{table}[h]
\caption{Nestedness ($60\times80$ planted Ferrers matrices, flip rate
  $f$, $20$ seeds): column $|\tau|$ / flips to a nested matrix under the
  returned order (lower is better).}
\label{tab:nested}
\centering
\footnotesize
\setlength{\tabcolsep}{4pt}
\begin{tabular}{l l l l l}
\toprule
$f$ & Degree sort & Spectral & AltLAP, nested mask & Degree sort $\to$ AltLAP \\
\midrule
0.0 & 0.994 / 0 & 0.586 / 1277 & 0.994 / 0 & 0.994 / 0 \\
0.02 & 0.967 / 134 & 0.573 / 1382 & 0.965 / 129 & 0.963 / 131 \\
0.05 & 0.942 / 301 & 0.544 / 1658 & 0.932 / 299 & 0.925 / 301 \\
0.1 & 0.906 / 574 & 0.548 / 1712 & 0.881 / 586 & 0.862 / 587 \\
0.15 & 0.876 / 819 & 0.507 / 1931 & 0.817 / 843 & 0.806 / 842 \\
\bottomrule
\end{tabular}
\end{table}

\subsubsection{Block models against the state of the art}
\label{app:blockmodels}

\textbf{Sparse binary SBM at the detectability scale.}
Symmetric SBM with $k=4$ equal blocks, $n=500$,
$c_{\mathrm{in}} = c + (k-1)\lambda\sqrt{c}$ and
$c_{\mathrm{out}} = c - \lambda\sqrt{c}$, so $\lambda=1$ is the asymptotic
(Kesten--Stigum) detectability threshold~\citep{decelle2011asymptotic}; mean degree
$c\in\{8,16,32\}$, $20$ seeds. Panel: Bethe Hessian $H(r)=(r^2-1)I-rA+D$
with $r=\sqrt{c}$~\citep{saade2014bethe}; the non-backtracking operator
in Ihara--Bass form~\citep{krzakala2013spectral}; regularized spectral
clustering~\citep{qin2013regularized}; Leiden~\citep{traag2019leiden},
which estimates $k$ itself; the block mask fitted by Alternating LAP
from eight random starts, the LAP and exchange phases interleaved per
start (three rounds) and the best start kept by the criterion
(``ours''); and the same run started from the Bethe-Hessian partition.
All methods except Leiden receive the true $k$; our mask also encodes
equal block sizes, which the planted model satisfies and the spectral
methods do not use. On sparse $\{0,1\}$ graphs the LAP phase alone
stalls early (ARI $0.56$ at $\lambda=1.6$, $c=16$, with the exchange
phase disabled, $32$ starts); with the exchange phase the fit reaches
$0.82$.

\begin{table}[h]
\caption{Sparse SBM, $k=4$, $n=500$, mean ARI over $20$ seeds. ``Ours''
  is Alternating LAP from random starts; ``swap only'' is a
  Kernighan--Lin-style exchange search alone from random balanced
  partitions (same criterion and block sizes); the paired difference is
  against the Bethe Hessian, with the best panel method per row in
  bold.}
\label{tab:sbm}
\centering
\scriptsize
\setlength{\tabcolsep}{3pt}
\begin{tabular}{l l r r r r r r r l r}
\toprule
$\lambda$ & $c$ & Bethe H. & Non-backtr. & Reg.\ sp. & Leiden & Swap only & \textbf{Ours} & BH $\to$ ours & $\Delta$ vs BH [CI] & seeds $+$/$-$ \\
\midrule
1.3 & 8 & 0.42 & 0.38 & 0.41 & 0.13 & 0.05 & 0.15 & \textbf{0.47} & $-0.272$ [$-0.296,-0.248$] & 0/20 \\
1.3 & 16 & 0.43 & 0.39 & 0.35 & 0.15 & 0.08 & 0.22 & \textbf{0.51} & $-0.208$ [$-0.249,-0.167$] & 1/19 \\
1.3 & 32 & 0.47 & 0.39 & 0.40 & 0.13 & 0.11 & 0.35 & \textbf{0.57} & $-0.116$ [$-0.168,-0.059$] & 2/18 \\
1.6 & 8 & 0.73 & 0.72 & 0.73 & 0.61 & 0.13 & 0.70 & \textbf{0.78} & $-0.034$ [$-0.083,+0.009$] & 11/9 \\
1.6 & 16 & 0.74 & 0.74 & 0.75 & 0.71 & 0.32 & \textbf{0.82} & \textbf{0.82} & $+0.080$ [$+0.069,+0.093$] & 20/0 \\
1.6 & 32 & 0.78 & 0.78 & 0.78 & 0.83 & 0.58 & \textbf{0.86} & \textbf{0.86} & $+0.079$ [$+0.063,+0.095$] & 18/2 \\
2.0 & 8 & 0.92 & 0.91 & 0.92 & 0.94 & 0.79 & 0.94 & \textbf{0.95} & $+0.022$ [$+0.013,+0.030$] & 18/2 \\
2.0 & 16 & 0.92 & 0.92 & 0.92 & \textbf{0.95} & 0.94 & \textbf{0.95} & \textbf{0.95} & $+0.028$ [$+0.021,+0.036$] & 20/0 \\
2.0 & 32 & 0.94 & 0.94 & 0.94 & \textbf{0.97} & 0.96 & \textbf{0.97} & \textbf{0.97} & $+0.027$ [$+0.019,+0.034$] & 19/1 \\
2.5 & 8 & 0.99 & 0.98 & 0.99 & \textbf{0.99} & 0.98 & \textbf{1.00} & \textbf{1.00} & $+0.009$ [$+0.006,+0.012$] & 18/2 \\
2.5 & 16 & 0.99 & 0.99 & 0.99 & \textbf{0.99} & \textbf{0.99} & \textbf{1.00} & \textbf{1.00} & $+0.007$ [$+0.004,+0.010$] & 17/1 \\
2.5 & 32 & 0.99 & 0.99 & 0.99 & \textbf{1.00} & \textbf{1.00} & \textbf{1.00} & \textbf{1.00} & $+0.009$ [$+0.006,+0.013$] & 18/0 \\
\bottomrule
\end{tabular}
\end{table}

With equal block sizes the number of within-block pairs is fixed, so
maximizing the within-block edge count, which is our criterion on a
$\{0,1\}$ matrix, is the maximum-likelihood objective for the planted partition
at any $n$ and any $p>q$ (for $p<q$ the sign reverses; for $k=2$ it is minimum bisection;
\citealp{abbe2016exact}). The crossover of Section~\ref{sec:exp_blocks}
moves with the mean degree (Table~\ref{tab:sbm}). Below it the
criterion values are $0.33$ (ours) against $0.24$ (Bethe-Hessian
partition) in error reduction at $c=16$, $\lambda=1.3$, with the ARIs in
the opposite order; the gap between the MAP and the Bayes-optimal
(belief-propagation) estimator near the threshold is the reason belief
propagation and its spectral linearizations were
introduced~\citep{decelle2011asymptotic,krzakala2013spectral,saade2014bethe}.
We do not establish inconsistency of the criterion at this $n$, nor that
the search reaches its global optimum. Refining the Bethe-Hessian
partition with our criterion costs $0.3$--$0.5$\,s on top of its
$0.04$\,s.

\textbf{Weighted planted partition.}
$A=\mu\cdot\mathbf{1}[\text{same block}]+E$ with $\mu>0$ and $E$
symmetric with i.i.d.\ $\mathcal{N}(0,\sigma^2)$ upper-triangular entries,
$\sigma=1$, $n=300$, $k=4$ equal blocks, $20$ seeds. This is the regime
of the block experiments in the paper, where minimizing the \emph{signed}
off-mask mass is the maximum-likelihood criterion for the partition.
Spectral clustering on the weighted matrix is the natural baseline; the
graph methods are defined for graphs and receive the matrix sparsified
at its strongest $10\%$ of entries. The squared loss rewards large
negative off-mask entries as if they were structure, which is why it
fails on this generator, whose blocks differ in mean but not in variance
(the ablation column); the
linear loss, which on a $\{0,1\}$ matrix coincides with the squared one,
is the likelihood here. A Kernighan--Lin-style swap search on the same
linear criterion tracks the cold run from $\mu{=}0.4$ on and lies
$0.02$--$0.03$ below it at the two weakest signals, so on these
dense matrices the criterion, not the particular optimizer, carries the
result.

\begin{table}[h]
\caption{Weighted planted partition ($n=300$, $k=4$, $\sigma=1$), mean ARI
  over $20$ seeds; paired difference of ours (AltLAP-lin, cold start)
  against spectral clustering. ``Squared'' is AltLAP on the squared loss
  of Definition~\ref{def:pattern_dist} (the ablation row); ``KL (lin.)''
  is a Kernighan--Lin-style swap search on the same linear criterion from
  random balanced starts, an alternative optimizer of our objective.}
\label{tab:weighted}
\centering
\scriptsize
\setlength{\tabcolsep}{1.5pt}
\begin{tabular}{l r r r r r r r r r l r}
\toprule
$\mu$ & Spectral & Reg.\ sp. & Bethe H. & Non-b. & Leiden & Squared & KL (lin.) & \textbf{AltLAP-lin} & Spec.\ $\to$ ours & $\Delta$ vs spectral [CI] & seeds $+$/$-$ \\
\midrule
0.30 & 0.403 & 0.028 & 0.064 & 0.052 & 0.033 & 0.003 & 0.324 & 0.356 & \textbf{0.556} & $-0.046$ [$-0.111,+0.024$] & 7/13 \\
0.35 & 0.680 & 0.073 & 0.119 & 0.102 & 0.054 & 0.004 & 0.807 & 0.827 & \textbf{0.838} & $+0.147$ [$+0.126,+0.169$] & 20/0 \\
0.40 & 0.850 & 0.154 & 0.218 & 0.179 & 0.092 & 0.006 & 0.927 & 0.932 & \textbf{0.932} & $+0.081$ [$+0.058,+0.105$] & 19/1 \\
0.50 & 0.975 & 0.482 & 0.560 & 0.445 & 0.354 & 0.018 & \textbf{0.994} & \textbf{0.995} & \textbf{0.995} & $+0.020$ [$+0.014,+0.027$] & 18/1 \\
0.60 & 0.996 & 0.821 & 0.814 & 0.813 & 0.866 & 0.050 & \textbf{1.000} & \textbf{1.000} & \textbf{1.000} & $+0.004$ [$+0.002,+0.008$] & 7/0 \\
\bottomrule
\end{tabular}
\end{table}

At the weakest signal the cold run is within the interval of spectral
clustering and the spectral-started run is the entry that pays:
$+0.154$ [$+0.118,+0.188$] and $+0.159$ [$+0.135,+0.182$] over
spectral clustering at $\mu=0.30$ and $0.35$.

\paragraph{Block sizes.}
The block mask fixes the block sizes, which above are the planted (equal)
ones. Table~\ref{tab:block_sizes} plants blocks of sizes $1{:}1{:}2{:}2$
as well and gives the solver the equal-size mask, the true sizes, or
sizes estimated as the cluster sizes of the size-free spectral method in
the same panel (regularized spectral clustering on the graphs, spectral
clustering on the weighted matrices; no labels), next to the size-free
baselines, the Kernighan--Lin-style swap search on the same criterion
with the true sizes (KL) and Alternating LAP started from the
Bethe-Hessian partition. With equal planted blocks the estimated sizes
cost almost nothing. With
unequal blocks the equal-size mask is misspecified and falls below the
size-free methods, the true sizes give the best result in the panel, and
the estimated sizes recover most of the gap, above the partition they
were taken from. The swap search from random starts needs the dense
signal of the weighted matrices and stays low on the sparse graphs, so
the LAP steps, not the exchanges alone, do the work there.

\begin{table}[h]
\caption{Block recovery with the block sizes in the mask equal, true, or
  estimated from a spectral partition (ARI). Top: sparse SBM, $n{=}500$,
  $k{=}4$, $c{=}16$, $20$ seeds, ours $=$ Alternating LAP. Bottom: weighted
  planted partition, $n{=}300$, $k{=}4$, $\sigma{=}1$, $10$ seeds, $16$
  restarts, ours $=$ AltLAP-lin.}
\label{tab:block_sizes}
\centering
\footnotesize
\setlength{\tabcolsep}{3pt}
\begin{tabular}{l l r r r r r r r r}
\toprule
& & & & & & & \multicolumn{3}{c}{Ours, mask sizes} \\
\cmidrule(lr){8-10}
Planted sizes & $\lambda$ & Bethe H. & Reg.\ spectral & Leiden & KL (true) & BH $\to$ ours (true) & equal & true & estimated \\
\midrule
equal & 1.6 & 0.74 & 0.75 & 0.74 & 0.32 & 0.79 & 0.82 & 0.82 & 0.80 \\
equal & 2.0 & 0.92 & 0.92 & 0.95 & 0.94 & 0.95 & 0.95 & 0.95 & 0.94 \\
1:1:2:2 & 1.6 & 0.59 & 0.54 & 0.52 & 0.66 & 0.52 & 0.49 & 0.84 & 0.65 \\
1:1:2:2 & 2.0 & 0.89 & 0.85 & 0.93 & 0.92 & 0.68 & 0.59 & 0.96 & 0.89 \\
\midrule
Planted sizes & $\mu$ & Spectral & KL-lin (true) & Squared loss (true) & & & equal & true & estimated \\
\midrule
equal & 0.4 & 0.84 & 0.93 & 0.00 & & & 0.93 & 0.93 & 0.89 \\
equal & 0.6 & 0.99 & 1.00 & 0.03 & & & 1.00 & 1.00 & 0.99 \\
1:1:2:2 & 0.4 & 0.62 & 0.94 & 0.01 & & & 0.55 & 0.94 & 0.79 \\
1:1:2:2 & 0.6 & 0.98 & 1.00 & 0.07 & & & 0.63 & 1.00 & 0.98 \\
\bottomrule
\end{tabular}
\end{table}

\textbf{Real labelled networks.} The same panel on standard networks with
published communities (Table~\ref{tab:realnets}; equal-size masks, or
``+sizes'': block sizes re-estimated from the order by change-point
segmentation, three rounds, no oracle information). Reference
partitions: the two factions
of Zachary's karate club~\citep{zachary1977information}; the two groups of nine women of the Davis
Southern Women data~\citep{davis1941deepsouth,freeman2003finding}, with the bipartite women--events
graph used as a symmetric adjacency on all $32$ nodes and ARI computed on
the $18$ women; the three classes of the political-books network
(compiled by V.~Krebs; \citealp{newman2006modularity}) and the
twelve conferences of the football network~\citep{girvan2002community}. All
methods receive $k$; PatLR uses rank $1$; results are means over $10$
seeds (standard deviations $\le0.03$, zero for the deterministic
methods). Les Mis\'erables has no published partition (the
$\mathrm{block}(19,19,19,20)$ mask of Section~\ref{sec:exp_patlr} is an
equal split, not a ground truth) and is not scored here. Two details
behind Section~\ref{sec:exp_blocks}: on Karate the low-rank term of
PatLR carries the degree profile, which the plain block fit has to place
inside the blocks; on polbooks ($49/43/13$) the equal-size mask improves
the error reduction while the ARI does not move, a balanced partition
being a different problem from community detection.

\begin{table}[h]
\caption{Community recovery on labelled networks (ARI). ``+sizes'':
  block sizes re-estimated from the order (change-point segmentation)
  instead of equal sizes; ``BH $\to$'' initializes from the
  Bethe-Hessian partition. PatLR uses rank $1$.}
\label{tab:realnets}
\centering
\footnotesize
\begin{tabular}{l r r r r}
\toprule
Method & Karate ($k{=}2$) & Davis ($k{=}2$) & Polbooks ($k{=}3$) & Football ($k{=}12$) \\
\midrule
Spectral clustering & 0.882 & 0.000 & 0.465 & 0.813 \\
Regularized spectral & 0.882 & 0.000 & 0.624 & 0.869 \\
Bethe Hessian & 0.882 & \textbf{1.000} & \textbf{0.659} & 0.897 \\
Non-backtracking & 0.882 & \textbf{1.000} & 0.405 & \textbf{0.906} \\
Leiden & 0.465 & 0.405 & 0.621 & 0.807 \\
\midrule
AltLAP, equal blocks & 0.772 & 0.777 & 0.455 & 0.739 \\
AltLAP + sizes & 0.772 & \textbf{1.000} & 0.473 & 0.733 \\
BH $\to$ AltLAP & 0.772 & 0.777 & 0.479 & 0.763 \\
BH $\to$ AltLAP + sizes & 0.772 & 0.777 & 0.479 & 0.793 \\
PatLR, equal blocks & \textbf{1.000} & 0.777 & 0.492 & 0.743 \\
PatLR + sizes & \textbf{1.000} & 0.777 & 0.476 & 0.775 \\
BH $\to$ PatLR & 0.772 & \textbf{1.000} & 0.477 & 0.752 \\
BH $\to$ PatLR + sizes & 0.772 & \textbf{1.000} & 0.481 & 0.764 \\
\bottomrule
\end{tabular}
\end{table}

\subsubsection{Composite target: a band inside blocks}
\label{app:composite}

Planted matrices with $k\in\{3,5\}$ blocks (equal or unequal sizes), a
band of width $w\in\{1,3\}$ inside each block, noise
$\sigma\in\{0.1,0.3\}$, $n=150$, $20$ seeds: $16$ configurations. The
target mask is the band \emph{inside} the blocks (block AND band); the
union mask would be blind to the within-block order. Metrics: error
reduction on the target mask, ARI of the block assignment read off the
order, and within-block Kendall $\tau$. Methods: spectral clustering
(blocks only), global Fiedler ordering, a two-stage pipeline (cluster,
then spectral order within each cluster), QAP-FAQ and AltLAP on the
composite mask, AltLAP with a change-point estimate of the block
boundaries (for unequal sizes), and AltLAP warm-started from the
two-stage pipeline.

\begin{table}[h]
\caption{Composite band-in-block target, averages over the $16$
  configurations and $20$ seeds; per-configuration values in
  Table~\ref{tab:composite_full}.}
\label{tab:composite}
\centering
\footnotesize
\begin{tabular}{l r r r}
\toprule
Method & target ER & block ARI & within-block $\tau$ \\
\midrule
Spectral clustering & 0.057 & \textbf{1.000} & 0.09 \\
Fiedler ordering & 0.050 & 0.839 & 0.10 \\
Two-stage (cluster, then order) & 0.115 & \textbf{1.000} & 0.45 \\
QAP-FAQ (composite mask) & 0.137 & 0.181 & 0.18 \\
AltLAP (composite mask) & 0.171 & 0.781 & 0.20 \\
AltLAP + boundary estimation & 0.173 & 0.877 & 0.20 \\
\textbf{Two-stage, then AltLAP} & \textbf{0.190} & \textbf{1.000} & \textbf{0.47} \\
\midrule
Two-stage, convex seriation inside & 0.124 & \textbf{1.000} & 0.46 \\
\textbf{Two-stage (convex), then AltLAP} & \textbf{0.193} & \textbf{1.000} & \textbf{0.48} \\
\bottomrule
\end{tabular}
\end{table}

Table~\ref{tab:composite} averages over the configurations and
Table~\ref{tab:composite_full} gives each one with its paired interval;
the summary is in Section~\ref{sec:exp_blocks}. Details: the refining
pass re-estimates the block boundaries from the order by change-point
segmentation, so the block sizes are not given; it raises the target
error reduction by $+17\%$ to $+257\%$ relative (median $+96\%$), from
a median $45\%$ of the planted order's value for the pipeline to $94\%$
for the refined run ($90$--$99\%$ at $w{=}3$, $\sigma{=}0.1$), and by
$+11\%$ to $+227\%$ when the pipeline uses convex seriation inside the
clusters; the within-block $\tau$ gain at $w{=}3$ is $+0.02$ to
$+0.09$ with the interval above zero in $7$ of $8$ configurations, and
at $w{=}1$ it is within $\pm0.02$, a band of width one determining little
of the order. From a cold start the composite mask alone gives block ARI
$0.78$, and $0.88$ with the boundary estimate (e.g.\ $0.45\to0.89$ on
$k=3$, unequal, $w=3$, $\sigma=0.3$); for the within-block order the hard
mask truncates the smooth decay, and the pipeline's seriation is the
better starting point, which is why the two are combined. FAQ from its
default starts does not recover the blocks (Table~\ref{tab:composite}).

\begin{table}[h]
\caption{Composite target per configuration ($20$ seeds): the two-stage
  pipeline (spectral clustering, spectral seriation inside each cluster)
  and the same pipeline followed by one AltLAP pass on the band-in-block
  mask; ``oracle'' is the error reduction of the planted order on that
  mask. Paired differences with $95\%$ bootstrap intervals; block ARI is
  $1.00$ for both in every configuration. The last two columns repeat the
  ER for the pipeline with convex seriation inside the clusters
.}
\label{tab:composite_full}
\centering
\scriptsize
\setlength{\tabcolsep}{1.7pt}
\begin{tabular}{l l l r r r r l r r l r r}
\toprule
& & & & \multicolumn{4}{c}{target ER} & \multicolumn{3}{c}{within-block $\tau$} & \multicolumn{2}{c}{ER, convex inside} \\
\cmidrule(lr){5-8}\cmidrule(lr){9-11}\cmidrule(lr){12-13}
$k$ & $w$ & $\sigma$ & oracle & pipel. & $+$Alt & rel. & $\Delta$ [CI] & pipel. & $+$Alt & $\Delta$ [CI] & pipel. & $+$Alt \\
\midrule
3 eq. & 1 & 0.1 & 0.130 & 0.043 & 0.104 & +142\% & $+0.061$ [$+0.060,+0.062$] & 0.31 & 0.32 & $+0.007$ [$-0.009,+0.023$] & 0.045 & 0.105 \\
3 eq. & 1 & 0.3 & 0.086 & 0.025 & 0.084 & +239\% & $+0.059$ [$+0.058,+0.061$] & 0.17 & 0.16 & $-0.009$ [$-0.029,+0.010$] & 0.026 & 0.084 \\
3 eq. & 3 & 0.1 & 0.305 & 0.196 & 0.288 & +47\% & $+0.093$ [$+0.088,+0.097$] & 0.71 & 0.76 & $+0.054$ [$+0.038,+0.071$] & 0.215 & 0.290 \\
3 eq. & 3 & 0.3 & 0.214 & 0.100 & 0.200 & +100\% & $+0.100$ [$+0.096,+0.103$] & 0.41 & 0.43 & $+0.024$ [$-0.001,+0.050$] & 0.100 & 0.200 \\
3 uneq. & 1 & 0.1 & 0.116 & 0.036 & 0.091 & +154\% & $+0.055$ [$+0.053,+0.057$] & 0.29 & 0.30 & $+0.003$ [$-0.010,+0.015$] & 0.038 & 0.093 \\
3 uneq. & 1 & 0.3 & 0.079 & 0.022 & 0.077 & +257\% & $+0.055$ [$+0.054,+0.057$] & 0.17 & 0.15 & $-0.018$ [$-0.039,+0.004$] & 0.024 & 0.077 \\
3 uneq. & 3 & 0.1 & 0.276 & 0.163 & 0.248 & +53\% & $+0.085$ [$+0.082,+0.089$] & 0.63 & 0.68 & $+0.055$ [$+0.041,+0.071$] & 0.178 & 0.257 \\
3 uneq. & 3 & 0.3 & 0.199 & 0.083 & 0.179 & +115\% & $+0.096$ [$+0.093,+0.098$] & 0.40 & 0.44 & $+0.030$ [$+0.013,+0.050$] & 0.090 & 0.185 \\
5 eq. & 1 & 0.1 & 0.193 & 0.090 & 0.168 & +87\% & $+0.078$ [$+0.076,+0.080$] & 0.44 & 0.46 & $+0.019$ [$+0.009,+0.028$] & 0.096 & 0.166 \\
5 eq. & 1 & 0.3 & 0.110 & 0.043 & 0.108 & +153\% & $+0.065$ [$+0.063,+0.068$] & 0.24 & 0.22 & $-0.021$ [$-0.037,-0.004$] & 0.046 & 0.108 \\
5 eq. & 3 & 0.1 & 0.415 & 0.353 & 0.413 & +17\% & $+0.060$ [$+0.055,+0.065$] & 0.87 & 0.96 & $+0.089$ [$+0.080,+0.099$] & 0.372 & 0.414 \\
5 eq. & 3 & 0.3 & 0.263 & 0.173 & 0.260 & +50\% & $+0.087$ [$+0.084,+0.090$] & 0.56 & 0.63 & $+0.065$ [$+0.048,+0.082$] & 0.183 & 0.262 \\
5 uneq. & 1 & 0.1 & 0.166 & 0.072 & 0.138 & +93\% & $+0.067$ [$+0.065,+0.068$] & 0.40 & 0.41 & $+0.013$ [$+0.003,+0.023$] & 0.076 & 0.140 \\
5 uneq. & 1 & 0.3 & 0.100 & 0.036 & 0.097 & +168\% & $+0.061$ [$+0.060,+0.062$] & 0.21 & 0.21 & $-0.006$ [$-0.026,+0.014$] & 0.040 & 0.099 \\
5 uneq. & 3 & 0.1 & 0.369 & 0.272 & 0.347 & +28\% & $+0.076$ [$+0.072,+0.080$] & 0.80 & 0.86 & $+0.062$ [$+0.050,+0.071$] & 0.303 & 0.361 \\
5 uneq. & 3 & 0.3 & 0.243 & 0.142 & 0.230 & +62\% & $+0.088$ [$+0.084,+0.091$] & 0.51 & 0.55 & $+0.046$ [$+0.017,+0.074$] & 0.151 & 0.238 \\
\bottomrule
\end{tabular}
\end{table}

\subsection{Attention Head Analysis: Full Details}
\label{app:attention_details}

This appendix gives the pipeline behind Section~\ref{sec:exp_attention},
the per-model results, and the identity-order control. All numbers come
from one run of the unified scoring described below.

\subsubsection{Score, Vocabulary and a Synthetic Check}
\label{app:attention_theory}

The score $\tilde{s}_M(A)$ of~\eqref{eq:adaptive_score} on a causal
attention matrix (rows queries, columns keys) is the minimum mass
outside the mask after a common reordering of queries and keys,
normalized by $\|A\|_F^2 (1-\rho_M)^{3/2}$
(equation~\eqref{eq:adaptive_score}). The strict upper triangle is
identically zero and contributes nothing to $d_M$ for any mask and
permutation, so no symmetrization is needed. The masks are the plain
geometric masks of Section~\ref{sec:prelim}, not their intersection
with the lower triangle (the arrowhead is the full hub row, hub column
and diagonal, which is what the closed form of Appendix~\ref{app:special}
assumes); a common permutation does not preserve the causal triangle
itself, and the structural labels describe the reordered matrix.
The nine masks of Section~\ref{sec:exp_attention} were fixed ahead of
time to span the regimes of the standard taxonomy of
attention~\citep{clark2019bert,voita2019analyzing,xiao2023efficient}:
local routing (bands of width $1$, $2$, $4$), segment- or topic-level
grouping (blocks, $k=2,3,4$), sink-like global routing of every query to
a few distinguished keys such as BOS (arrowhead), modular-local heads
(three blocks plus the band $|i-j|\le2$) and pure self-loops (diagonal).
The vocabulary is a controlled basis set, not an exhaustive one. At
$n{=}512$ the block+band mask (blocks of $170$ plus a remainder block
of $2$, plus the band) differs from block-$k$3 (blocks of $170$, $170$,
$172$) in $0.3\%$ of the entries, so near-ties between the two labels are
expected and the two are read together below.

\begin{figure}[h]
  \centering
  \includegraphics[width=\linewidth]{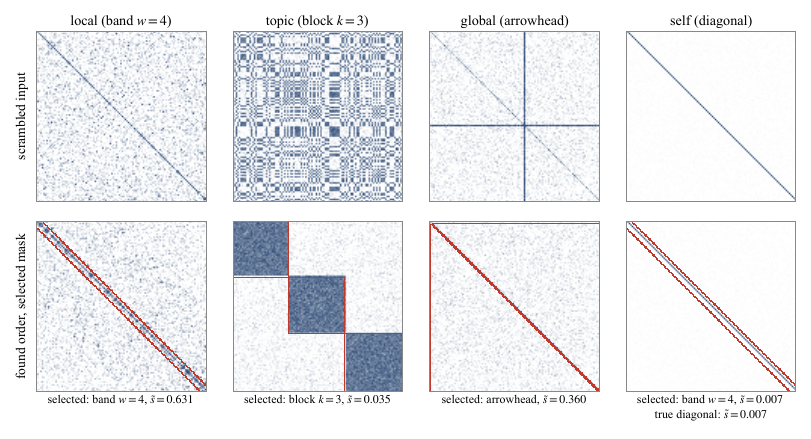}
  \caption{Synthetic attention heads, one instance per type: the
    scrambled input (top) and the same matrix in the order returned for
    the selected mask, mask outlined (bottom), with the
    density-normalized scores of the selected and, where different, of
    the true mask.}
  \label{fig:attention_synthetic}
\end{figure}

\paragraph{Synthetic check.}
Four head types with a constructed ground truth are scored with the
full pipeline at $n{=}128$ ($20$ instances each, scrambled by a random
permutation; Table~\ref{tab:synthetic_attention},
Figure~\ref{fig:attention_synthetic}): a local head (Gaussian kernel of
width $1.5$ tokens plus noise, so that the band $|i-j|\le4$ holds $99\%$
of the kernel mass), a topic head (three equal blocks plus noise), a
global head (every query attends the first key, plus a diagonal and
noise) and a self head (a dominant diagonal plus noise). The global and
topic heads are identified in every instance, the topic head either as
block-$k$3 or as the block+band mask that contains its blocks; the local
head is identified in most instances, with block masks selected otherwise
at a small margin (the noise carries two thirds of the energy of these
instances). The self head is a boundary case of the vocabulary: every
mask that contains the diagonal fits it, the optimized thin bands pack
slightly more noise than the diagonal alone, and the label is a thin band
in every instance, by a small margin.
A pure self-loop head is therefore reported by this pipeline as a thin
band, and the band labels of real heads should be read as covering it.

\begin{table}[h]
\caption{Synthetic attention heads scored with the paper's pipeline
  (9-mask vocabulary, density-normalized score, identity candidate,
  $n{=}128$, $20$ instances per type): instances in which the true mask
  is the label, the labels selected, and the median relative margin of
  the selected mask over the runner-up.}
\label{tab:synthetic_attention}
\centering
\footnotesize
\setlength{\tabcolsep}{4pt}
\begin{tabular}{l l c l c}
\toprule
Head type & True mask & True mask selected & Labels selected & Margin \\
\midrule
% <<SYNTHROWS
Local & band $w{=}4$ & 15/20 & band $w{=}4$ (15), block+band (3), block $k{=}3$ (1), block $k{=}4$ (1) & 0.03 \\
Topic & block $k{=}3$ & 12/20 & block $k{=}3$ (12), block+band (8) & 0.56 \\
Global & arrowhead & 20/20 & arrowhead (20) & 0.60 \\
Self & diagonal & 0/20 & band $w{=}4$ (20) & 0.02 \\
% SYNTHROWS>>
\bottomrule
\end{tabular}
\end{table}

\subsubsection{Pipeline}
\label{app:tech_attention}

\textbf{Models and prompts.} GPT-2~\citep[$117$M parameters, $12$ layers
$\times$ $12$ heads;][]{radford2019language}, Qwen2.5-0.5B ($24{\times}14$) and Qwen2.5-7B
($28{\times}28$)~\citep{qwen2025qwen25}, loaded from the Hugging Face hub (\texttt{gpt2},
\texttt{Qwen/Qwen2.5-0.5B}, \texttt{Qwen/Qwen2.5-7B}). The prompts are
the first $20$ passages of at least $5n$ characters formed from
consecutive non-heading lines of the WikiText-2 validation
split~\citep{merity2017pointer} (its
first two to five articles), tokenized by each model's tokenizer and
truncated to $n\in\{128,512\}$ tokens (two passages fall a few tokens
short of $512$); the passages differ between the two lengths.

\textbf{Extraction.} One forward pass per (model, prompt, length) on a
single GPU (one H100; about two hours for the full grid of models,
prompts and lengths) returns the $L{\times}H{\times}n{\times}n$ tensor of
post-softmax attention weights (fp16 for the Qwen models, fp32 for
GPT-2; non-finite rows, which occur in $\le0.1\%$ of the rows of
Qwen2.5-7B, are zeroed).

\textbf{Scoring.} For every (layer, head, prompt) and every mask of the
vocabulary we compute (i) the optimized score, Alternating LAP from a
spectral start with $8$ iterations and $2$ restarts at $n{=}512$ ($12$
and $3$ at $n{=}128$). On a causal $A$ the symmetric eigensolver of
Appendix~\ref{app:init} reads the lower triangle, so the operator it
diagonalizes is $D - W$ with $W = A + A^\top - \mathrm{diag}\,A$ and
$D=\mathrm{diag}(A\mathbf{1})$, the identity for softmax rows; the start
is therefore the order by the second-largest eigenvector of the
symmetrized attention matrix $W$, not the Fiedler vector of its graph
Laplacian (whose degree matrix would be $\mathrm{diag}(W\mathbf{1})$).
The objective itself is not symmetrized. (ii) The identity score, under the native
token order, and (iii) the mean score under five random permutations
(the null level). The score of a mask is the lower of (i) and (ii); for
the arrowhead mask the closed-form optimum of
Appendix~\ref{app:special} is taken as well (it is attained to within
$1\%$ by the solver for $99.6$--$99.9\%$ of the pairs). The label of a
pair is the mask with the lowest score; its margin is the relative gap
to the runner-up. Scoring one $512$-token head takes about $12$
core-seconds (Appendix~\ref{app:scaling}).

\textbf{Intervals.} All intervals are $95\%$ bootstrap intervals over
prompts (the $20$ prompts resampled with replacement, the statistic
recomputed on all heads), i.e.\ they treat the prompt, not the (head,
prompt) pair, as the unit of replication.

\subsubsection{Per-Model Results}
\label{app:attention_experiments}

\begin{figure}[h]
  \centering
  \includegraphics[width=\linewidth]{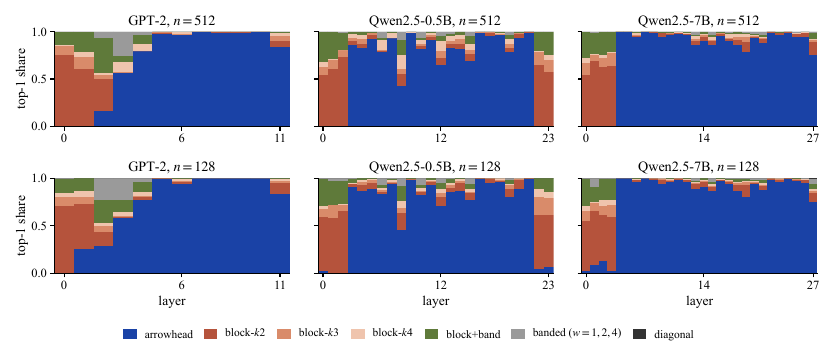}
  \caption{Share of the top-$1$ label per layer for the three models
    (columns) at $n{=}512$ (top) and $n{=}128$ (bottom), $20$ prompts.}
  \label{fig:attention_profiles}
\end{figure}

\begin{figure}[h]
  \centering
  \includegraphics[width=\linewidth]{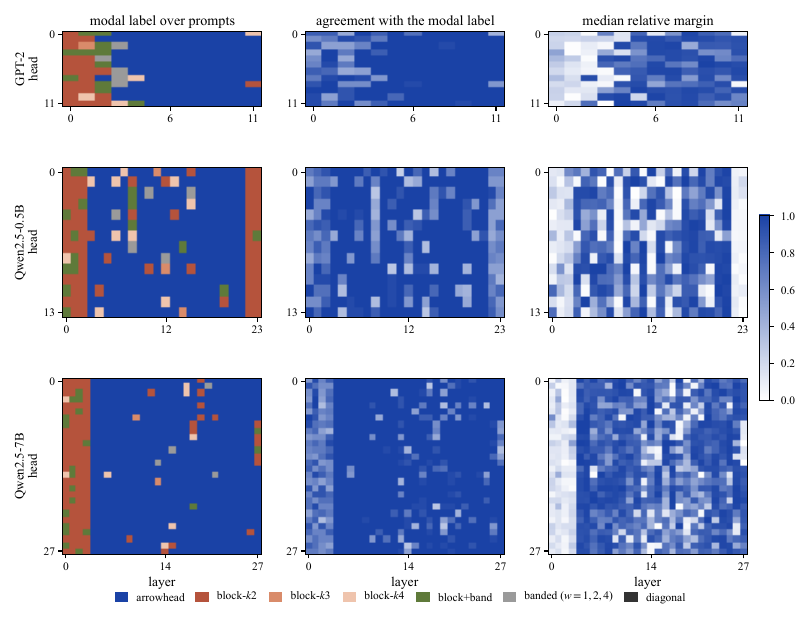}
  \caption{Per-head view at $n{=}512$, one row per model: the modal
    label over the $20$ prompts (left), the share of prompts carrying
    that label (middle) and the median relative margin of the selected
    mask over the runner-up (right).}
  \label{fig:attention_headmap}
\end{figure}

\begin{figure}[h]
  \centering
  \includegraphics[width=\linewidth]{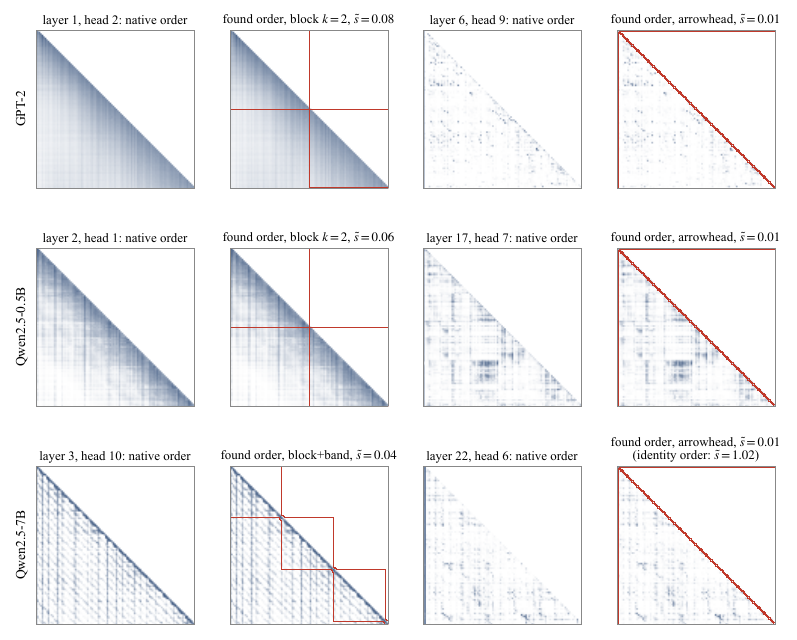}
  \caption{Representative heads of each model at $n{=}128$ (one passage
    per model), logarithmic colour scale: a block-labelled head (left pair) and an
    arrowhead head (right pair), each in the native token order and in
    the order found for the selected mask, mask outlined, with the
    density-normalized score. The block-labelled heads are wide local
    heads, fitted by the native order; the Qwen2.5-7B arrowhead head has
    its sink at a key other than the first token (identity score $1.02$),
    which the reordering brings to the front.}
  \label{fig:permuted_detail}
\end{figure}

Table~\ref{tab:attention_summary} gives the per-model numbers behind
Section~\ref{sec:exp_attention} at both lengths;
Figure~\ref{fig:attention_profiles} the per-layer label shares and
Figure~\ref{fig:attention_headmap} the per-head modal label, its
agreement over prompts and its margin. The depth profile is the same in
all six (model, length) cells: block labels in the first three to four
layers, arrowhead after them, with Qwen2.5-0.5B returning to block-$k$2
in its last two layers at both lengths. Within a model the label shares
move by a few points between $n{=}128$ and $n{=}512$, and the modal
label of a head is the same at both lengths for $91$--$95\%$ of the
heads ($0.70$--$0.77$ among the heads whose modal label is not arrowhead
at both lengths, so the block-type labels are less stable than the
arrowhead label and the layer profile). On the
$10$-bin relative-depth grid the three models agree in $8/10$ bins at
both lengths; pairwise agreement is $0.9$ for GPT-2 vs.\ Qwen2.5-7B and
for Qwen2.5-0.5B vs.\ Qwen2.5-7B and $0.8$ for GPT-2 vs.\ Qwen2.5-0.5B.
The last column of Table~\ref{tab:attention_summary} is the accuracy of
a multinomial logistic regression that predicts the label of a pair from
four scalar descriptors of the head on that prompt (row entropy, mean
maximum weight, sparsity, mean distance from the diagonal), trained on
$19$ prompts and tested on the held-out one; it is well above the
majority label and well below one, so these descriptors partially
predict the labels under the tested logistic-regression model. About
three quarters of the
pairs carry a confident label, a margin over the runner-up above $20\%$
(column ``confident'').
Figure~\ref{fig:permuted_detail} shows representative heads before and
after reordering: the block-labelled heads of the first layers are wide
local heads whose mass already lies in the diagonal blocks in the native
order (the native order is within $1\%$ of the optimum for $74$--$84\%$
of the block-labelled pairs at $n{=}512$, $57$--$71\%$ at $n{=}128$),
and on the Qwen2.5-7B passages whose sink is not the first token the
reordering moves the sink key to the front.

\begin{table}[h]
\caption{Per-model attention results (unified scoring). Label shares in
  \% of (head, prompt) pairs: arrowhead with its $95\%$ interval over
  prompts, block-$k$2, the other block masks and block+band together,
  bands and diagonal together. Agreement: mean per-head agreement with
  the head's modal label over the $20$ prompts, for all heads, for the
  heads whose modal label is arrowhead, and for the others. Confident:
  share of pairs whose margin over the runner-up exceeds $20\%$. Null:
  median ratio of the selected mask's score to its score under random
  permutations. Scalars: leave-one-prompt-out accuracy of predicting the
  label from four scalar descriptors / majority baseline.}
\label{tab:attention_summary}
\centering
\scriptsize
\setlength{\tabcolsep}{3pt}
\begin{tabular}{l r l r r r r r r r r r}
\toprule
 & & \multicolumn{4}{c}{label share (\%)} & \multicolumn{3}{c}{agreement over prompts} & & & \\
\cmidrule(lr){3-6}\cmidrule(lr){7-9}
Model & $n$ & arrowhead [CI] & block-$k$2 & other blocks & bands & all & arrowhead & other & confident & null & scalars / maj. \\
\midrule
% <<SUMMARYROWS
GPT-2 & 512 & 69.3 [69.0, 69.7] & 14.8 & 12.9 & 3.0 & 0.91 & 0.99 & 0.74 & 0.77 & 0.07 & 0.87 / 0.69 \\
 & 128 & 72.2 [71.7, 72.6] & 12.8 & 10.8 & 4.1 & 0.90 & 0.98 & 0.68 & 0.78 & 0.08 & 0.86 / 0.72 \\
Qwen2.5-0.5B & 512 & 67.1 [66.5, 67.6] & 15.9 & 15.4 & 1.6 & 0.86 & 0.96 & 0.64 & 0.71 & 0.12 & 0.81 / 0.67 \\
 & 128 & 69.1 [68.3, 69.8] & 16.3 & 12.9 & 1.7 & 0.86 & 0.96 & 0.62 & 0.74 & 0.12 & 0.81 / 0.69 \\
Qwen2.5-7B & 512 & 79.9 [79.4, 80.3] & 11.4 & 8.2 & 0.5 & 0.92 & 0.98 & 0.67 & 0.83 & 0.09 & 0.89 / 0.80 \\
 & 128 & 81.3 [80.8, 81.8] & 10.2 & 7.4 & 1.0 & 0.91 & 0.97 & 0.63 & 0.85 & 0.09 & 0.89 / 0.81 \\
% SUMMARYROWS>>
\bottomrule
\end{tabular}
\end{table}

\subsubsection{Identity-Order Control and Long-Context Control}
\label{app:identity}

\textbf{Are the permutations needed?}
Table~\ref{tab:identity} scores every (head, prompt) pair at $n{=}512$
under three rules: the identity order only, the optimized order only
(spectral start, no identity candidate), and the unified rule that keeps
the identity as a candidate for every mask. The optimized score of a mask can
be worse than its identity score, because the solver is started from the
spectral start and not from the identity, and on causal attention the
native order already is the band; comparing identity labels with
optimized-only labels therefore counts label changes caused by competing
masks getting worse. Under the unified rule only a few percent of the
labels of GPT-2 and Qwen2.5-0.5B change, against about a fifth under the
optimized-only rule. On Qwen2.5-7B the change is concentrated on
particular passages: on $15$ of the $20$ passages at $n{=}512$ ($14$ at
$n{=}128$) the native-order arrowhead share is $0.82$ and $3$--$5\%$ of
the labels change, as on the smaller models; on the remaining passages,
all of which begin with the token ``The'', the sink is not the first
token, the native-order arrowhead share is below $0.02$, and $83$--$85\%$
of the labels change. With reordering the arrowhead share is
$0.77$--$0.83$ on every Qwen2.5-7B passage. This is also why the
interval over prompts of the Qwen2.5-7B label-change rate in
Table~\ref{tab:identity} is wide. The last column checks the solver
against the closed-form arrowhead optimum (Appendix~\ref{app:special}).

\begin{table}[h]
\caption{Identity-order control at $n{=}512$, $20$ prompts. Arrowhead
  share of the (head, prompt) labels under the identity order, under the
  optimized order only, and under the unified rule (identity kept as a
  candidate); share of labels that differ from the identity labels under
  the unified rule ($95\%$ interval over prompts) and under the
  optimized-only rule; share of pairs on which the solver's arrowhead
  score is within $1\%$ of the closed-form optimum.}
\label{tab:identity}
\centering
\footnotesize
\setlength{\tabcolsep}{4pt}
\begin{tabular}{l r r r r r r}
\toprule
 & \multicolumn{3}{c}{arrowhead share} & \multicolumn{2}{c}{labels changed vs identity} & \\
\cmidrule(lr){2-4}\cmidrule(lr){5-6}
Model & identity & optimized & unified & unified [CI] & optimized & solver $=$ exact \\
\midrule
% <<IDENTITYROWS
GPT-2 & 0.70 & 0.70 & 0.69 & 3.5\% [3, 4] & 19\% & 99.9\% \\
Qwen2.5-0.5B & 0.70 & 0.68 & 0.67 & 6.5\% [6, 7] & 24\% & 99.8\% \\
Qwen2.5-7B & 0.61 & 0.80 & 0.80 & 23.4\% [8, 39] & 34\% & 99.6\% \\
% IDENTITYROWS>>
\bottomrule
\end{tabular}
\end{table}

\textbf{Position subsampling does not transfer.}
A tempting shortcut for long contexts is to score a random subset of
$128$ of the $512$ positions. The label obtained this way agrees with the
full-length label for only $0.40$ / $0.41$ / $0.41$ of (head, prompt)
pairs (GPT-2 / Qwen2.5-0.5B / Qwen2.5-7B; $5$ prompts, two draws each,
unified scoring at the sub-length budget), and the per-head modal label
over these draws agrees with the full-length one for about half of the
heads, against $0.91$--$0.95$ agreement of the per-head modal label
between the contiguous $n{=}128$ and $n{=}512$ contexts. Non-contiguous
subsampling
destroys the band and block geometry; long contexts should be analysed
on contiguous windows or with the scaling routes of
Appendix~\ref{app:scaling}.

\subsection{Wall-Clock Scaling of Alternating LAP}
\label{app:scaling}

Planted banded instances in the Gaussian regime of
Section~\ref{sec:exp_algo} ($w{=}5$), one Fiedler-initialized run of the
LAP phase with an iteration cap of $10$, $3$
seeds at every $n$; the exchange phase is timed separately below. All
times are single-thread wall-clock seconds (BLAS and the assignment
solver pinned to one core) on an Apple M3 with $18$\,GB of memory; peak
resident memory is the maximum over the three runs. Two noise series are reported: the per-entry noise fixed at
$\sigma_n{=}0.05$ (the regime of Table~\ref{tab:baselines}), under which
the band holds a $1/n$ fraction of the energy and the error reduction of
the planted order itself falls from $0.80$ to $0.20$; and the noise
scaled as $\sigma_n \propto n^{-1/2}$, which keeps the planted order's
error reduction at $0.80$ for every $n$.
Table~\ref{tab:scaling} gives the measurements.

\begin{table}[h]
\caption{Wall-clock scaling of the LAP phase of one Alternating LAP run
  (single thread, Apple M3). Times in seconds; ``LAPs'' is the part of an iteration
  spent in the two assignment solves; ER is the error reduction of the
  solver and, in parentheses, of the planted order. FAQ: one run of
  QAP-FAQ on the same instances and objective. Means over $3$ seeds
  (FAQ at $n\ge4000$ and the fixed-SNR series at $n{=}8000$: one seed).}
\label{tab:scaling}
\centering
\footnotesize
\setlength{\tabcolsep}{4pt}
\begin{tabular}{r rrrrr r rr rr}
\toprule
& \multicolumn{5}{c}{time} & & \multicolumn{2}{c}{ER, $\sigma_n{=}0.05$} & \multicolumn{2}{c}{ER, $\sigma_n\propto n^{-1/2}$} \\
\cmidrule(lr){2-6}\cmidrule(lr){8-9}\cmidrule(lr){10-11}
$n$ & Fiedler & per iter. & LAPs & total & FAQ & memory & AltLAP & FAQ & AltLAP & time \\
\midrule
500  & 0.0 & 0.04 & 0.03 & 0.4 & 0.7 & 0.2\,GB & 0.51 (0.80) & 0.57 & 0.51 (0.80) & 1.0 \\
1000 & 0.1 & 0.14 & 0.10 & 1.5 & 4.7 & 0.4\,GB & 0.44 (0.67) & 0.48 & 0.52 (0.80) & 1.6 \\
2000 & 1.0 & 0.73 & 0.49 & 8.3 & 32.5 & 0.6\,GB & 0.34 (0.50) & 0.38 & 0.55 (0.80) & 7.5 \\
4000 & 10.9 & 5.4 & 3.1 & 65 & 249 & 1.3\,GB & 0.24 (0.34) & 0.25 & 0.59 (0.80) & 46 \\
8000 & 105 & 37 & 18 & 479 & 1936 & 3.2\,GB & 0.15 (0.20) & 0.16 & 0.63 (0.80) & 334 \\
\bottomrule
\end{tabular}
\end{table}

\paragraph{Cost.}
An iteration of the LAP phase is two $O(n^3)$ cost-matrix products and
two assignment solves; the solves take three quarters of it at small $n$ and half at
$n{=}8000$, and the iteration time decreases as the iterates converge
(first and last iteration at $n{=}8000$: $78$\,s and $22$\,s), so the
total grows by $4$--$8\times$ per doubling of $n$. The dense Fiedler
start (a full eigendecomposition) becomes a fifth of the total at the
largest size. Memory is a handful of dense $n\times n$ doubles. FAQ on
the same objective costs $2$--$4\times$ more and reaches a similar error
reduction from one start.

\paragraph{Exchange phase.}
On the same instances, started from the output of the LAP phase above,
the exchange descent of Appendix~\ref{app:init} takes $0.8$--$1.1$\,s at
$n{=}500$ ($20$--$28$ rounds), $6.4$--$7.6$\,s at $n{=}1000$ ($32$--$39$
rounds) and $24$--$48$\,s at $n{=}2000$ ($25$--$48$ rounds), and raises
the error reduction by $0.07$--$0.08$, $0.06$--$0.07$ and
$0.04$--$0.05$. A round is dominated by the gain evaluation, two
$n\times n$ products: $0.4$\,s at $n{=}2000$, $2.7$\,s at $n{=}4000$
and $20$\,s at $n{=}8000$ on this machine, so at the observed $25$--$50$
rounds the phase would take a few minutes at $n{=}4000$ and of the order
of $10$--$20$ minutes at $n{=}8000$, comparable with the LAP phase; it
was not run at these two sizes. On the same machine PatLR (rank $3$,
$12$ iterations, $3$ restarts) takes $3.0$\,s at $n{=}200$, $22$\,s at $n{=}500$ and $134$\,s at
$n{=}1000$; Adaptive PatLR ($N{=}80$, $\ell{=}3$, default vocabulary)
$2.9$\,s at $n{=}200$ and $31$\,s at $n{=}500$, i.e.\ $1$--$1.4\times$
the cost of one PatLR call. The benchmark panels and the attention
scoring run as single-threaded processes on a $24$-core AMD EPYC;
scoring one $512\times512$ attention head against the $9$-mask
vocabulary ($8$ iterations, $2$ restarts per mask, plus the identity and
five random permutations) takes about $12$ core-seconds there, so the
full grid of Section~\ref{sec:exp_attention} ($1{,}264$ heads $\times$
$20$ prompts at $n{=}512$, and again at $n{=}128$) is about $85$
core-hours ($22$ processes).

\paragraph{Scaling routes.}
On one M3 core a solve on a planted banded instance thus takes about
two seconds at $n{=}500$ and, for the LAP phase alone, minutes at
$n{=}8000$ (Table~\ref{tab:scaling}), and a vocabulary scan of a single
$8000$-token matrix would take hours per head. The cost matrix is dense
(a banded mask only makes it $O(n^2)$ to assemble), so the assignment
solve is the bottleneck; sparse assignment with restricted candidates,
coarsen-then-refine, and GPU Sinkhorn or auction
solvers~\citep{cuturi2013sinkhorn,bertsekas1988auction} are left for
future work.

\paragraph{Quality against $n$.}
The iteration cap is not binding: with the cap raised to $30$ the
solver stops by its own convergence test after $9$--$12$ iterations at
$n\le 2000$ and the error reduction changes by less than $0.001$. Under
fixed per-entry noise the error reduction falls with $n$ because the
attainable reduction falls: the solver reaches a stable two thirds to
three quarters of the planted order's value at every $n$. With the
matrix SNR held fixed the error reduction does not degrade with size; it
rises slowly with $n$. Under fixed per-entry noise the Fiedler start
alone reduces the error by $0.1$ at $n{=}500$ and by less than $0.04$ at
$n\ge 1000$ (the Laplacian of $|A|$ is dominated by the dense noise), so
the reduction is produced by the LAP iterations.

\clearpage
% =============================================================================
\section*{NeurIPS Paper Checklist}
% =============================================================================

\begin{enumerate}

\item {\bf Claims}
    \item[] Question: Do the main claims made in the abstract and introduction accurately reflect the paper's contributions and scope?
    \item[] Answer: \answerYes{}
    \item[] Justification: Claims match Sections~\ref{sec:prelim}--\ref{sec:experiments} and the appendices.

\item {\bf Limitations}
    \item[] Question: Does the paper discuss the limitations of the work performed by the authors?
    \item[] Answer: \answerYes{}
    \item[] Justification: Section~\ref{sec:discussion}: cost, local optima, vocabulary scope, proxy objective, theory scope.

\item {\bf Theory assumptions and proofs}
    \item[] Question: For each theoretical result, does the paper provide the full set of assumptions and a complete (and correct) proof?
    \item[] Answer: \answerYes{}
    \item[] Justification: Full proofs in Appendices~\ref{app:proofs}--\ref{app:theory}; assumptions stated with each result.

    \item {\bf Experimental result reproducibility}
    \item[] Question: Does the paper fully disclose all the information needed to reproduce the main experimental results of the paper to the extent that it affects the main claims and/or conclusions of the paper (regardless of whether the code and data are provided or not)?
    \item[] Answer: \answerYes{}
    \item[] Justification: Setups, hyperparameters and seed counts per experiment (typically $20$, exceptions stated) in Section~\ref{sec:experiments} and Appendix~\ref{app:experiments}.

\item {\bf Open access to data and code}
    \item[] Question: Does the paper provide open access to the data and code, with sufficient instructions to faithfully reproduce the main experimental results, as described in supplemental material?
    \item[] Answer: \answerNo{}
    \item[] Justification: No code release; Algorithms~\ref{alg:alternating_lap}--\ref{alg:adaptive} fully specified; all data public.

\item {\bf Experimental setting/details}
    \item[] Question: Does the paper specify all the training and test details (e.g., data splits, hyperparameters, how they were chosen, type of optimizer) necessary to understand the results?
    \item[] Answer: \answerYes{}
    \item[] Justification: Section~\ref{sec:experiments} and Appendix~\ref{app:experiments}.

\item {\bf Experiment statistical significance}
    \item[] Question: Does the paper report error bars suitably and correctly defined or other appropriate information about the statistical significance of the experiments?
    \item[] Answer: \answerYes{}
    \item[] Justification: $95\%$ bootstrap intervals: over seeds (typically $20$) for the benchmarks, over the $20$ prompts for attention.

\item {\bf Experiments compute resources}
    \item[] Question: For each experiment, does the paper provide sufficient information on the computer resources (type of compute workers, memory, time of execution) needed to reproduce the experiments?
    \item[] Answer: \answerYes{}
    \item[] Justification: Attention extraction: one H100, about $2$ hours. Benchmark panels and attention scoring: AMD EPYC, single-threaded processes, about $85$ core-hours. Scaling study and PatLR timings: Apple M3 (Appendix~\ref{app:scaling}).
    
\item {\bf Code of ethics}
    \item[] Question: Does the research conducted in the paper conform, in every respect, with the NeurIPS Code of Ethics \url{https://neurips.cc/public/EthicsGuidelines}?
    \item[] Answer: \answerYes{}
    \item[] Justification: Conforms.

\item {\bf Broader impacts}
    \item[] Question: Does the paper discuss both potential positive societal impacts and negative societal impacts of the work performed?
    \item[] Answer: \answerNA{}
    \item[] Justification: Foundational matrix analysis; no direct societal impact.
    
\item {\bf Safeguards}
    \item[] Question: Does the paper describe safeguards that have been put in place for responsible release of data or models that have a high risk for misuse (e.g., pre-trained language models, image generators, or scraped datasets)?
    \item[] Answer: \answerNA{}
    \item[] Justification: No such assets released.

\item {\bf Licenses for existing assets}
    \item[] Question: Are the creators or original owners of assets (e.g., code, data, models), used in the paper, properly credited and are the license and terms of use explicitly mentioned and properly respected?
    \item[] Answer: \answerYes{}
    \item[] Justification: All assets public and cited; UCI CC~BY~4.0, GPT-2 MIT, Qwen2.5 Apache~2.0, WikiText-2 CC~BY-SA~3.0.

\item {\bf New assets}
    \item[] Question: Are new assets introduced in the paper well documented and is the documentation provided alongside the assets?
    \item[] Answer: \answerNA{}
    \item[] Justification: None released.

\item {\bf Crowdsourcing and research with human subjects}
    \item[] Question: For crowdsourcing experiments and research with human subjects, does the paper include the full text of instructions given to participants and screenshots, if applicable, as well as details about compensation (if any)? 
    \item[] Answer: \answerNA{}
    \item[] Justification: None.

\item {\bf Institutional review board (IRB) approvals or equivalent for research with human subjects}
    \item[] Question: Does the paper describe potential risks incurred by study participants, whether such risks were disclosed to the subjects, and whether Institutional Review Board (IRB) approvals (or an equivalent approval/review based on the requirements of your country or institution) were obtained?
    \item[] Answer: \answerNA{}
    \item[] Justification: Not applicable.

\item {\bf Declaration of LLM usage}
    \item[] Question: Does the paper describe the usage of LLMs if it is an important, original, or non-standard component of the core methods in this research? Note that if the LLM is used only for writing, editing, or formatting purposes and does \emph{not} impact the core methodology, scientific rigor, or originality of the research, declaration is not required.
    \item[] Answer: \answerNA{}
    \item[] Justification: LLMs are study subjects only.

\end{enumerate}

\end{document}